\documentclass[10pt,twocolumn,letterpaper]{article}

\usepackage{cvpr}              

\usepackage{graphicx}
\usepackage{amsmath}
\usepackage{amssymb}
\usepackage{booktabs}
\usepackage{multirow}
\usepackage{microtype}
\usepackage{url}
\usepackage{color}
\usepackage[dvipsnames]{xcolor}
\usepackage{algorithm,algpseudocode}

\newcommand{\modelname}{OVTrack\xspace}
\newcommand{\parsection}[1]{\noindent\textbf{#1} }

\usepackage[pagebackref,breaklinks,colorlinks]{hyperref}

\definecolor{codeblue}{rgb}{0.25,0.5,0.5}

\algnewcommand\algorithmicinput{\textbf{Input:}}
\algnewcommand\INPUT{\item[\algorithmicinput]}
\algnewcommand\algorithmicinputt{\textbf{Track:}}
\algnewcommand\TRACK{\item[\algorithmicinputt]}
\newcommand{\LineComment}[1]{\hfill \textcolor{codeblue}{\# #1}}

\usepackage[capitalize]{cleveref}
\crefname{section}{Sec.}{Secs.}
\Crefname{section}{Section}{Sections}
\Crefname{table}{Table}{Tables}
\crefname{table}{Tab.}{Tabs.}

\def\cvprPaperID{3419} 
\def\confName{CVPR}
\def\confYear{2023}

\begin{document}

\title{OVTrack: Open-Vocabulary Multiple Object Tracking}

\author{
 Siyuan Li\thanks{Equal contribution.} \qquad
 Tobias Fischer\footnotemark[1] \qquad
 Lei Ke \qquad Henghui Ding \qquad \\
 Martin Danelljan \qquad Fisher Yu \\ 
 Computer Vision Lab, ETH Z{\"u}rich \\
  \url{https://www.vis.xyz/pub/ovtrack/} 
}

\maketitle

\begin{abstract}
The ability to recognize, localize and track dynamic objects in a scene is fundamental to many real-world applications, such as self-driving and robotic systems. Yet, traditional multiple object tracking (MOT) benchmarks rely only on a few object categories that hardly represent the multitude of possible objects that are encountered in the real world. This leaves contemporary MOT methods limited to a small set of pre-defined object categories. 
In this paper, we address this limitation by tackling a novel task, open-vocabulary MOT, that aims to evaluate tracking beyond pre-defined training categories. We further develop \modelname, an open-vocabulary tracker that is capable of tracking arbitrary object classes. Its design is based on two key ingredients: First, leveraging vision-language models for both classification and association via knowledge distillation; second, a data hallucination strategy for robust appearance feature learning from denoising diffusion probabilistic models.
The result is an extremely data-efficient open-vocabulary tracker that sets a new state-of-the-art on the large-scale, large-vocabulary TAO benchmark, while being trained solely on static images.
\end{abstract}


\section{Introduction}
\label{sec:intro}

\begin{figure}[t]
  \centering
   \includegraphics[width=1.0\linewidth]{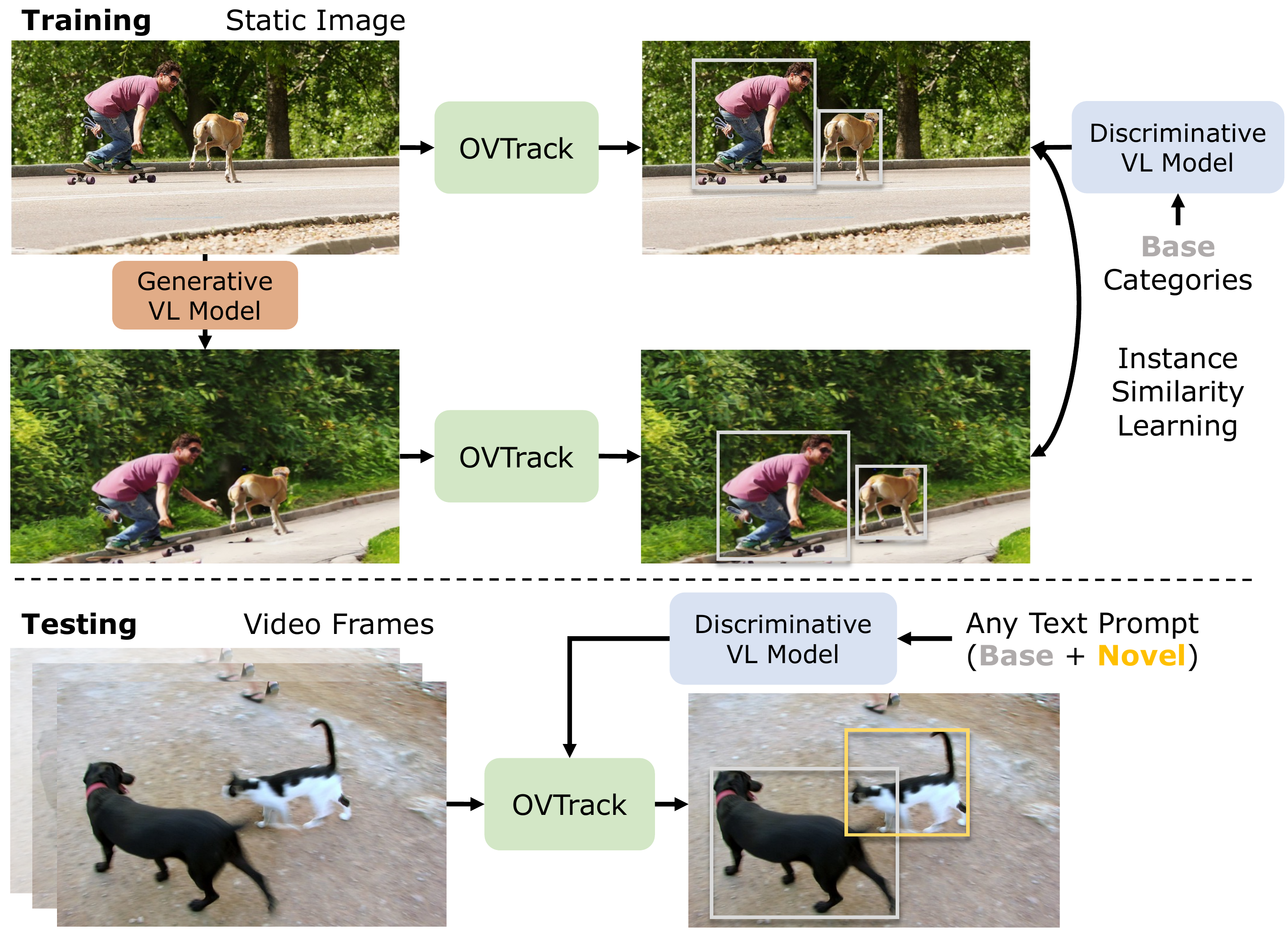}
    \vspace{-2mm}
   \caption{\textbf{OVTrack.} We approach the task of open-vocabulary multiple object tracking. During training, we leverage vision-language (VL) models both for generating samples and knowledge distillation. During testing, we track both base and novel classes unseen during training by querying a vision-language model.}
    \vspace{-2mm}
   \label{fig:teaser}
\end{figure}

Multiple Object Tracking (MOT) aims to recognize, localize and track objects in a given video sequence. It is a cornerstone of dynamic scene analysis and vital for many real-world applications such as autonomous driving, augmented reality, and video surveillance.
%
Traditionally, MOT benchmarks~\cite{dendorfer2021motchallenge, geiger2012we, bdd100k, dave2020tao, sun2020scalability} define a set of semantic categories that constitute the objects to be tracked in the training and testing data distributions. The potential of traditional MOT methods~\cite{leibe2008robust, milan2013continuous, bewley2016simple, bergmann2019tracking} is therefore limited by the taxonomies of those benchmarks. As consequence, contemporary MOT methods struggle with unseen events, leading to a gap between evaluation performance and real-world deployment.

To bridge this gap, previous works have tackled MOT in an open-world context. In particular, O\v{s}ep~\etal~\cite{ovsep2016multi, ovsep2018track} approach generic object tracking by first segmenting the scene and performing tracking before classification. Other works have used class agnostic localizers~\cite{dave2019towards, ovsep20204d} to perform MOT on arbitrary objects.
Recently, Liu~\etal~\cite{Liu_2022_CVPR} defined open-world tracking, a task that focuses on the evaluation of previously unseen objects. In particular, it requires any-object tracking as a stage that precedes object classification.
This setup comes with two inherent difficulties. First, in an open-world context, densely annotating all objects is prohibitively expensive. Second, without a pre-defined taxonomy of categories, the notion of \textit{what is} an object is ambiguous. As a consequence, Liu~\etal resort to recall-based evaluation, which is limited in two ways. Penalizing false positives (FP) becomes impossible,~\ie we cannot measure the tracker \textit{precision}. Moreover, by evaluating tracking in a class-agnostic manner, we lose the ability to evaluate how well a tracker can infer the semantic category of an object.

In this paper, we propose open-vocabulary MOT as an effective solution to these problems.
Similar to open-world MOT, open-vocabulary MOT aims to track multiple objects beyond the pre-defined training categories.
However, instead of dismissing the classification problem and resorting to recall-based evaluation, we assume that \textit{at test time} we are given the classes of objects we are interested in.
This allows us to apply existing closed-set tracking metrics~\cite{yang2019video, li2022tracking} that capture both precision and recall, while still evaluating the tracker's ability to track arbitrary objects during inference.

We further present the first \textbf{O}pen-\textbf{V}ocabulary \textbf{T}racker, \modelname (see Fig.~\ref{fig:teaser}). 
To this end, we identify and address two fundamental challenges to the design of an open-vocabulary multi-object tracker.
The first is that closed-set MOT methods are simply not capable of extending their pre-defined taxonomies. The second is data availability,~\ie scaling video data annotation to a large vocabulary of classes is extremely costly. 
Inspired by existing works in open-vocabulary detection~\cite{bansal2018zero, gu2021open, zhou2022detecting, du2022learning}, we replace our classifier with an embedding head, which allows us to measure similarities of localized objects to an open vocabulary of semantic categories. In particular, we distill knowledge from CLIP~\cite{radford2021learning} into our model by aligning the image feature representations of object proposals with the corresponding CLIP image and text embeddings. 

Beyond detection, association is the core of modern MOT methods. It is driven by two affinity cues: motion and appearance. In an open-vocabulary context, motion cues are brittle since arbitrary scenery contains complex and diverse camera and object motion patterns. 
In contrast, diverse objects usually exhibit heterogeneous appearance.
However, relying on appearance cues requires robust representations that generalize to novel object categories.
We find that CLIP feature distillation helps in learning better appearance representations for improved association.
This is especially intriguing since object classification and appearance modeling are usually distinct in the MOT pipeline~\cite{wojke2017simple, bergmann2019tracking, fischer2022qdtrack}.

Learning robust appearance features also requires strong supervision that captures object appearance changes in different viewpoints, background, and lighting. 
To approach the data availability problem, we utilize the recent success of denoising diffusion probabilistic models (DDPMs) in image synthesis~\cite{ramesh2021zero, Rombach_2022_CVPR} and propose an effective data hallucination strategy tailored to appearance modeling. In particular, from a static image, we generate both simulated positive and negative instances along with random background perturbations.

The main contributions are summarized as follows:\vspace{-2mm}
\begin{enumerate}
\setlength\itemsep{0em}
    \item  We define the task of open-vocabulary MOT and provide a suitable benchmark setting on the large-scale, large-vocabulary MOT benchmark TAO~\cite{dave2020tao}.
    \item We develop \modelname, the first open-vocabulary multi-object tracker. It leverages vision-language models to improve both classification and association compared to closed-set trackers.
    \item We propose an effective data hallucination strategy that allows us to address the data availability problem in open-vocabulary settings by leveraging DDPMs.
\end{enumerate}
Owing to its thoughtful design, \modelname sets a new state-of-the-art on the challenging TAO benchmark~\cite{dave2020tao}, outperforming existing trackers by a significant margin while being trained \textit{on static images only}. In addition, \modelname is capable of tracking \textit{arbitrary} object classes (see Fig.~\ref{fig:qualitative}), overcoming the limitation of closed-set trackers.
\begin{figure}[t]
    \centering
    \footnotesize
    \setlength\tabcolsep{0.2mm}
    \resizebox{1.0\linewidth}{!}{
    \begin{tabular}{ccc}
        \toprule
        $t$ & $t +2$ & $t + 4$\\ \midrule
        \includegraphics[width=0.34\linewidth]{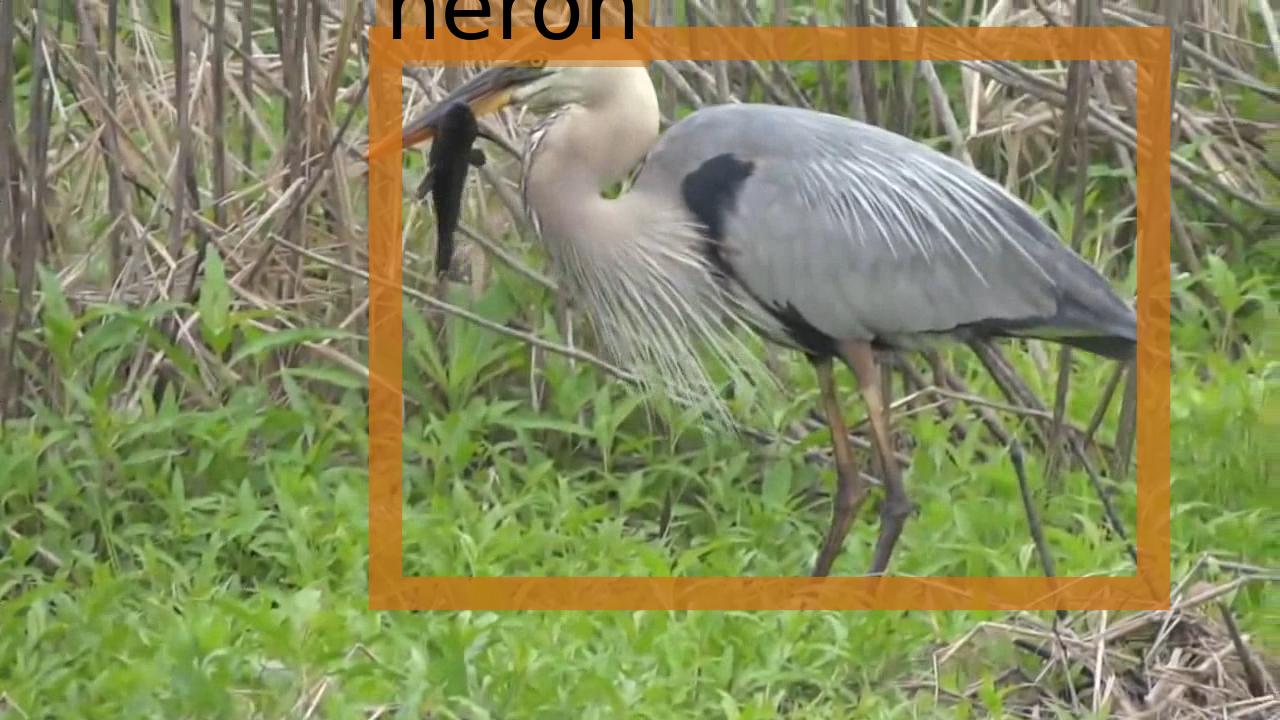} &
        \includegraphics[width=0.34\linewidth]{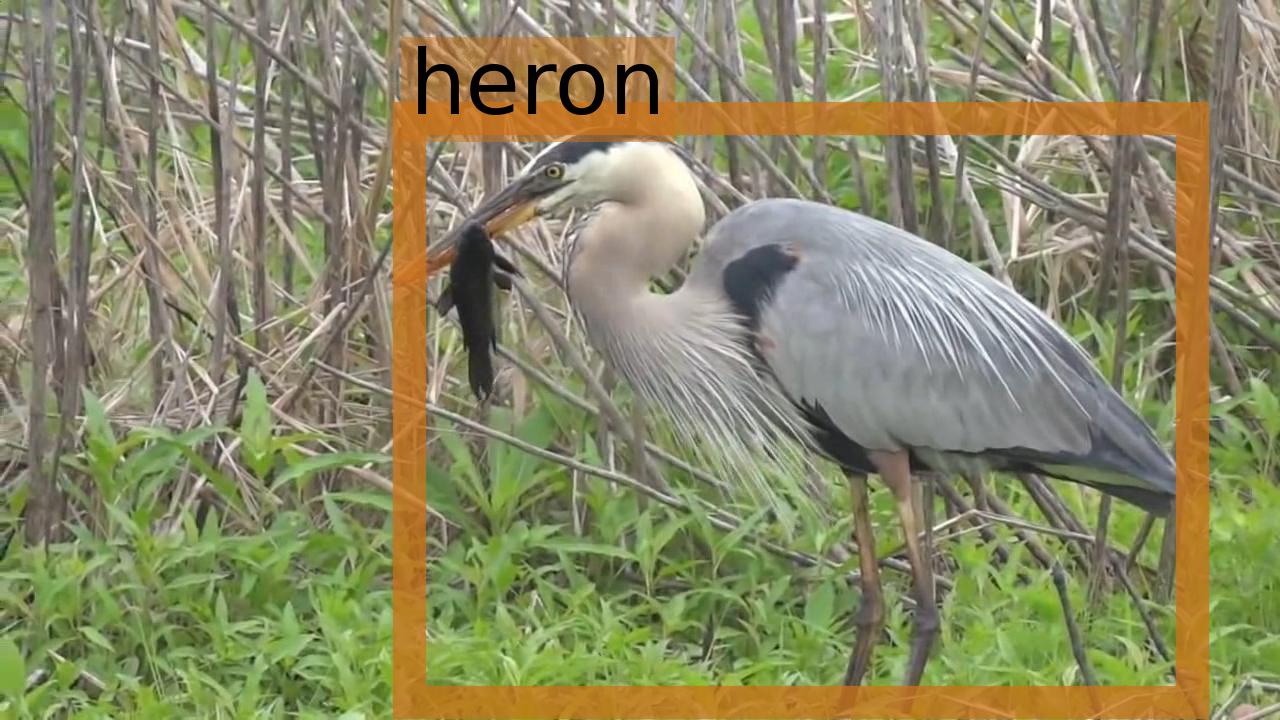} &
        \includegraphics[width=0.34\linewidth]{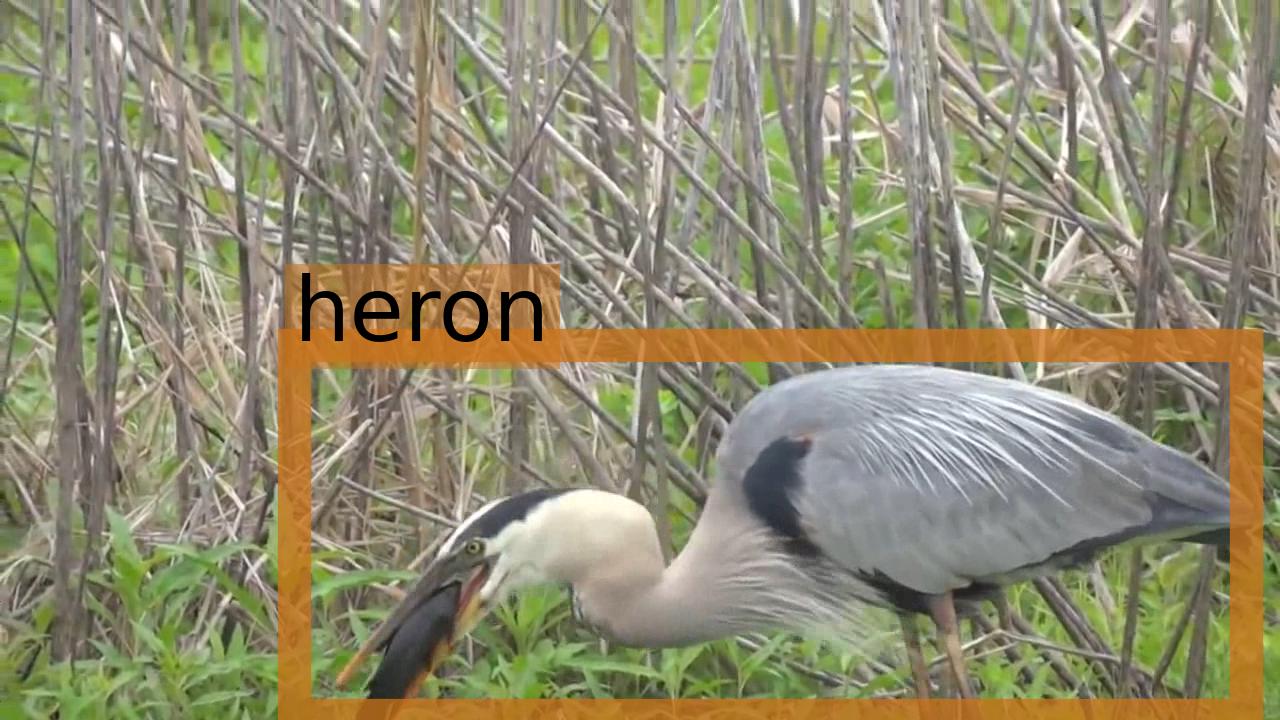} \\
        \includegraphics[width=0.34\linewidth]{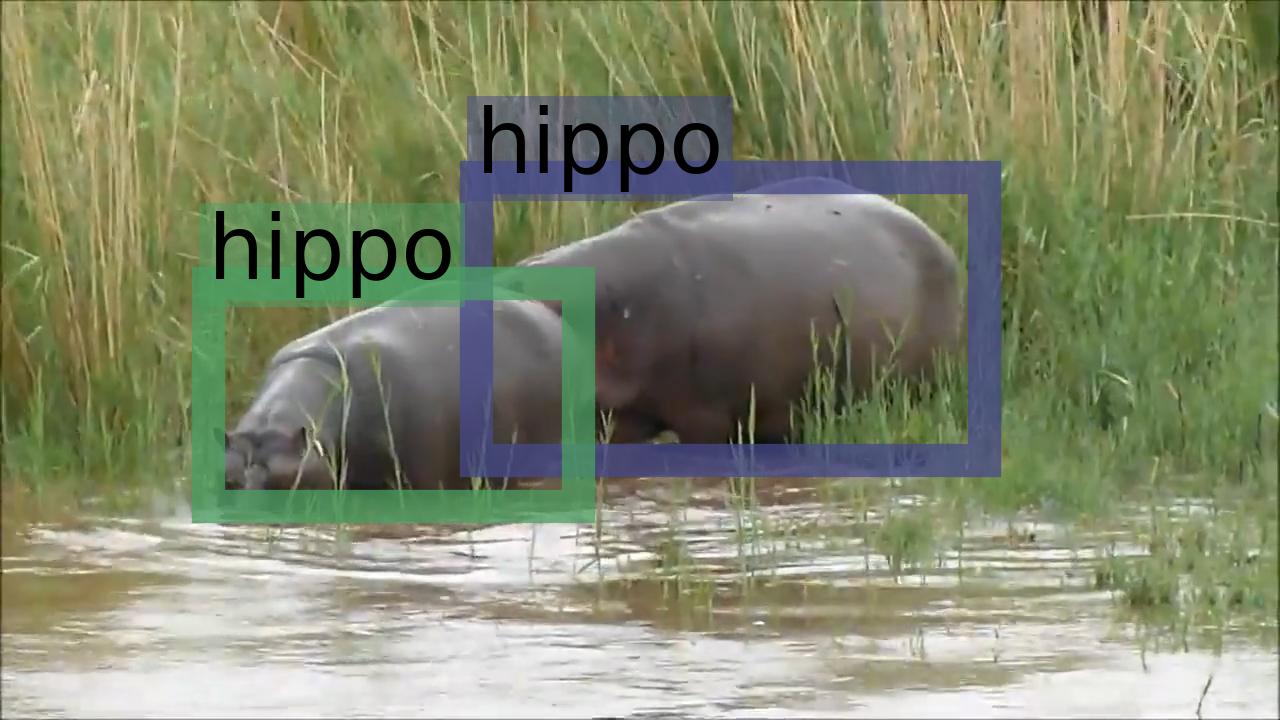} &
        \includegraphics[width=0.34\linewidth]{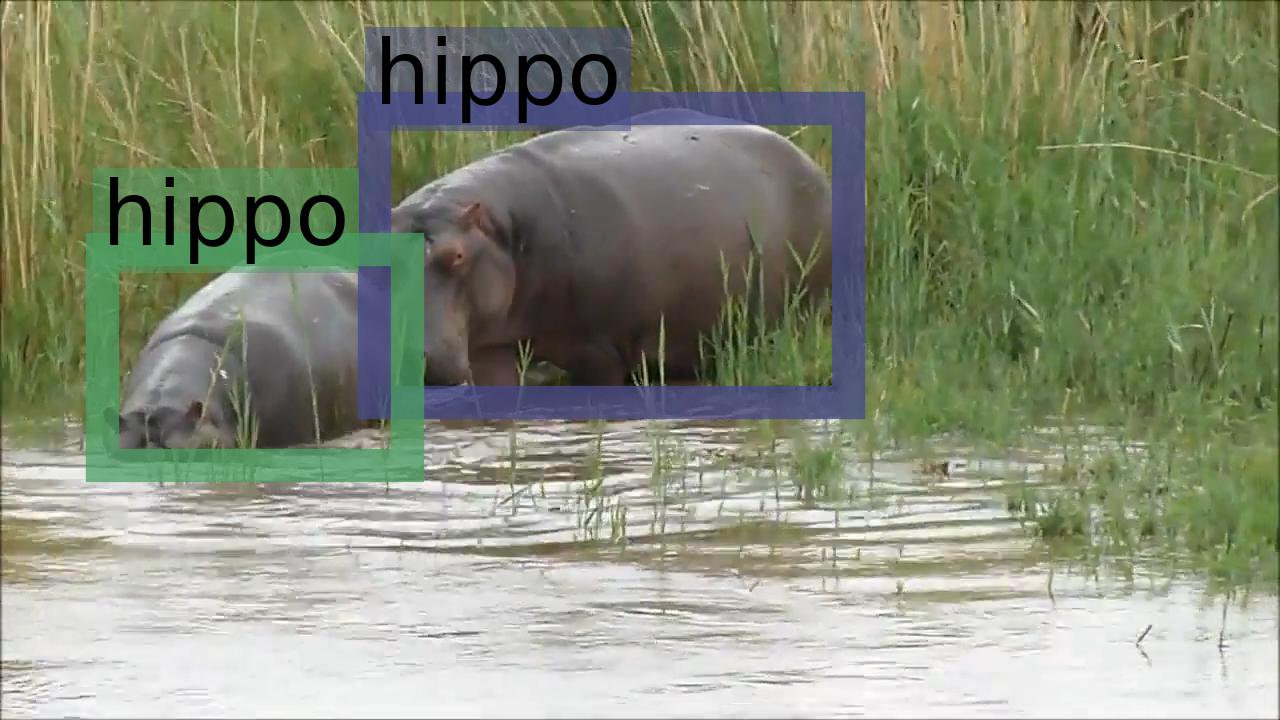} &
        \includegraphics[width=0.34\linewidth]{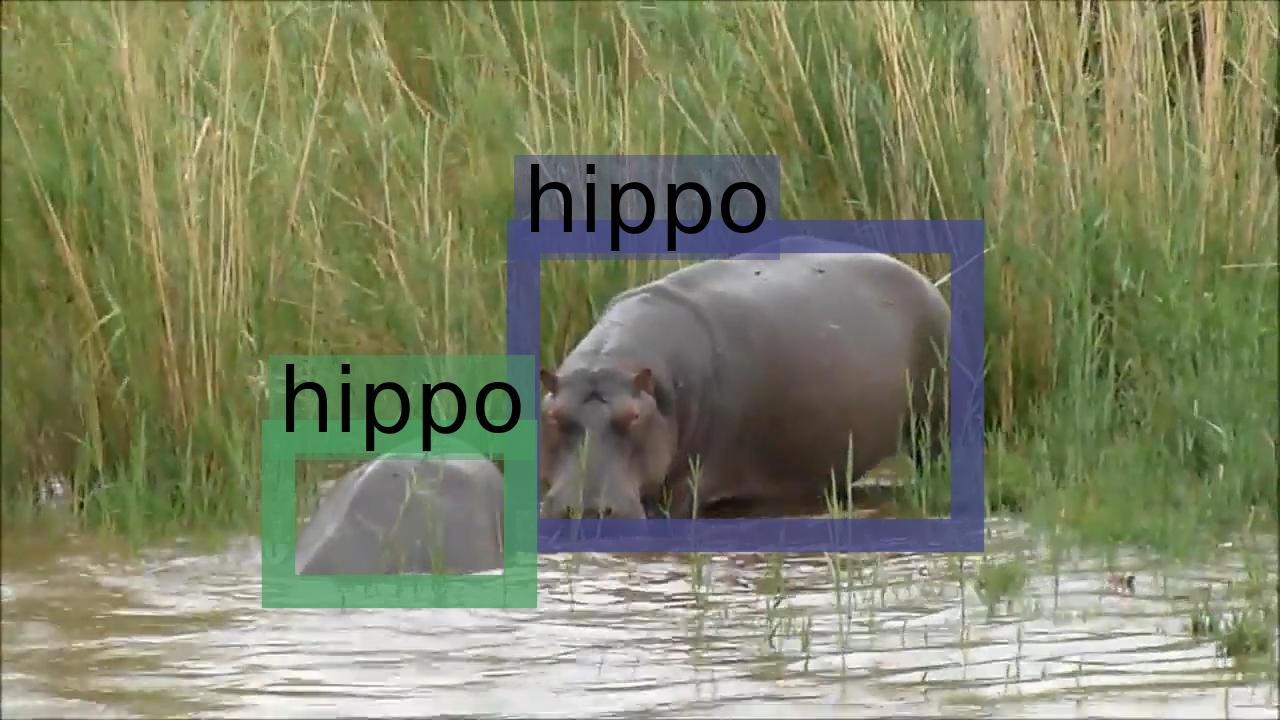} \\
        \includegraphics[width=0.34\linewidth]{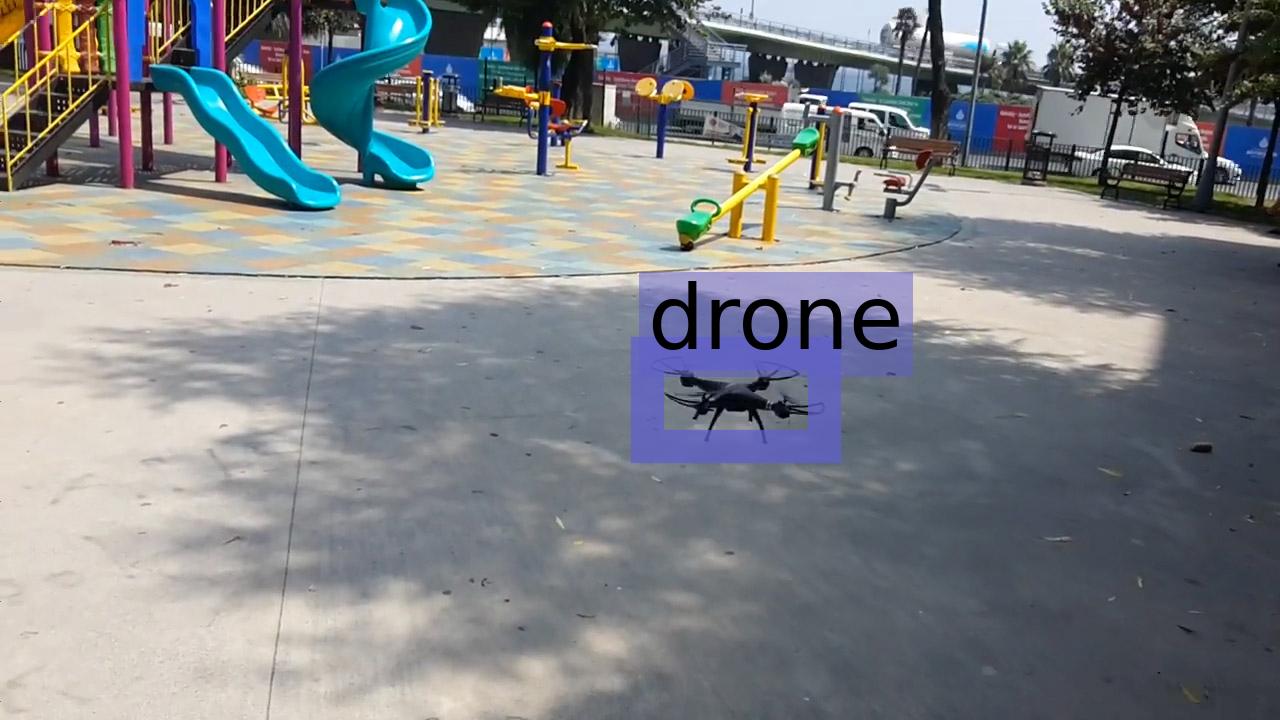} &
        \includegraphics[width=0.34\linewidth]{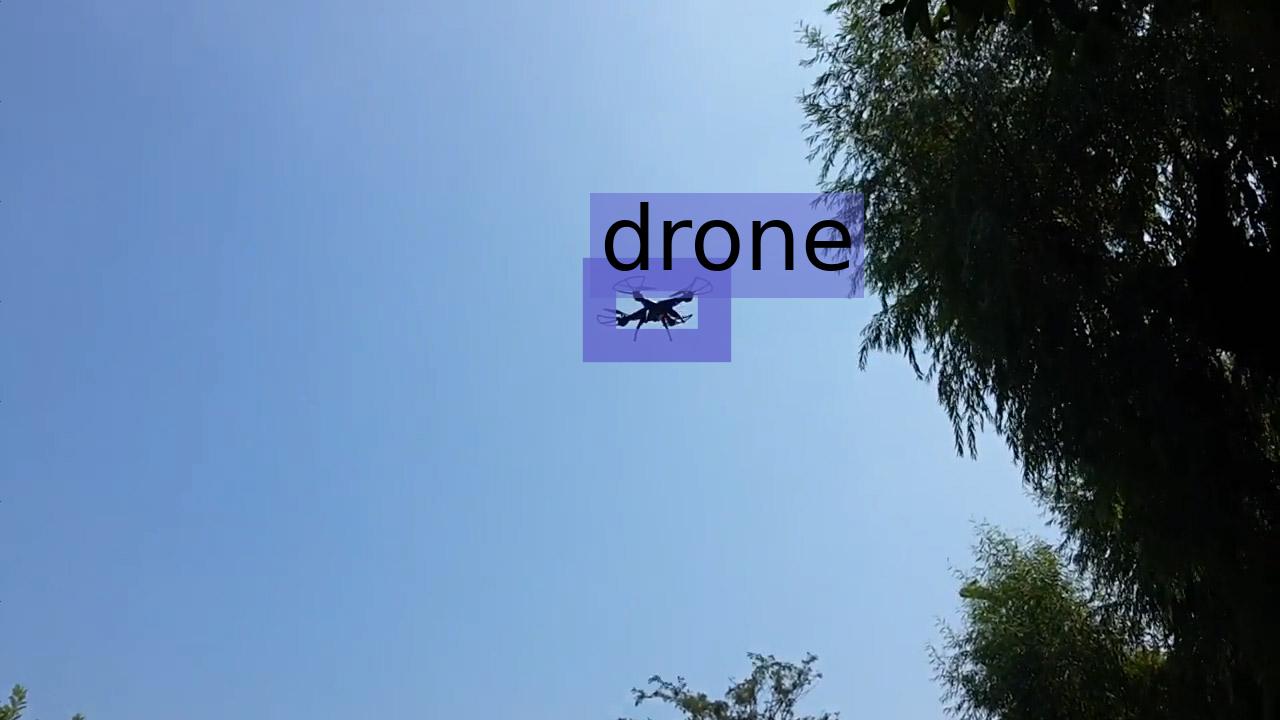} &
        \includegraphics[width=0.34\linewidth]{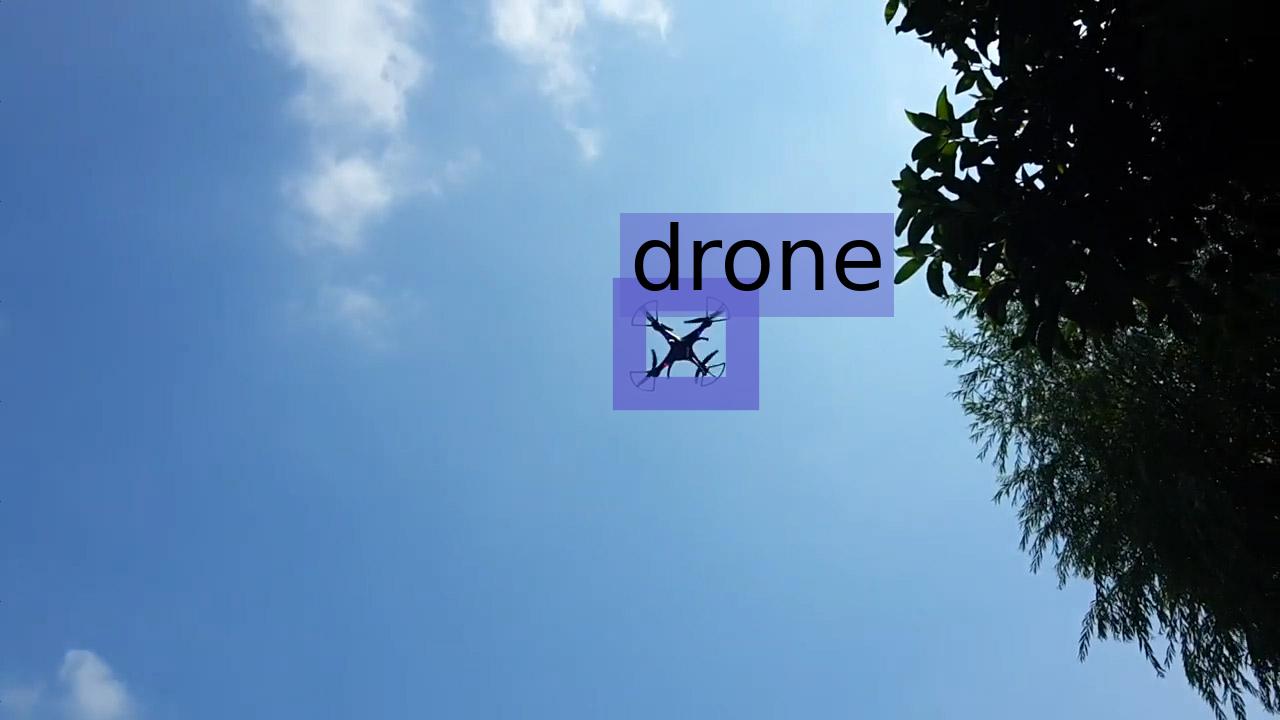} \\
        \bottomrule
    \end{tabular}}\vspace{-2mm}%
    \caption{\textbf{\modelname qualitative results.} We condition our tracker on text prompts unseen during training, namely `heron', `hippo' and `drone', and successfully track the corresponding objects in the videos. The box color depicts object identity.
    }
    \label{fig:qualitative}
    \vspace{-2mm}
\end{figure}


\section{Related work}

\parsection{Multiple object tracking.}
The dominant paradigm in MOT literature is tracking-by-detection~\cite{ramanan2003finding}, where objects are first detected in each frame, and subsequently associated across time. Thus, many works have focused on data association, aiming to exploit similarity cues such as visual appearance~\cite{crf, laura2016, amir2017, anton2017, wojke2017simple, bergmann2019tracking, pang2021quasi, fischer2022qdtrack}, 2D object motion~\cite{bewley2016simple, xiao2018simple, ioutracker, goturn, d&t} or 3D object motion~\cite{mitzel2012taking, held2013precision, osep2017combined, ovsep2018track, beyondpixels, luiten2020track} most effectively.
Recently, researchers have focused on learning data association with graph neural networks~\cite{braso2020learning, schulter2017deep} or transformers~\cite{meinhardt2022trackformer, sun2020transtrack, zeng2021motr, zhou2022global}.
However, those works dismiss a more profound problem in the tracking-by-detection pipeline that precedes data association: Contemporary object detectors~\cite{frcnn, maskrcnn, liu2016ssd, redmon2016you, redmon2017yolo9000} are designed for closed-set scenarios where all objects appear frequently in the training \textit{and} testing data distributions.
Hence, Dave~\etal~\cite{dave2020tao} proposed a new benchmark, TAO, that focuses on studying MOT in the long-tail of the object category distribution. On this benchmark, GTR~\cite{zhou2022global}, AOA~\cite{du2021aoa}, QDTrack~\cite{fischer2022qdtrack} and TET~\cite{li2022tracking} achieve impressive performance. 
However, those works are still limited to pre-defined object categories and thus do not scale to the diversity of real-world settings. Our work enables tracking of unseen classes from an open vocabulary.

\parsection{Open-world detection and tracking.}
Open-world detection methods aim to detect any salient object in a given input image \textit{irrespective} of its category and \textit{beyond} the training data distribution in particular.
However, object classification under such a setting is ill-posed since novel classes will be unknown by definition~\cite{bendale2015towards, joseph2021towards}.
As such, open-world detection methods utilize class agnostic localizers~\cite{dave2019towards} and treat classification as a clustering problem~\cite{joseph2021towards}, estimating a similarity between novel instances and grouping them into novel classes via incremental learning.

Instead, open-vocabulary object detection methods aim to detect arbitrary, but \textit{given} classes of objects at test time~\cite{zareian2021open}.
For this, Bansal~\etal~\cite{bansal2018zero} connect an object detector with word representations~\cite{pennington2014glove}.
Recently, models like CLIP~\cite{radford2021learning} learn visual representations from natural language supervision. Their main advantage over word representations is better alignment of visual concepts and language description. Consequently, many works have focused on leveraging image-text representations for open-vocabulary and few-shot object detection~\cite{gu2021open, zhou2022detecting}. ViLD~\cite{gu2021open} distills CLIP image features, while Detic~\cite{zhou2022detecting} leverages classification data for joint training. Other works have focused on learning good language prompts for open-vocabulary object detection~\cite{du2022learning}.

Fewer works have tackled the open-world problem in the MOT domain. Existing works perform scene segmentation and class agnostic tracking before classification~\cite{mitzel2012taking, ovsep2016multi, ovsep2018track} or utilize class-agnostic proposal generation~\cite{dave2019towards, ovsep20204d}, similar to open-world detection methods.
Liu~\etal~\cite{Liu_2022_CVPR} propose an open-world tracking benchmark, TAO-OW, that evaluates class-agnostic tracking as a task that precedes classification. However, this comes with the limitation that the evaluation only captures tracker recall and no classification accuracy.
Instead of dismissing classification, we pose the problem in a different way,~\ie \textit{at test time} we know the novel classes we are interested in. This allows us to capture both the precision and recall of novel classes in our evaluation, while our method maintains the ability to track any object.

\parsection{Learning tracking from static images.}
Since labelled video data is expensive to acquire at scale, recent methods have proposed to use static images to supervise MOT methods~\cite{zhou2020tracking, zhang2021fairmot, woobridging, fischer2022qdtrack}.
CenterTrack~\cite{zhou2020tracking} proposes to learn motion offsets from static images by random translation of the input, while FairMOT~\cite{zhang2021fairmot} treats objects in a dataset of static images as unique classes to distinguish. 
Inspired by recent progress in self-supervised representation learning~\cite{oord2018representation, he2020momentum, chen2020simple}, Fischer~\etal~\cite{fischer2022qdtrack} propose to utilize data augmentation in combination with a contrastive learning objective to learn appearance-based tracking from static images.
We go beyond classic data augmentation used in existing works by generating positive and negative examples of objects along with background perturbations via generative models, offering a more targeted approach to guiding appearance similarity learning from static images.

\parsection{Data generation for tracking.}
While aforementioned methods alleviate the data availability problem in MOT, there is still room for improvement when it comes to data generation in video tasks. Therefore, a large body of research has focused on data generation strategies that can benefit tracking methods~\cite{gaidon2016virtual, richter2016playing, khoreva2019lucid, fabbri2021motsynth, chen2021geosim, kim2021drivegan, hu2022monocular}.
While early works focused on obtaining synthetic data from computer graphics engines~\cite{gaidon2016virtual, richter2016playing, fabbri2021motsynth, hu2022monocular}, newer approaches combine 3D assets with generative models for improved realism~\cite{kim2021drivegan, chen2021geosim}. Fewer works have tackled data generation with generative models only~\cite{khoreva2019lucid}.
Recently, DDPMs~\cite{ho2020denoising, ramesh2021zero, Rombach_2022_CVPR} showed impressive results in image synthesis. We leverage their data generation fidelity to address the data availability problem that is particularly pronounced in open-vocabulary MOT with a novel data hallucination strategy tailored to appearance modeling.

\begin{figure*}[t]
  \centering
  \vspace{-0.1in}
   \includegraphics[width=1.0\linewidth]{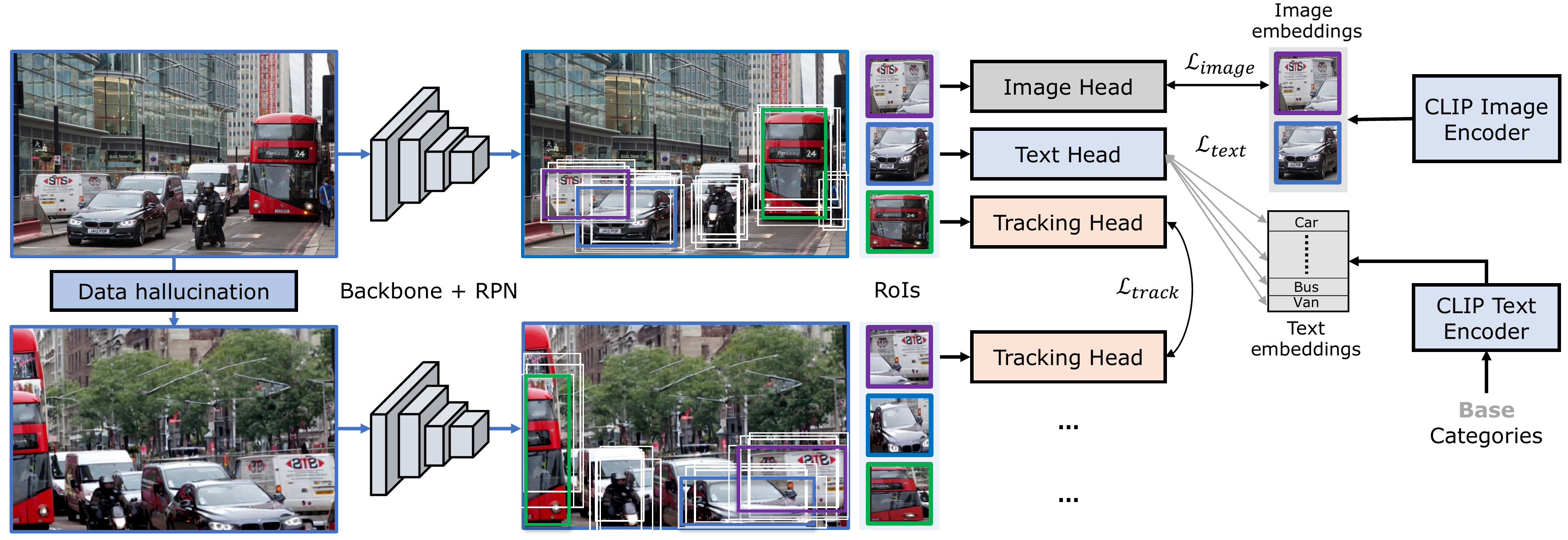}
\vspace{-6mm}
   \caption{\textbf{\modelname training.} From a single static $I_\text{key}$, we generate $I_\text{ref}$ with our data hallucination strategy. We extract RoIs via a RPN~\cite{frcnn} and perform knowledge distillation from CLIP~\cite{radford2021learning} via the embeddings of the text and image heads. Note that we train classification only on $\mathcal{C}^\text{base}$. Further, we obtain appearance embeddings from the tracking head and apply our instance similarity loss on the image pair.}
   \vspace{-2mm}
   \label{fig:training}
\end{figure*}

\section{Open-Vocabulary MOT}
In real-world scenarios, object categories follow a long-tailed distribution with a rich vocabulary. The remarkable diversity of the open world cannot be covered by a monolithic dataset.
However, existing MOT benchmarks focus on closed-set evaluation, with often only a handful of object classes being evaluated. Furthermore, the task setup requires trackers to only track objects within a small set of training categories.
To bridge the gap between existing MOT benchmarks and algorithms and real-world settings, we propose the task of open-vocabulary MOT and define its training and evaluation setup as follows. 

At training time, we train a tracker $M$ on the training data distribution $\mathcal{D}^\text{train} = \{ \mathbf{X}^\text{train}, \mathcal{A}^\text{train} \}$ that contains video sequences $\mathbf{X}^\text{train}$ and their respective annotations $\mathcal{A}^\text{train}$ of objects with semantic categories $\mathcal{C}^\text{base} \subset \mathbb{N}$. Each annotation $\mathbf{\alpha} \in \mathcal{A}^\text{train}$ consists of a set of states $\{\alpha_t\}_{t \in T}$ for each frame $t \in T$ that the object is visible in. A state $\alpha_t = ( \mathbf{b}_t, c_t )$ comprises the object class $c \in \mathbb{N}$ and the 2D bounding box $\mathbf{b} = [x, y, w, h]$, where $(x, y)$ is the center location in pixel coordinates and $(w, h)$ are width and height, respectively.
At test time, we are given video sequences $\mathbf{X}^\text{test}$ and a set of object classes $\mathcal{C}^\text{novel} \subset \mathbb{N} \setminus \mathcal{C}^\text{base}$ that we are interested in. 
We aim to find all tracks $\mathcal{T}$ of objects in $\mathbf{X}^\text{test}$ belonging to classes $\mathcal{C}^\text{base} \cup \mathcal{C}^\text{novel}$. Each track state $\tau_t = ( \mathbf{b}_t, p_t, c_t ) \in \mathcal{T}$ contains predicted object confidence $p \in [0, 1]$, class $c \in \mathbb{N}$ and 2D bounding box $\mathbf{b} = [x, y, w, h]$.
The important distinctions to closed-set tracking are two-fold: \textbf{1)} We evaluate the tracker $M$ not only on $\mathcal{C}^\text{base}$ but also on $\mathcal{C}^\text{novel}$ with $\mathcal{C}^\text{novel} \cap \mathcal{C}^\text{base} = \emptyset$, and \textbf{2)} While $\mathcal{C}^\text{novel}$ is known at test time, our setup requires the tracker $M$ to track arbitrary object classes $c \in \mathbb{N}$ since $\mathcal{C}^\text{novel}$ remains unknown \textit{at training time}. In particular, the evaluation of $\mathcal{C}^\text{novel}$ illustrates the ability of tracker $M$ to track \textit{any} unknown class $c \in \mathbb{N} \setminus \mathcal{C}^\text{base}$, while the classes in $\mathcal{C}^\text{novel}$ serve \textit{as proxy}.

\subsection{Benchmark}
\label{sec:benchmark}
We utilize the large-scale, large-vocabulary MOT dataset TAO~\cite{dave2020tao} to establish a suitable benchmark for open-vocabulary MOT.
TAO mostly follows the taxonomy of LVIS~\cite{gupta2019lvis}, which divides classes according to their occurrence into frequent, common and rare classes.
To obtain our held-out set $\mathcal{C}^\text{novel}$, we follow open-vocabulary detection literature~\cite{gu2021open} and use the rare classes as defined by LVIS.
The intuition behind this is that the occurrence of rare classes is correlated with uncommon scenarios and events that we are particularly interested in evaluating.

With respect to evaluation, the advantage of defining $C^\text{novel}$ is that we can apply closed-set tracking metrics in a straightforward manner, while open-world MOT~\cite{Liu_2022_CVPR} needs to resort to recall-based evaluation.
Further, previous works~\cite{Liu_2022_CVPR, li2022tracking} have shown that the official evaluation metric in TAO, Track mAP~\cite{yang2019video}, is sub-optimal in terms of handling FPs in presence of missing annotations. On the contrary, the recently proposed TETA metric~\cite{li2022tracking} handles this shortcoming via local cluster evaluation. Also, TETA disentangles classification from localization and association performance. Thus, we choose TETA as the evaluation metric for our setup to provide a comprehensive insight into the localization, association, and open-vocabulary classification performance of tracker $M$.

\section{\modelname}
\label{sec:design}
We present our \textbf{O}pen-\textbf{V}ocabulary \textbf{T}racker, \modelname. We address two perspectives of its design: \textbf{1)} \textit{Model perspective}: We show how to handle the open-vocabulary setting in the localization, classification, and association modules of the tracker in Section~\ref{sec:design}; \textbf{2)} \textit{Data perspective}: Collecting and annotating the necessary amount of training videos is impractical for open-vocabulary MOT. Therefore, we contribute a novel training approach for learning object tracking \textbf{without} video data in Section~\ref{sec:data}.

\begin{figure*}[t]
  \centering
   \includegraphics[width=1.0\linewidth]{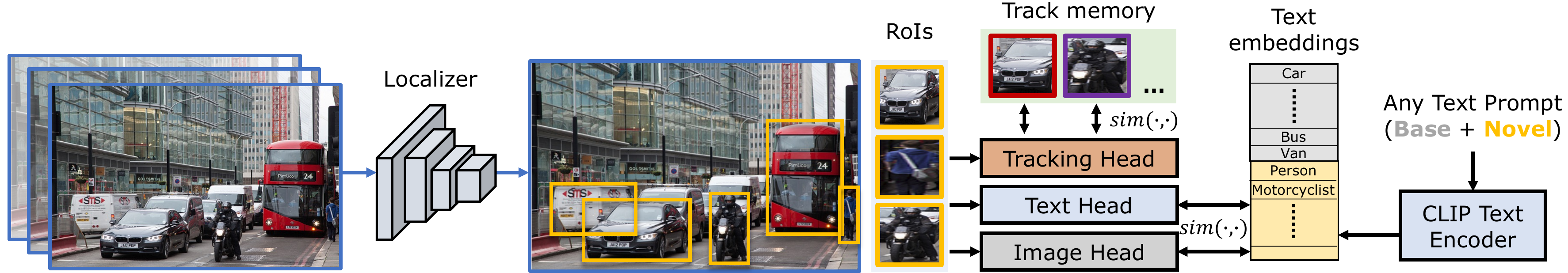}
\vspace{-6mm}
   \caption{\textbf{\modelname inference.} Given an input video stream, we track objects of arbitrary classes,~\eg $\mathcal{C}^\text{base} \cup \mathcal{C}^\text{novel}$. We first localize objects agnostic of their class, then assign a semantic class label via the text embedding head, and finally associate them to existing tracks by comparing their appearance embedding $\mathbf{q}$ obtained from the tracking head to the appearance embeddings in the track memory.}
   \vspace{-1mm}
   \label{fig:inference}
\end{figure*}
\subsection{Model design}
\label{sec:design}
We decompose \modelname's functionality into localization, classification and association and discuss our open-vocabulary design philosophy in tackling difficulties for each of those parts. The model design is illustrated in Fig.~\ref{fig:training}.

\parsection{1) Localization:}
To localize objects of arbitrary and possibly unknown classes $c \in \mathbb{N}$ in a video, we train Faster R-CNN~\cite{frcnn} in a class-agnostic manner,~\ie we use only the RPN and regression losses defined in~\cite{frcnn}.
We find that this localization procedure can generalize well to object classes that are unknown at training time, as also validated by previous works~\cite{dave2019towards, gu2021open, zhou2022detecting}.
During training, we use RPN proposals as object candidates $P$ for greater diversity, while during inference, we use the refined RCNN outputs as object candidates. Each candidate $r \in P$ is defined by confidence $p_r$ and bounding box $\mathbf{b}_r$.

\parsection{2) Classification:}
Existing closed-set trackers~\cite{bewley2016simple, bergmann2019tracking, fischer2022qdtrack, zhou2020tracking} can only track objects of categories in $\mathcal{C}^\text{base}$,~\ie objects present and annotated in the training data distribution $\mathcal{D}^\text{train}$. 
To enable open-vocabulary classification, we need to be able to configure the classes we are interested in without re-training.
Inspired by open-vocabulary detection literature~\cite{bansal2018zero}, we connect our Faster R-CNN with the vision-language model CLIP~\cite{radford2021learning} that has been pre-trained on over 400 million image-text pairs for contrastive learning. 

After extracting the RoI feature embeddings $\mathbf{f}_r=\mathcal{R}(\phi(I), \mathbf{b}_r), \forall r \in P$ from the backbone $\phi$, we replace the original classifier in Faster R-CNN with a text head and add an image head generating the embeddings $\hat{\mathbf{t}}_r$ and $\hat{\mathbf{i}}_r$, for each $\mathbf{f}_r$.
We use the CLIP text and image encoders to supervise the heads following~\cite{du2022learning,gu2021open}.
In particular, we use the class names to generate text prompts $\mathcal{P}(c) = \{\mathbf{v}_1^c, ..., \mathbf{v}_L^c, \mathbf{w}_c  \}$ that consist of $L$ context vectors $\mathbf{v}^c$ and a class name embedding $\mathbf{w}_c$.
We feed the prompts into the CLIP text encoder $\mathcal{E}$, generating text embeddings $\mathbf{t}_c = \mathcal{E}(\mathcal{P}(c)), \forall c \in \mathcal{C}^\text{base}$.
We compute the affinity between the predicted embeddings $\hat{\mathbf{t}}_r$ and their CLIP counterpart $\mathbf{t}_c$. 
\begin{align}
\mathbf{z}(r)&\!=\![\operatorname{cos}(\hat{\mathbf{t}}_r, \mathbf{t}_{bg}), \operatorname{cos}(\hat{\mathbf{t}}_r, \mathbf{t}_1), \cdots, \operatorname{cos}(\hat{\mathbf{t}}_r, \mathbf{t}_{|\mathcal{C}^\text{base}|})] \\
\mathcal{L}_\text{text}&=\frac{1}{|P|} \sum_{r \in P} \mathcal{L}_{\mathrm{CE}}(\operatorname{softmax}(\mathbf{z}(r) / \lambda), c_r),
\end{align}
where $\operatorname{cos}(\mathbf{v}, \mathbf{k}) = \frac{\textbf{v} \cdot \textbf{k}}{||\textbf{v}|| ||\textbf{k}||}$, $\mathbf{t}_{bg}$ a learned background prompt, $\lambda$ a temperature parameter, $\mathcal{L}_{\mathrm{CE}}$ the cross-entropy loss and $c_r$ is the class label of $r$. 
Furthermore, we align each $\hat{\mathbf{i}}_r$ with the CLIP image encoder $\mathcal{I}$. For each $r$, we crop the input image to $\mathbf{b}_r$, and resize it to the required input size to obtain the image embedding $\mathbf{i}_r = \mathcal{I}(\mathcal{R}(I, \mathbf{b}_r))$.
We minimize the distance between the corresponding $\hat{\mathbf{i}}_r$ and $\mathbf{i}_r$.
\begin{equation}
    \mathcal{L}_\text{image} = \frac{1}{|P|} \sum_{r \in P} ||\hat{\mathbf{i}}_r - \mathbf{i}_r||_1.
\end{equation}

\parsection{3) Association:}
An open-vocabulary tracker should handle diverse scenarios that comprise complex camera motion and heterogeneous object motion patterns. However, those patterns are difficult to model especially when there are not enough video annotations available~\cite{Liu_2022_CVPR, fischer2022qdtrack}. 
Therefore, we rely on appearance cues to robustly track objects in an open-vocabulary context.
Specifically, we employ a contrastive learning approach inspired by~\cite{fischer2022qdtrack, li2022tracking}. Given an image pair $( I_\text{key}, I_\text{ref} )$ we extract RoIs from both images and match the RoIs to the annotations using intersection-over-union (IoU).
For each matched RoI in $I_\text{key}$ with appearance embedding $\mathbf{q} \in Q$, we cluster objects $Q^+$ with the same identity and divide objects $Q^-$ with different identity in $I_\text{ref}$.
\begin{align}
&\mathrm{PosD}(\mathbf{q}) = \frac{1}{\left | Q^+(\mathbf{q})) \right |}\sum_{\mathbf{q}^+ \in Q^+}\mathrm{exp}(\mathbf{q} \cdot \mathbf{q}^+)/\tau),\\
&\mathrm{Sim}(\mathbf{q}) = \frac{\mathrm{exp}(\mathbf{q} \cdot \mathbf{q}^+ / \tau)}{\mathrm{PosD}(\mathbf{q})+ \sum_{\mathbf{q}^- \in Q^-}\mathrm{exp}(\mathbf{q} \cdot \mathbf{q}^- / \tau)},\\
&\mathcal{L}_\text{track} = -\sum_{\mathbf{q} \in Q} \frac{1}{|Q^+(\mathbf{q})|}\sum_{\mathbf{q}^+ \in Q^+(\mathbf{q})}\mathrm{log}(\mathrm{Sim}(\mathbf{q}^+)).
\label{eq:multipos}
\end{align}
We further apply an auxiliary loss $\mathcal{L}_\text{aux}$ to constrain the magnitude of the logits following~\cite{fischer2022qdtrack}. 

During inference, we use straightforward appearance feature similarity for associating existing tracks $\mathcal{T}$ with objects in $P$. In particular, for each track $\tau \in \mathcal{T}$ and its corresponding appearance embedding $\mathbf{q}_{\tau}$, we compare its similarity with all candidate objects $r \in P$ using appearance embedding $\mathbf{q}_{r}$. We measure the similarity $\textbf{s}(\tau, r)$ of existing tracks with the candidate objects using both bi-directional softmax~\cite{fischer2022qdtrack} and cosine similarity.
We assign $r$ to the track $\tau$ with its maximum similarity, if $\textbf{s}(\tau, r) > \beta$. If $r$ does not have a matching track, it starts a new track if its confidence $p_{r} > \gamma$ and is discarded otherwise. The inference pipeline is illustrated in Fig.~\ref{fig:inference}.

\begin{figure*}[t]
  \centering
   \includegraphics[width=1.0\linewidth]{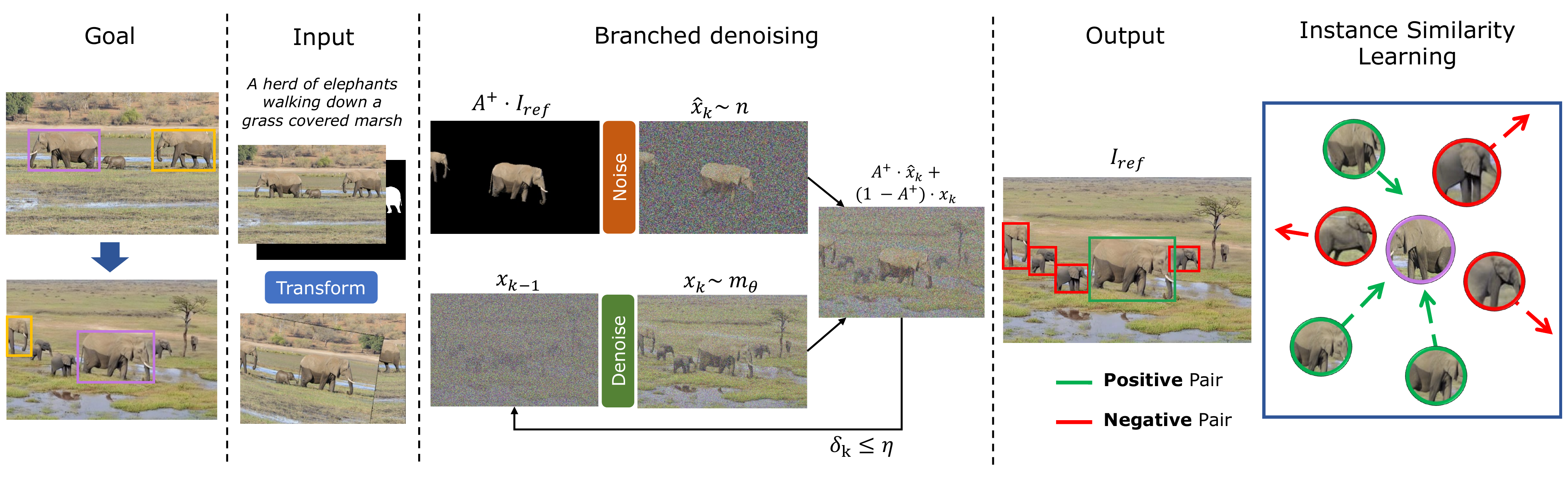}
    \vspace{-6mm}
   \caption{\textbf{Data hallucination strategy.} Given an input image, its annotations and its caption, we generate $x_0 \sim n$ and input it to the diffusion model~\cite{Rombach_2022_CVPR}, which progressively denoises $x$ from $\delta_0 = 0.75$ to $\eta$. At the same time, the foreground regions are kept fixed at each iteration. Specifically, we compose the generated images $x_k \sim m_\theta$ with foreground regions $\hat{x}_k \sim n$ at the current noise level, to obtain $x_k$ as input for the next iteration. Finally, we arrive at $I_\text{ref}$ which we use for instance similarity learning.}
   \label{fig:overview_data}
   \vspace{-2mm}
\end{figure*}

\subsection{Learning to track without video data}
\label{sec:data}
In this section, we focus on how to train \modelname in an open-vocabulary context. In particular, open-vocabulary MOT is challenging from a data perspective since we need to localize, classify, and associate possibly unknown objects of extremely diverse appearance \emph{with fixed method components}. Thus, it is of fundamental importance to align our training data distribution $\mathcal{D}^\text{train}$ with the conditions found in evaluation. However, existing video datasets lack the diversity of contemporary image datasets. Hence, it is essential for open-vocabulary trackers to leverage not only video but more importantly static image data during training.

We use the large-scale, diverse image dataset LVIS~\cite{gupta2019lvis} to train \modelname. In particular, for each image $I_\text{key}$ we generate a reference image $I_\text{ref}$. 
Referring to our instance similarity loss in Eq.~\ref{eq:multipos}, the appearance similarity learning is constituted by contrasting positive and negative examples $\mathbf{q}^+$ and $\mathbf{q}^-$. The corollary of this is that learning will be optimal if $\mathbf{q}^+$ consists of examples with distortions commonly encountered in video data, such as change in object scale, viewpoint or lighting, while $\mathbf{q}^-$ contains examples with appearance changes associated with object identity, such as different material.
While distortions like translation, scaling, and rotation can be simulated via classic data augmentation strategies~\cite{zhou2020tracking, zhang2021fairmot, fischer2022qdtrack}, there are certain phenomena like changes in object viewpoint, lighting or context that cannot be simulated by these.

Therefore, to simulate all desired properties of our instance embedding space in $I_\text{ref}$, we combine classic data augmentations with a DDPM-based data hallucination strategy. 
We use the data generation fidelity of stable diffusion~\cite{Rombach_2022_CVPR} to simulate $\mathbf{q}^+$ and $\mathbf{q}^-$ via a specialized denoising process.
Generally, the denoising process of a DPPM can be viewed as the inversion of a forward process that maps an input $x$ to Gaussian white noise $\mathcal{N}(0, \mathbf{I})$.
In the forward direction, Gaussian noise is added to the input image according to a variance schedule $\delta_k$ in $K$ steps.
\begin{equation}
    n(x_k | x_{k-1}) = \mathcal{N}(x_k ; \sqrt{1 - \delta_k} \cdot x_{k-1}, \delta_k \mathbf{I}).
\end{equation}
In the backward direction, a neural network with parameters $\theta$ predicts the parameters $\mu$ and $\Sigma$ of a Gaussian distribution that reverses a forward step.
\begin{equation}
    m_\theta(x_{k-1} | x_k) = \mathcal{N}(x_{k-1} ; \mu_\theta(x_k, k), \Sigma_\theta(x_k, k) ).
\end{equation}

Our specialized denoising process is illustrated in Fig.~\ref{fig:overview_data}.
We initialize $I_\text{ref}$ as $I_\text{key}$ and apply a random geometric transformation to $I_\text{ref}$. 
Next, we use the instance mask annotations in LVIS to define the set of positive examples $A^+$. 
We divide each iteration of the denoising process of image $I_\text{ref}$ using the union of all object masks in $A^+$ into two branches following~\cite{lugmayr2022repaint}. In addition, we use the conditioning mechanism in~\cite{Rombach_2022_CVPR} to guide the backward process with the corresponding image caption.
To initialize the backward process, we set $x_0$ to $I_\text{ref}$ at $\delta_0 = 0.75$ via the forward process $n(x_k | x_{k-1})$. Note that $x$ corresponds to a latent representation of $I_\text{ref}$ obtained via the encoder of stable diffusion.
In each iteration, we apply $m_\theta(x_{k-1} | x_k)$ to obtain a new sample $x_k \sim m_\theta$. At the same time, we use the forward process $n(x_k | x_{k-1})$ on the areas of $A^+$ to generate $\hat{x}_k \sim n$. At the end of each reverse iteration, we compose the two versions via $x_k = A^+ \cdot \hat{x}_k + (1 - A^+) \cdot x_k$. 
We iterate until $\delta_k \leq \eta$. Finally, we apply $m_\theta(x_{k-1} | x_k)$ to the whole image, without branching, as a homogenization step between $A^+ x_k$ and $(1 - A^+)x_k$, while $\eta > \delta_k > 0$. 

By this process, we achieve three goals. First, we generate random perturbations of the background. Second, we keep the areas of $A^+$ close to its original content in each denoising step so that positive instances are integrated well into the new background. Third, we generate distractor objects by caption guided hallucination.

\begin{table*}[t]
\footnotesize
\centering
\caption{\textbf{Open-vocabulary MOT comparison.} We compare our method with existing closed-set trackers and off-the-shelf open-vocabulary baselines on base and novel classes on the validation and test sets of TAO~\cite{dave2020tao}. We indicate the classes and data the methods trained on. Note that methods using TAO data utilize videos for training. All methods use ResNet50~\cite{gao2019res2net} as backbone.}\vspace{-3mm}
\resizebox{0.8\linewidth}{!}{
\begin{tabular}{l|cc|ccc|cccc|cccc}
\toprule
\multicolumn{1}{c|}{Method}  & \multicolumn{2}{c|}{Classes} & \multicolumn{3}{c|}{Data}  & \multicolumn{4}{c|}{Base}   & \multicolumn{4}{c}{Novel}\\ \midrule
\textbf{Validation set} & Base & Novel & CC3M & LVIS & TAO & TETA & LocA & AssocA & ClsA & TETA & LocA & AssocA & ClsA  \\ \midrule

QDTrack~\cite{fischer2022qdtrack} & \checkmark & \checkmark & - & \checkmark & \checkmark & 27.1& 45.6 &24.7 &11.0   & 22.5 &42.7 &24.4 & 0.4 \\
TETer~\cite{li2022tracking} & \checkmark & \checkmark & - & \checkmark & \checkmark & 30.3 &47.4 &31.6 &12.1 & 25.7 &45.9 &31.1 & 0.2  \\ 


DeepSORT (ViLD)~\cite{wojke2017simple} & \checkmark & - &-& \checkmark & \checkmark & 26.9 & 47.1 & 15.8 & 17.7 & 21.1 & 46.4 & 14.7 & \textbf{2.3} \\
Tracktor++ (ViLD)~\cite{bergmann2019tracking} & \checkmark & - &- &\checkmark & \checkmark & 28.3 & 47.4 & 20.5 & 17.0 & 22.7 & 46.7 & 19.3 & 2.2   \\ \midrule
\textbf{\modelname}  & \checkmark & - & - & \checkmark & - & \textbf{35.5} & \textbf{49.3} & \textbf{36.9}   & \textbf{20.2} & \textbf{27.8} & \textbf{48.8}   & \textbf{33.6}   & 1.5 \\ \midrule
\underline{RegionCLIP~\cite{zhong2022regionclip}} & & & & & & & & & & & & &  \\
+ DeepSORT ~\cite{wojke2017simple} & \checkmark & - & \textcolor{red}{\checkmark} & \checkmark & \checkmark & 28.4 &	52.5 &	15.6 &	17.0 &24.5 &	49.2 & 15.3  & 9.0 \\

+ Tracktor++ ~\cite{bergmann2019tracking} & \checkmark & - &\textcolor{red}{\checkmark} &\checkmark &\checkmark& 29.6 & 52.4 & 19.6 & 16.9 & 25.7 & 50.1 & 18.9 & 8.1 \\

+ \textbf{\modelname}   & \checkmark & - &\textcolor{red}{\checkmark}& \checkmark & - & \textbf{36.3} & \textbf{53.9} & \textbf{36.3} & \textbf{18.7} & \textbf{32.0} & \textbf{51.4} & \textbf{33.2} & \textbf{11.4} \\ \midrule \midrule

\textbf{Test set} & Base & Novel & CC3M & LVIS & TAO & TETA & LocA & AssocA & ClsA & TETA & LocA & AssocA & ClsA \\ \midrule

QDTrack~\cite{fischer2022qdtrack} & \checkmark & \checkmark & - & \checkmark & \checkmark &25.8& 43.2 &23.5 &10.6 & 20.2 & 39.7 & 20.9 & 0.2 \\
TETer~\cite{li2022tracking} & \checkmark & \checkmark & - & \checkmark & \checkmark & 29.2 & 44.0 & 30.4 & 10.7 & 21.7 & 39.1 & 25.9 & 0.0 \\ 

DeepSORT (ViLD)~\cite{wojke2017simple} & \checkmark & - &-& \checkmark & \checkmark & 24.5 & 43.8 & 14.6 & 15.2 & 17.2 & 38.4 & 11.6 & 1.7 \\

Tracktor++ (ViLD)~\cite{bergmann2019tracking} & \checkmark & - &- &\checkmark & \checkmark & 26.0 & 44.1 & 19.0 & 14.8 & 18.0 & 39.0 & 13.4 & 1.7 \\ \midrule
\textbf{\modelname}   & \checkmark & - & - & \checkmark & - & \textbf{32.6}& \textbf{45.6} & \textbf{35.4} & \textbf{16.9} & \textbf{24.1} & \textbf{41.8} & \textbf{28.7} & \textbf{1.8} \\ \midrule
\underline{RegionCLIP~\cite{zhong2022regionclip}} & & & & & & & & & & & & &  \\
+ DeepSORT ~\cite{wojke2017simple} & \checkmark & - &\textcolor{red}{\checkmark}& \checkmark & \checkmark & 27.0 & 49.8  & 15.1 & 16.1 & 18.7 & 41.8 & 9.1 & 5.2 \\

+ Tracktor++ ~\cite{bergmann2019tracking} & \checkmark & - &\textcolor{red}{\checkmark} &\checkmark &\checkmark& 28.0 & 49.4 & 18.8 & 15.7 & 20.0 & 42.4 & 12.0 & 5.7 \\

+ \textbf{\modelname} & \checkmark & - &\textcolor{red}{\checkmark} &\checkmark &- & \textbf{34.8} & \textbf{51.1} & \textbf{36.1} & \textbf{17.3} & \textbf{25.7} & \textbf{44.8} & \textbf{26.2} & \textbf{6.1} \\
\bottomrule
\end{tabular}}
\label{tab:openvocab}
\vspace{-2mm}
\end{table*}

\section{Experiments}
\begin{table}[t]
\centering%
\footnotesize
\caption{\textbf{Closed-set MOT Track mAP comparison.} We compare to existing trackers on TAO~\cite{dave2020tao} validation. Competing methods use ResNet101~\cite{he2016deep}, we use ResNet50 as backbone. All methods use Faster R-CNN~\cite{frcnn}. We include results with stronger detectors and additional data in gray. $\dagger$ does not use videos for training.}\vspace{-2mm}
\resizebox{0.95\linewidth}{!}{
\begin{tabular}{l|cc|ccc}
\toprule
Method        & Track mAP50 & Track mAP75 & Track mAP \\ \midrule
SORT-TAO~\cite{dave2020tao} & 13.2       & -          & -        \\
QDTrack~\cite{fischer2022qdtrack}   & 15.9       & 5.0          & 10.6     \\
GTR$\dagger$~\cite{zhou2022global}  & 20.4      & -          & -        \\
TAC~\cite{woo2022tracking}     & 17.7      & 5.80       & 7.30     \\
BIV~\cite{woobridging}  & 19.6      & 7.30       & 13.6    \\ \midrule 
\textbf{OVTrack}$\dagger$ & \textbf{21.2}      & \textbf{10.6}      & \textbf{15.9}    \\ \midrule 

\textcolor{gray}{GTR + CenterNet2}$\dagger$~\cite{zhou2022global}  & \textcolor{gray}{22.5}      & -          & -        \\
\textcolor{gray}{AOA}~\cite{du2021aoa} & \textcolor{gray}{25.8} & - & - \\\bottomrule 
\end{tabular}}
\vspace{-3mm}
\label{comparison:tao}
\end{table}

\subsection{Evaluation metrics}

\parsection{TETA.} The tracking-every-thing accuracy (TETA)~\cite{li2022tracking} is calculated from three independent scores.
First, the localization accuracy (LocA) is calculated by matching all annotated boxes $\alpha$ to the predicted boxes of $\mathcal{T}$ without taking classification into account:
$\mathrm{LocA} = \frac{|\mathrm{TPL}|}{|\mathrm{TPL}| + |\mathrm{FPL}| + |\mathrm{FNL}|}$.
Next, classification accruacy (ClsA) is computed based on all well-localized TPL, comparing the predicted semantic classes to the matched ground-truths:
$\mathrm{ClsA}$~\!$=$~\!$\frac{|\mathrm{TPC}|}{|\mathrm{TPC}| + |\mathrm{FPC}| + |\mathrm{FNC}|}$.
Finally, association accuracy (AssocA) is computed in a similar fashion, comparing the identity of associated ground truths with well-localized predictions: $\mathrm{AssocA} = \frac{1}{|\mathrm{TPL}|}\sum_{b \in \mathrm{TPL}}\frac{|\mathrm{TPA}(b)|}{|\mathrm{TPA}(b)| + |\mathrm{FPA}(b)| + |\mathrm{FNA}(b)|}$.
The TETA score is computed as the arithmetic mean of the three scores.

\parsection{Track mAP.} 
The Track mAP~\cite{yang2019video} is calculated using the 3D IoU between the bounding boxes of a predicted track $\tau$ and an annotated track $\alpha$ by
$\operatorname{IoU}_\text{3D}(\tau, \alpha) = \frac{\sum_{t \in T}\tau_t \cap \alpha_t}{\sum_{t \in T}\tau_t \cup \alpha_t}$.
It is used analogous to 2D bounding box IoU to calculate the popular average precision metric per class as in~\cite{gupta2019lvis}. The Track mAP is the average of the per-class scores across a set of $\operatorname{IoU}_\text{3D}$ thresholds.
\subsection{Implementation details}
We use ResNet50~\cite{he2016deep} with FPN~\cite{lin2017feature}. We filter object candidates $P$ by non-maximum suppression (NMS) with an IoU threshold of $0.7$ and randomly select $|P| = 256$ candidates per image. We set $\lambda = 0.07$ in $\mathcal{L}_\text{text}$.
We use a two-stage training process, first training the detection components following~\cite{gu2021open, du2022learning}, second fine-tuning the model for tracking with loss weights 0.25 for $\mathcal{L}_\text{track}$ and 1.0 for $\mathcal{L}_\text{aux}$ following~\cite{fischer2022qdtrack}.
We train on the LVIS dataset, with one hallucinated counterpart per image in the dataset. While we use the full dataset for the state-of-the-art comparison, we use a subset of 10,000 images for the ablation studies due to resource constraints.
Unless otherwise noted, we use the following data augmentations in training: resizing, random horizontal flipping, color jittering, random affine transformation, and mosaic composition with varying parameters between $I_\text{key}$ and $I_\text{ref}$. We use $\eta = 0.02$ for data generation.
For inference, we select object candidates $P$ by NMS with an IoU threshold of $0.5$. We keep a track memory of 10 frames to re-identify objects after occlusion and set $\beta = 0.5$ and $\gamma = 0.0001$ (see Sec.~\ref{sec:design}).

\begin{table}[t]
\footnotesize
\centering
\caption{\textbf{Closed-set MOT TETA comparison.} We compare to existing trackers on the TAO~\cite{dave2020tao} validation. Benchmark results are taken from~\cite{li2022tracking}. All competing methods use ResNet101~\cite{he2016deep} except AOA~\cite{du2021aoa}, we use ResNet50 as backbone. All methods use Faster R-CNN~\cite{frcnn}. $\dagger$ does not use videos for training.}\vspace{-2mm}

\resizebox{0.72\linewidth}{!}{
\begin{tabular}{l|cccc}
\toprule
Method                            &   TETA   & LocA  & AssocA & ClsA  \\ \midrule 
SORT-TAO~\cite{dave2020tao}       & 24.8 & 48.1 & 14.3  & 12.1 \\
Tracktor~\cite{bergmann2019tracking}                          & 24.2  & 47.4 & 13.0  & 12.1 \\
DeepSORT~\cite{wojke2017simple}                         & 26.0  & 48.4 & 17.5  & 12.1 \\
AOA~\cite{du2021aoa}                               & 25.3  & 23.4 & 30.6  & \textbf{21.9} \\
Tracktor++~\cite{dave2020tao}                        & 28.0  & 49.0 & 22.8  & 12.1 \\
QDTrack~\cite{fischer2022qdtrack}                       & 30.0  & 50.5 & 27.4  & 12.1 \\
TETer~\cite{li2022tracking}                            & 33.3  & \textbf{51.6} & 35.0  & 13.2 \\ \midrule
\textbf{OVTrack}$\dagger$         & \textbf{34.7}  & 49.3 & \textbf{36.7}  & 18.1 \\
\bottomrule
\end{tabular}}
\vspace{-3mm}
\label{comparison:teta}
\end{table}

\subsection{Comparison to state-of-the-art}
\parsection{Open-vocabulary MOT.}
In Tab.~\ref{tab:openvocab}, we show the open-vocabulary MOT evaluation on the TAO validation and test sets, divided into base classes $\mathcal{C}^\text{base}$ and novel classes $\mathcal{C}^\text{novel}$. For details on the setup, please refer to the supplemental material. 
The baselines we establish are composed of both closed-set and open-vocabulary trackers.
We choose the two state-of-the-art closed-set trackers, TETer~\cite{li2022tracking} and QDTrack~\cite{fischer2022qdtrack}, trained on $\mathcal{C}^\text{base} \cup \mathcal{C}^\text{novel}$.
In addition, we combine off-the-shelf trackers DeepSORT~\cite{wojke2017simple} and Tracktor++~\cite{bergmann2019tracking} with the open-vocabulary detector ViLD~\cite{gu2021open} as baseline open-vocabulary trackers. These are, like \modelname, trained on $\mathcal{C}^\text{base}$ only.
Note that all baselines use video data for training, while we use only static images.

Our approach substantially outperforms all closed-set and open-vocabulary baselines. We achieve consistent improvement across LocA, AssocA, and ClsA on both base and novel classes.
The baselines trained on $\mathcal{C}^\text{base} \cup \mathcal{C}^\text{novel}$ can, in some cases, correctly classify objects in $\mathcal{C}^\text{novel}$ but achieve poor results.
On the contrary, both the open-vocabulary baselines and our tracker achieve significantly higher ClsA on novel classes.
However, we note that classification on the TAO dataset remains a very challenging task. The absolute ClsA scores on novel classes are low. This is partially due to the nature of the ClsA metric, which only considers top-1 classification accuracy, while classes on the TAO dataset are diverse and fine-grained. 

Therefore, we investigate the use of stronger, recently proposed open-vocabulary detectors. We combine RegionCLIP~\cite{zhong2022regionclip} with our off-the-shelf baselines and \modelname. We replace the localization and classification parts of \modelname with RegionCLIP while keeping the association fixed.
We observe that ClsA increases substantially for all trackers on novel classes. Our method achieves the best performance by a wide margin and achieves the best ClsA scores with 11.4 and 6.1 on the validation and test sets, respectively. Note however that RegionCLIP makes use of additional data.
 
\parsection{Closed-set MOT.}
In Tab.~\ref{comparison:tao} and Tab.~\ref{comparison:teta} we compare to existing works on the validation split of TAO using Track mAP and TETA metrics, respectively. 
Note that our method neither uses video data for training, nor is it trained on rare classes as defined in Sec.~\ref{sec:benchmark}, while all of the compared closed-set trackers train on video data and use the held-out rare classes for training as they are part of the closed-set evaluation in TAO.
We outperform all previous works by a sizable margin on both metrics.
By examining the TETA scores in Tab.~\ref{comparison:teta}, we observe that our tracker obtains $2.3$ points less in LocA compared to TETer~\cite{li2022tracking}. However, our approach beats TETer in terms of AssocA by $1.7$ points and greatly improves in ClsA by $4.9$ points, illustrating the positive effect of CLIP distillation on both classification and associated compared to closed-set trackers. This validates our design in Sec.~\ref{sec:design}. Note that while AOA~\cite{du2021aoa} has a better ClsA, it ensembles multiple few-shot detection and re-identification models trained on additional datasets as reported by previous works~\cite{zhou2022global, li2022tracking}. Overall, our approach surpasses the previous state-of-the-art by $1.4$ points in TETA and $2.3$ points in Track mAP while using a weaker backbone and the same detector.

\subsection{Ablation studies}
\parsection{CLIP knowledge distillation.}
In Tab.~\ref{tab:ablation_clip} we analyze the effect of the knowledge distillation described in Sec.~\ref{sec:design}.
In particular, we observe that using both $\mathcal{L}_\text{text}$ and $\mathcal{L}_\text{image}$ is more effective for classification than using only $\mathcal{L}_\text{text}$, improving ClsA significantly from 15.6 to 18.1, while LocA and AssocA stay at the same level of performance.

\begin{table}[t]
\footnotesize
\centering
\caption{\textbf{Ablation study on CLIP knowledge distillation.} We show that using both $\mathcal{L}_\text{text}$ and $\mathcal{L}_\text{image}$ is important to classification performance when doing CLIP knowledge distillation (Sec.~\ref{sec:design}).}\vspace{-2mm}
\resizebox{0.75\linewidth}{!}{
\begin{tabular}{cc|cccc}
\toprule
$\mathcal{L}_\text{text}$ & $\mathcal{L}_\text{image}$ & TETA & LocA & AssocA & ClsA   \\ \midrule
 $\checkmark$ & - & 34.0 & \textbf{50.5} & \textbf{35.7} & 15.6  \\ 
 $\checkmark$ & $\checkmark$ & \textbf{34.3} & 49.3 & 35.4 & \textbf{18.1} \\
\bottomrule
\end{tabular}}
\label{tab:ablation_clip}
\vspace{-1mm}
\end{table}

\parsection{Data hallucination strategy.}
We validate the effectiveness of our data hallucination strategy described in Sec.~\ref{sec:data} by training both the closed-set tracker TETer~\cite{li2022tracking} and our \modelname with it. Note that we choose to use SwinT~\cite{liu2021swin} with TETer and ResNet50~\cite{he2016deep} with our \modelname to achieve similar performance, in order to fairly compare the performance difference on both trackers.
Tab.~\ref{tab:ablation_aug} shows the TETA results on the TAO validation set. We observe that our data hallucination strategy improves the AssocA significantly for both trackers, while LocA and ClsA are comparable. In particular, we improve $2.4$ and $1.8$ points in AssocA for TETer and \modelname, respectively. Further, ensembling our data generation strategy with heavy data augmentations yields another $2.5$ and $1.8$ points improvement.
Overall, we show that our data generation strategy improves instance similarity learning across both closed-set and open-vocabulary trackers while being complementary to classic data augmentation.

\begin{table}[t]
\footnotesize
\caption{\textbf{Ablation study on data hallucination strategy.} We show that our data hallucination strategy (`DDPM', Sec.~\ref{sec:data}) improves the association of a closed-set tracker~\cite{li2022tracking} and our \modelname on TAO~\cite{dave2020tao} validation. We ensemble it with data augmentations, where `Standard' refers to random resize and horizontal flip, `Heavy' to color jitter, random affine transformation and mosaic. }\vspace{-2mm}
\resizebox{\linewidth}{!}{
\begin{tabular}{ccc|cccc}
\toprule
Standard & DDPM & Heavy & TETA & LocA & AssocA & ClsA  \\ \midrule
\multicolumn{1}{l}{\textbf{TETer-SwinT}}\\ \hline
\checkmark & - & -             & 32.3  &  50.7   & 30.6    &15.5       \\ 
\checkmark & \checkmark &   -   &33.2    &51.2    &33.0    &15.4    \\
\checkmark & \checkmark & \checkmark    &\textbf{34.3}    &\textbf{51.4}   &\textbf{35.5}    &\textbf{15.8}    \\ \midrule
\multicolumn{1}{l}{\textbf{\modelname}}\\
 \hline
\checkmark & - & -             & 32.5 & 48.9 & 31.1 & 17.6 \\
\checkmark & \checkmark & -      &33.3 & 48.9  & 32.9 & 18.0 \\
\checkmark & \checkmark & \checkmark    &\textbf{34.4} & \textbf{49.1}  & \textbf{35.7} & \textbf{18.3}\\
\bottomrule
    \end{tabular}}
\vspace{-3mm}
\label{tab:ablation_aug}
\end{table}

\section{Conclusion}
This work introduced open-vocabulary MOT as an effective solution to evaluating multi-object trackers beyond pre-defined training categories. We defined a suitable benchmark setting and presented \modelname, a data-efficient open-vocabulary tracker.
By using knowledge distillation from vision-language models, we improve tracking while going beyond limited dataset taxonomies. In addition, we put forth a data hallucination strategy tailored to instance similarity learning that addresses the data availability problem in open-vocabulary MOT.
As a result, \modelname learns tracking from static images and is able to track arbitrary objects in videos while outperforming existing trackers by a sizable margin on the large-scale, large-vocabulary TAO~\cite{dave2020tao} benchmark.

\section{Appendix}

In this supplementary material, we elaborate on our experimental setup, method details, and training and inference hyperparameters. Further, we provide additional ablation studies, dataset statistics, and results of our tracker and of our data hallucination strategy. 

\subsection{Dataset statistics}
Since we focus on tracking an arbitrary vocabulary of classes with our tracker, we use the only large-vocabulary MOT benchmark publicly available, namely TAO~\cite{dave2020tao} in all our experiments. However, to show that our method also works on other datasets, we provide a zero-shot generalization experiment on BDD100K~\cite{bdd100k} in Sec.~\ref{sec:ablations} of this appendix. Furthermore, we show qualitative results on arbitrary internet videos in Sec.~\ref{sec:qualitative}.

\parsection{TAO validation set.}
The 833 object classes in TAO have an overlap of 482 classes in LVIS.
In the validation set of TAO, 295 of the overlapping classes are present. 35 of these classes are defined as rare, which serve as our $\mathcal{C}^\text{novel}$. In total, there are 109,963 annotations across 988 validation sequences for evaluation, with 2,835 annotations in $\mathcal{C}^\text{novel}$.

\parsection{TAO test set.}
To evaluate open-vocabulary MOT on the TAO test set, we resort to the recently published BURST~\cite{athar2022burst} dataset that provides us with test set annotations for the TAO videos. This is due to the fact that the TAO test set annotations are not publicly available. However, we need the test set annotations to split the evaluation into base and novel classes.
In particular, we use the instance mask annotations in BURST to create 2D bounding boxes which serve as our ground truth for evaluation on the TAO test set.

In the test set of TAO, there are 324 of the overlapping classes mentioned above present. 33 of these classes are defined as rare, which serve as our $\mathcal{C}^\text{novel}$. In total, there are 164,501 annotations across 1,419 test sequences for evaluation, with 2,263 annotations in $\mathcal{C}^\text{novel}$.




\subsection{Experiment details}

\parsection{Training details.}
To train \modelname we use a two-stage training scheme. In particular, we first train the detector for 20 epochs on LVIS~\cite{gupta2019lvis} using standard data augmentations and without hallucinated images following~\cite{du2022learning}. We use pre-trained backbone weights from~\cite{wei2021aligning} which are trained self-supervised for 400 epochs on ImageNet~\cite{deng2009imagenet}.
For the first stage of training, we use SGD optimizer with a learning rate of $0.02$, momentum of $0.9$, weight decay of $0.0001$, a batch size of $16$ and decay the learning rate by a factor of 10 at epochs $[8, 16]$.
In the second stage of training, we train the tracking head for 6 epochs on LVIS~\cite{gupta2019lvis} with our hallucinated reference images. We use the same optimizer and learning rate settings and decay the learning rate at epochs $[3, 5]$.

\parsection{Experiment details.}
For the comparison on open-vocabulary MOT, all methods train using the same training schedule and dataset versions. In particular, we use LVISv1 annotations to train our model and the baselines. The baselines, namely QDTrack~\cite{fischer2022qdtrack} and TETer~\cite{li2022tracking} are trained according to the schedules mentioned in the respective papers,~\ie 24 epochs on LVIS and a subsequent fine-tuning of the tracking head on TAO for 12 epochs. We initialize the detection modules following~\cite{du2022learning}. We train our method with a similar, but shorter schedule as described above.
For the closed-set MOT comparisons, we take the same model as above and compare with the numbers reported in the respective papers.
For our ablation studies, we use the same 6 epoch fine-tuning as above.
For data hallucination, we use the combined LVISv1 and COCO annotations as used in ~\cite{dave2020tao, fischer2022qdtrack,zhou2022global}. Note that for data hallucination, we only add objects with a bounding box area greater than $64^2$ to $A^{+}$.


\subsection{Method details}
We provide details of our network architecture, losses, and inference scheme.
For the tracking and image heads, we use a standard \textit{4-conv-1-fc} architecture each. The text embedding and bounding box regression, share a single head with the \textit{4-conv-1-fc} architecture, with two parallel linear layers on top for text embedding and box regression outputs.
In terms of network losses, we attach the formula of $\mathcal{L}_\text{aux}$ described in the main paper in Sec.~4.1.
\begin{equation}
    \mathcal{L}_\text{aux} = \left(\frac{\textbf{q} \cdot \textbf{q}'}{||\textbf{q}|| ||\textbf{q}'||} - e\right)^2,
\end{equation}
where $e = 1$ if the two samples $\mathbf{q},\mathbf{q}' \in Q$ have the same identity and 0 otherwise. Note also that, to better align the text embeddings $\mathbf{t}_c$ with the task at hand, we use learned context vectors following~\cite{du2022learning}. This is because CLIP is trained with image-text pairs that usually contain only a single or a few instances, unlike the potentially crowded scenes encountered in MOT. 

In terms of inference, we provide the formula for the bi-softmax matching that we use for association:
\begin{equation}\small
    \textbf{s}(\tau, r) = \frac{1}{2} \left[\frac{ \text{exp}(\textbf{q}_{r} \cdot \textbf{q}_{\tau})}{\sum_{r' \in P} \text{exp}(\textbf{q}_{r'} \cdot \textbf{q}_{\tau} )} +
    \frac{\text{exp}(\textbf{q}_{r} \cdot \textbf{q}_{\tau})}{ \sum_{\tau' \in \mathcal{T}} \text{exp}(\textbf{q}_{r} \cdot \textbf{q}_{\tau'} )}\right].
\end{equation}
Moreover, we employ a temporal voting scheme among the frame-level object classification results to decide the final video object category in a given test sequence. Due to the different evaluation criteria of TETA and Track mAP, we use slightly different detector post-processing for inference in our experiments. For Track mAP evaluation, we set $|P| = 300$ and use class-specific non-maximum suppression (NMS). For TETA evaluation, we set $|P| = 50$ and use class-agnostic NMS.
Overall, our inference scheme is illustrated in Algorithm~\ref{alg:inference}.

\begin{algorithm}[t]
    \caption{Inference pipeline of \modelname for associating objects across a video sequence.}
    \label{alg:inference}
    \begin{algorithmic}[1]
        \INPUT frame index $t$, object candidates $r \in P$, confidence $p_r$, detection embeddings $\textbf{q}_r$, and track embeddings $\textbf{q}_\tau$ for all $\tau \in \mathcal{T}$.
        \State \texttt{DuplicateRemoval}($P$)
        \For{$r \in P, \tau \in \mathcal{T}$} \LineComment{compute matching scores}
        \State \textbf{f}$(r, \tau) = $ \texttt{similarity}($\mathbf{q}_r, \mathbf{q}_\tau$)
        \EndFor
        \For{$r \in P$} \LineComment{track management}
        \State $c$ = \texttt{max}$\left(\textbf{f}(r)\right)$ \LineComment{match confidence}
        \State $\tau_{\texttt{match}}$ = \texttt{argmax}$\left(\textbf{f}(r)\right)$ \LineComment{matched track ID}
        \If{$c > \beta$ \textbf{and} $p_i >$ $\beta_{\texttt{obj}}$} \LineComment{object match found}
        \State \texttt{updateTrack}$\left(\tau_{\texttt{match}}, r, \textbf{q}_r, t\right)$  \LineComment{update track}
        \ElsIf{$p_r > \gamma$}
        \State \texttt{createTrack}$\left(r, \textbf{q}_r, t\right)$ \LineComment{create new track}
        \EndIf
        \EndFor
    \end{algorithmic}
\end{algorithm}

\parsection{Data generation pipeline.}
As stated in Sec.~4.2 and 5.2 of the main paper, we apply data augmentations in combination with our data hallucination strategy to simulate all perturbations commonly encountered in video data. 
We implement this process stochastically so that the image $I_\text{ref}$ is generated from a random sample of transformations. The set of transformations is composed of random resize, flip, affine transformation, color jitter, mosaic, and data hallucination.

\begin{table}[t]
\centering%
\footnotesize
\caption{\textbf{Open-Vocabulary MOT Track mAP comparison.} We compare to existing trackers on TAO~\cite{dave2020tao} validation and test sets. All methods use ResNet50 as backbone. All methods use Faster R-CNN~\cite{frcnn}. Only our method does not use videos for training.}\vspace{-2mm}
\resizebox{0.95\linewidth}{!}{
\begin{tabular}{l|ccc|ccc}
\toprule
Method &  \multicolumn{3}{c|}{Base Classes}   & \multicolumn{3}{c}{Novel Classes} \\ \midrule
\textbf{Validation set} & mAP50 & mAP75 &  mAP & mAP50 & mAP75 &  mAP \\ \midrule

QDTrack~\cite{fischer2022qdtrack}   & 14.7  & 5.2   & 10.0 & 8.3   & 3.8   & 6.0  \\
TETer~\cite{li2022tracking} &14.1 & 5.1 & 9.6 & 8.5  & 3.9 & 6.2 \\  \midrule
\modelname & \textbf{21.0} & \textbf{10.1} & \textbf{15.6} & \textbf{23.0} & \textbf{14.5} & \textbf{18.8} \\ \midrule \midrule
\textbf{Test set} & mAP50 & mAP75 &  mAP & mAP50 & mAP75 &  mAP \\ \midrule

QDTrack~\cite{fischer2022qdtrack} &  11.6 & 3.3 & 7.5  & 1.6  & 0.4 & 1.0\\
TETer~\cite{li2022tracking} & 11.3 & 3.1 & 7.2  & 1.7  & 0.6 & 1.2 \\ \midrule
\modelname & \textbf{17.9} & \textbf{7.7} & \textbf{12.9} & \textbf{13.2} & \textbf{3.0} & \textbf{8.2} \\

\bottomrule 
\end{tabular}}
\label{tab:comparison_tao}
\end{table}

\subsection{Ablation studies and additional results}
\label{sec:ablations}
\parsection{Open-vocabulary MOT.}
We add an additional comparison to closed-set trackers in the open-vocabulary setting using the official TAO metrics in Tab.~\ref{tab:comparison_tao}. We observe that also on the official Track mAP metrics, our \modelname outperforms existing closed-set trackers by a wide margin.

\begin{figure*}[t]
    \centering
    \footnotesize
    \setlength\tabcolsep{1.0pt}
    \resizebox{1.0\linewidth}{!}{
    \begin{tabular}{cccc}
        \toprule
        
        & Original & w/o lang. prompt & w/o geo. trans. \\ \midrule
        & \raisebox{-0.5\height}{\includegraphics[width=0.34\linewidth]{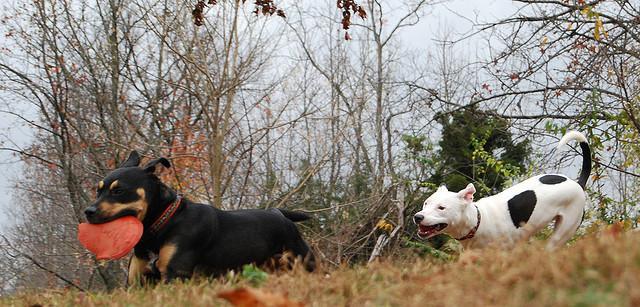}} &
        \raisebox{-0.5\height}{\includegraphics[width=0.34\linewidth]{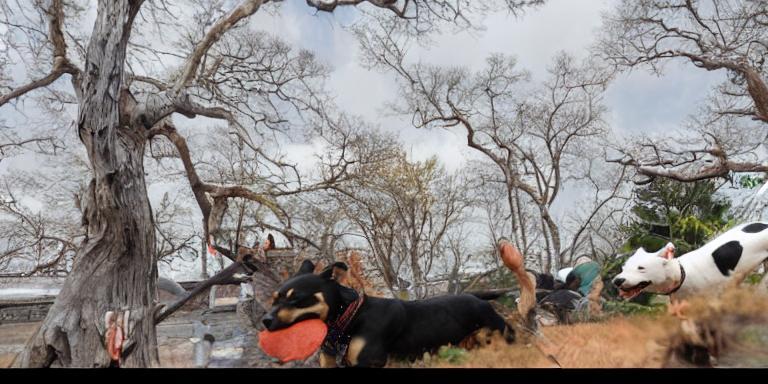}} &
        \raisebox{-0.5\height}{\includegraphics[width=0.34\linewidth]{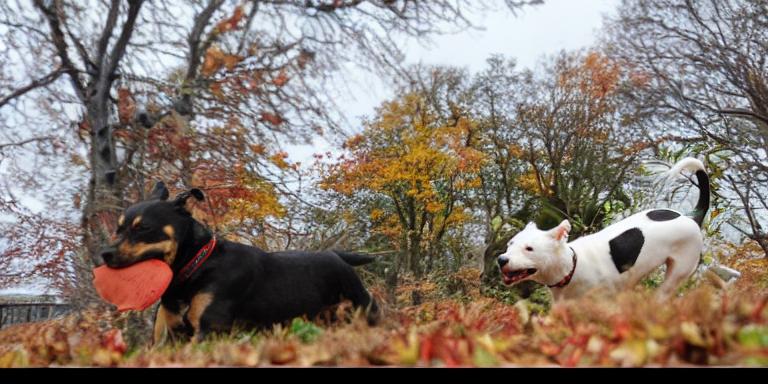}} \\ \midrule \midrule

        & $0.25$ & $\mathbf{0.75}$ & $1.0$\\ \midrule
         \rotatebox[origin=c]{90}{$\delta_0$} & \raisebox{-0.5\height}{\includegraphics[width=0.34\linewidth]{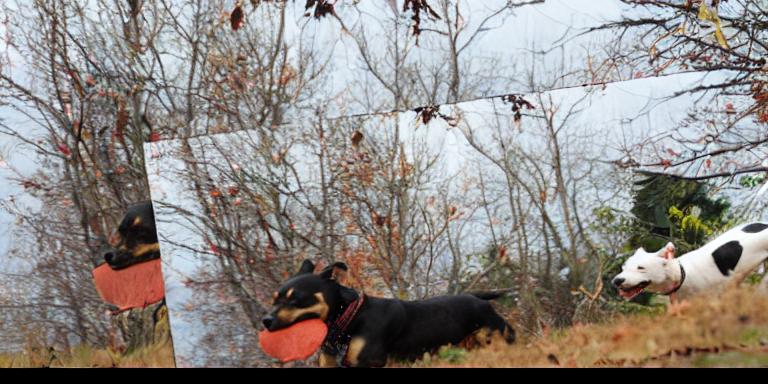}} &
        \raisebox{-0.5\height}{\includegraphics[width=0.34\linewidth]{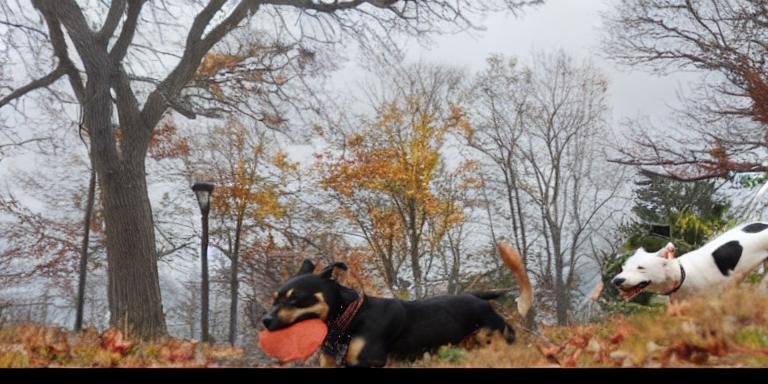}} &
        \raisebox{-0.5\height}{\includegraphics[width=0.34\linewidth]{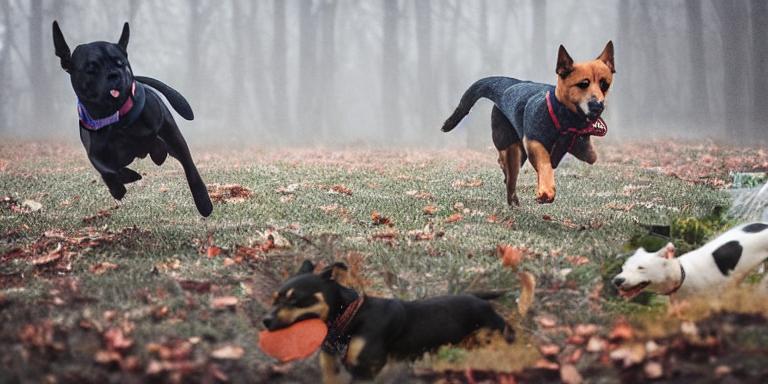}} \\
         \midrule
        & $\mathbf{50}$ & $100$ & $250$\\ \midrule
        \rotatebox[origin=c]{90}{$K$} & \raisebox{-0.5\height}{\includegraphics[width=0.34\linewidth]{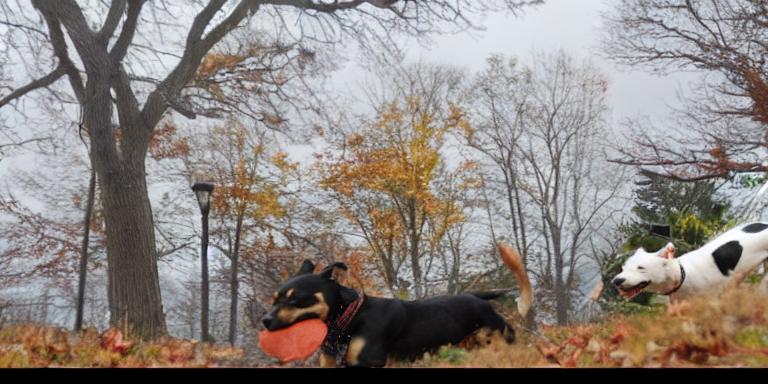}}&
        \raisebox{-0.5\height}{\includegraphics[width=0.34\linewidth]{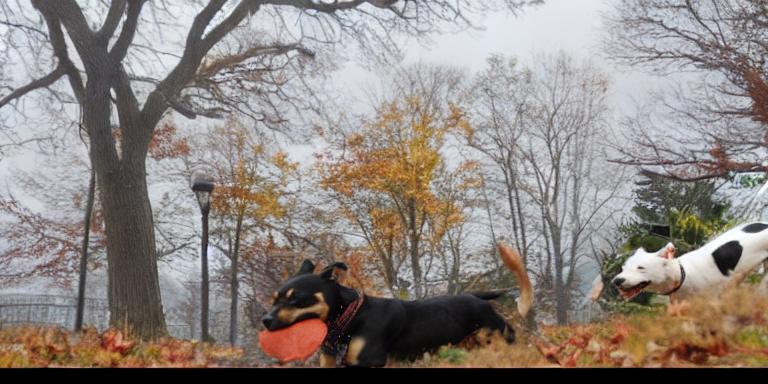}} &
        \raisebox{-0.5\height}{\includegraphics[width=0.34\linewidth]{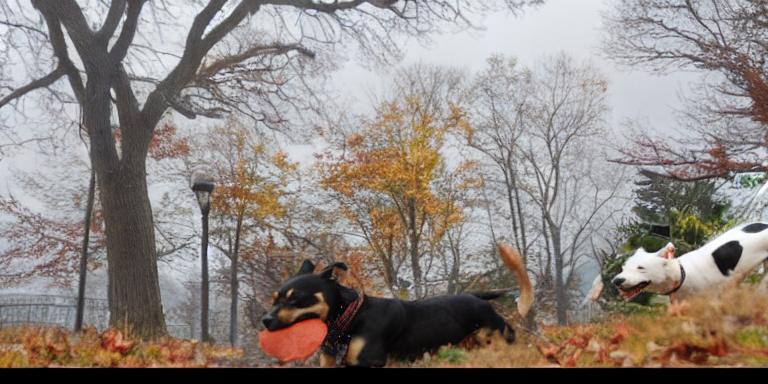}} \\ \midrule
        & $0.0$ & $\mathbf{0.01}$ & $0.1$\\ \midrule
        \rotatebox[origin=c]{90}{$\eta$}& \raisebox{-0.5\height}{\includegraphics[width=0.34\linewidth]{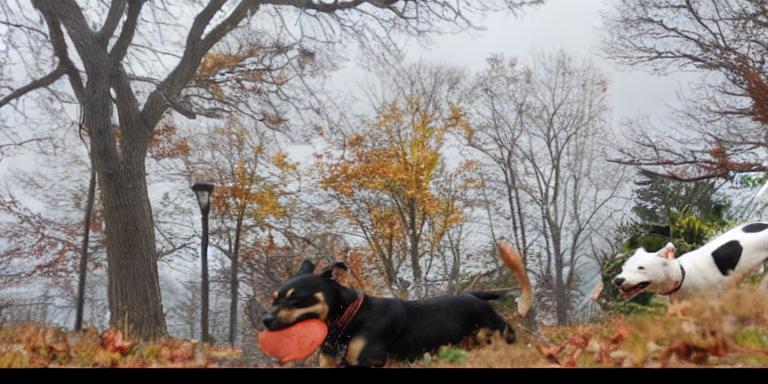}} &
        \raisebox{-0.5\height}{\includegraphics[width=0.34\linewidth]{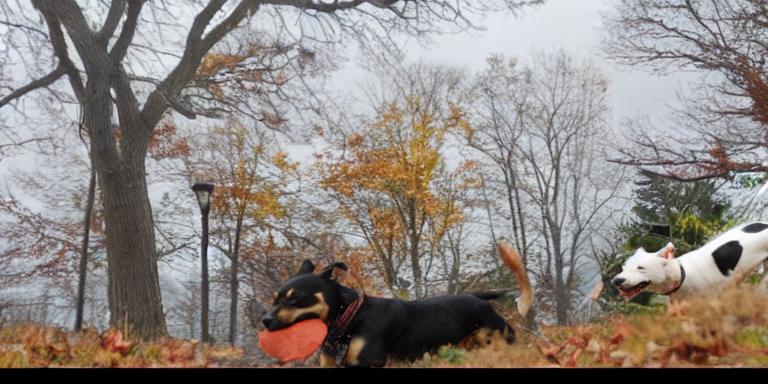}} &
        \raisebox{-0.5\height}{\includegraphics[width=0.34\linewidth]{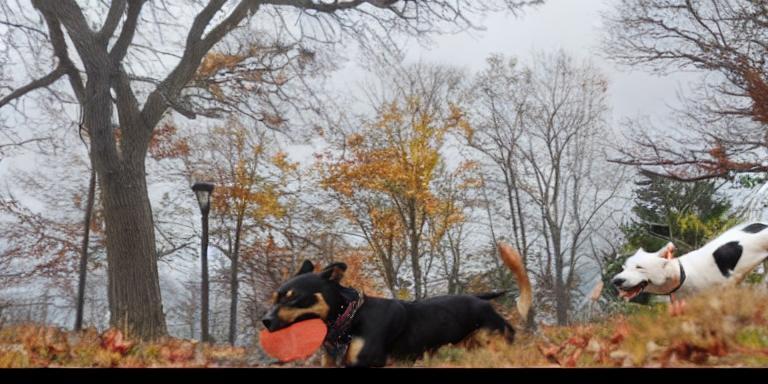}} \\
        \bottomrule
    \end{tabular}}
    \caption{\textbf{Data hallucination hyperparameters.} We show the influence leaving out of language prompt and geometric transformation. In addition, we examine different values for the noise level $\delta_0$, the number of denoising steps $K$ and the homogenization threshold $\eta$. We indicate the value we choose for each of those parameters in bold. 
    }
    \label{fig:hallucination}
\end{figure*}

\parsection{Data hallucination strategy.} 
In Fig.~\ref{fig:hallucination} we illustrate a variety of hyperparameters of the data hallucination process. We experiment with varying noise levels, number of iterations, and homogenization steps and choose the parameter configuration with the visually most appealing results.

In addition, we ablate the most important hyperparameters of our data hallucination strategy quantitatively in Tab.~\ref{tab:ablation_aug}. 
We use standard data augmentations,~\ie random resize and horizontal flip.
We observe that using hallucinated images without language prompt or geometric augmentations fails to improve the performance of the baseline trained without any hallucinated data. When adding the geometric augmentations, however, we see a clear improvement of $1.3$ points in AssocA over the baseline. Further adding the language prompt to condition the hallucination process improves the result by another $0.5$ points in AssocA, culminating in a $1.8$ points improvement. 
\begin{table}[t]
\footnotesize
\caption{\textbf{Data hallucination hyperparameters.} We show that using language prompts and geometric transformation of the input image before denoising is essential to our data hallucination strategy (`DDPM', paper Sec.~4.2). We use the TAO~\cite{dave2020tao} validation set.}
\resizebox{\linewidth}{!}{
\begin{tabular}{ccc|ccccc}
\toprule
DDPM & Lang. prompt & Geo. trans. & TETA & LocA & AssocA & ClsA  \\ \midrule
- & - & - & 32.5 & 48.9 & 31.1 & 17.6 \\
\checkmark & - & - & 32.6 & \textbf{49.0} & 30.6 & 17.2 \\
\checkmark & \checkmark & - & 32.3 & 48.9 & 30.7 & 17.2 \\
\checkmark & - & \checkmark & 32.8 & 48.9 & 32.4 & 17.1 \\
\checkmark & \checkmark & \checkmark & \textbf{33.3} & 48.9  & \textbf{32.9} & \textbf{18.0} \\
\bottomrule
\end{tabular}}
\label{tab:ablation_aug}
\end{table}

\parsection{Zero-shot generalization.} We test the ability of our tracker to adapt zero-shot to another dataset in comparison to closed-set trackers. We use the large-scale MOT benchmark BDD100K~\cite{bdd100k} for this experiment. Note that BDD100K has an overlapping class taxonomy with TAO. We apply our tracker conditioned on text prompts containing the class names in the BDD100K dataset. 
Further, for the closed-set baselines, we provide results where we masked out the logits of classes not present in BDD100K. 

Tab.~\ref{tab:zero_shot} shows the results using the TETA metric. Our tracker exhibits a much better transfer ability, outperforming the closed-set baselines by at least $6.4$ points in TETA. Our \modelname improves over the baselines in localization, association and particularly in classification, where the gap is the biggest with $6.3$ points in ClsA. Overall, we show that we are able to bridge the gap to the upper bound,~\ie a tracker trained on the target dataset.

\begin{table}[t]
\footnotesize
\caption{\textbf{Zero-shot generalization.} We test our model along with two closed set baselines, QDTrack~\cite{fischer2022qdtrack} and TETer~\cite{li2022tracking}, on the BDD100K~\cite{bdd100k} MOT validation split. We indicate the training data used to train each model. $\dagger$ denotes logit masking of classes not present in BDD100K.}
\begin{tabular}{l|c|cccc}
\toprule
Method & Training & TETA & LocA& AssocA & ClsA \\ \midrule
QDTrack$\dagger$ & LVIS, TAO & 35.6                     & 38.1                     & 28.5                       & 40.2                     \\
TETer$\dagger$ & LVIS, TAO   & 36.1                     & 36.4                     & 31.9                       & 40.2                     \\                     
QDTrack & LVIS, TAO          & 32.0                     & 25.9                     & 27.8                       & 42.4                     \\
TETer & LVIS, TAO            & 33.2                     & 24.5                     & 31.8                       & 43.4                     \\
Ours & LVIS & \textbf{42.5}                    & \textbf{41.0}                     & \textbf{36.7}                       & \textbf{49.7}                     \\ \midrule
\textcolor{gray}{TETer} & \textcolor{gray}{BDD100K}  & \textcolor{gray}{58.7}                     & \textcolor{gray}{47.2}                     & \textcolor{gray}{52.9}                       & \textcolor{gray}{76.0}                    \\
\bottomrule
\end{tabular}
\label{tab:zero_shot}
\end{table}

 \begin{figure*}[t]
    \centering
    \small
    \setlength\tabcolsep{0.5mm}
    \resizebox{1.0\linewidth}{!}{
    \begin{tabular}{ccccc}
        \toprule
        $t$ & $t +1$ & $t + 2$& $t + 3$& $t + 4$\\ \midrule
        \includegraphics[width=0.2\linewidth]{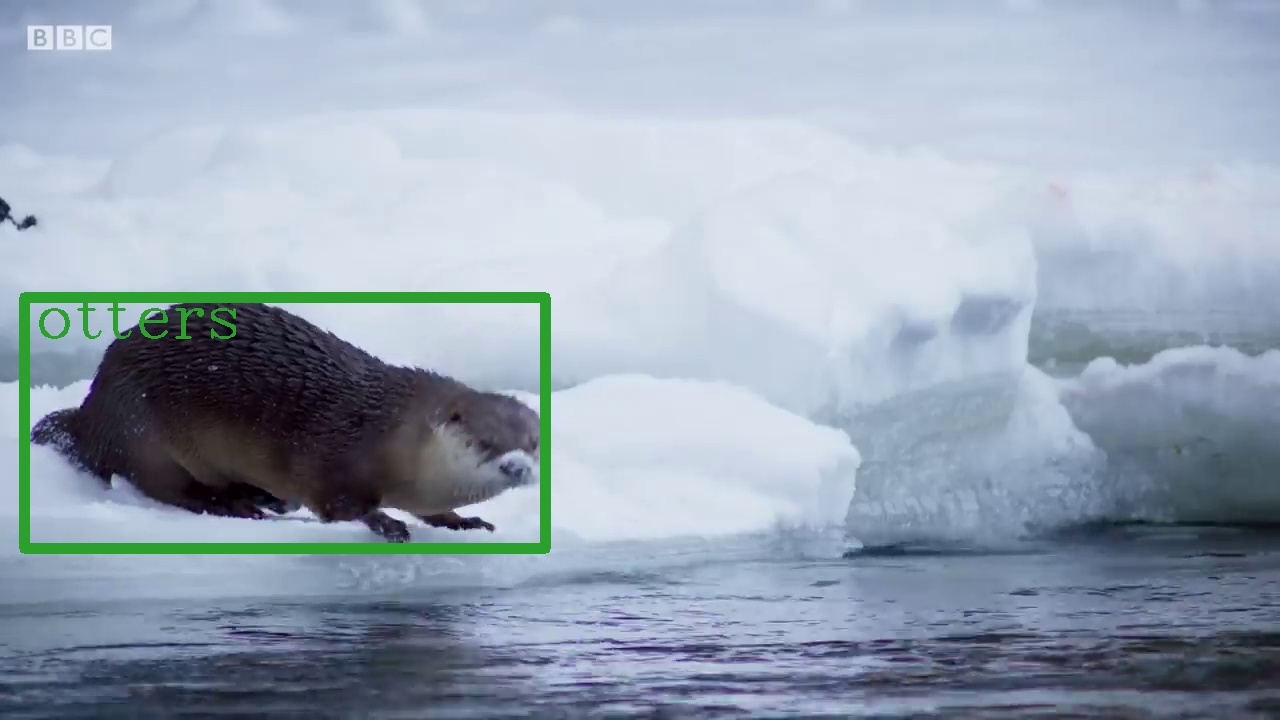} &
        \includegraphics[width=0.2\linewidth]{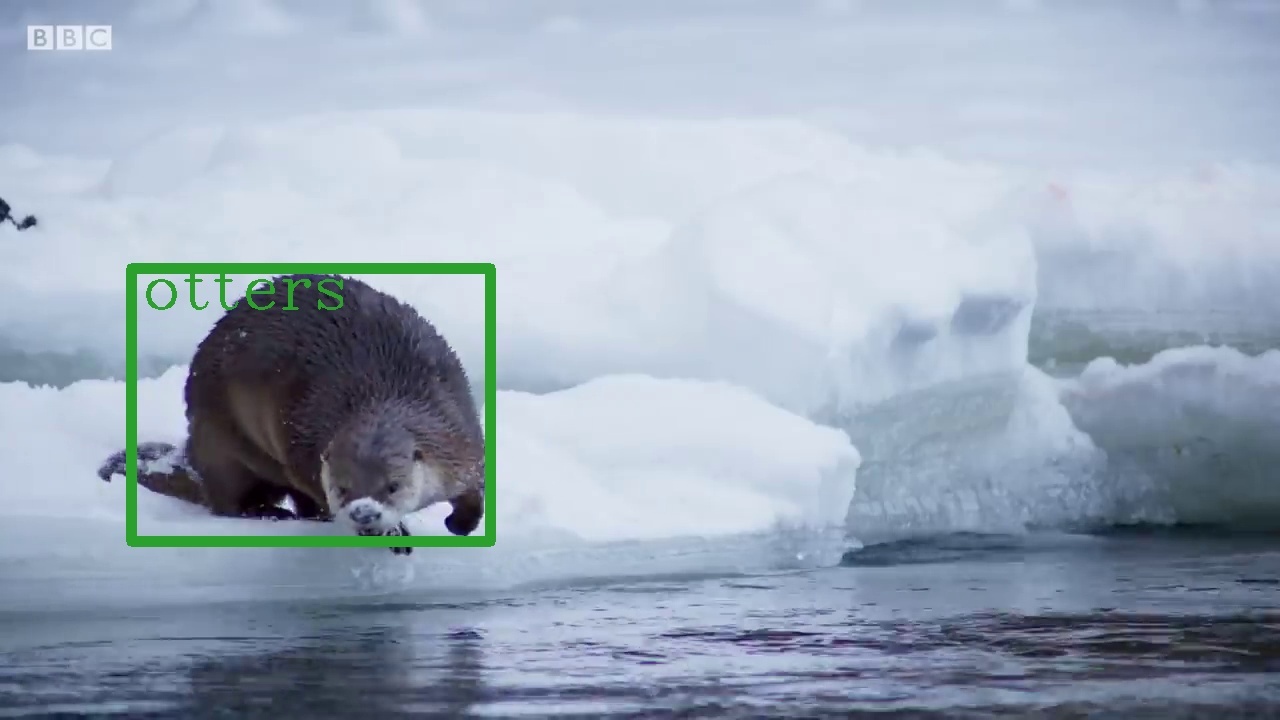} &
        \includegraphics[width=0.2\linewidth]{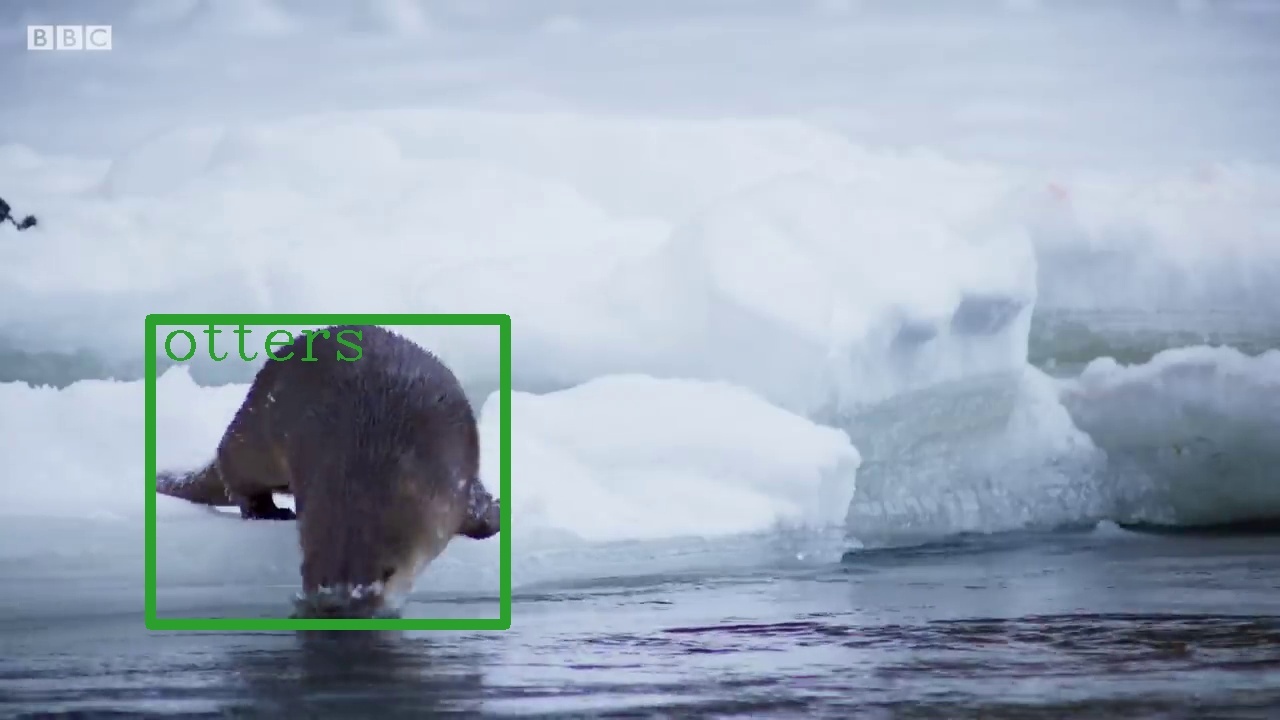} &
        \includegraphics[width=0.2\linewidth]{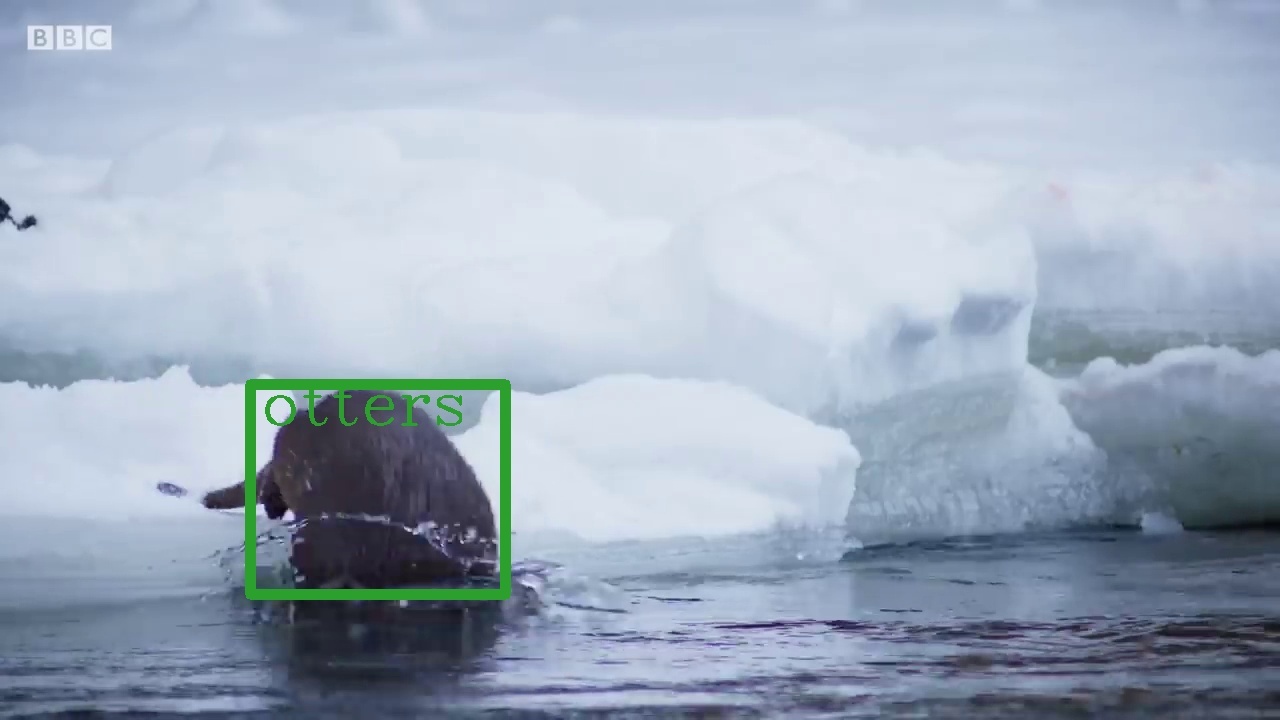} &
        \includegraphics[width=0.2\linewidth]{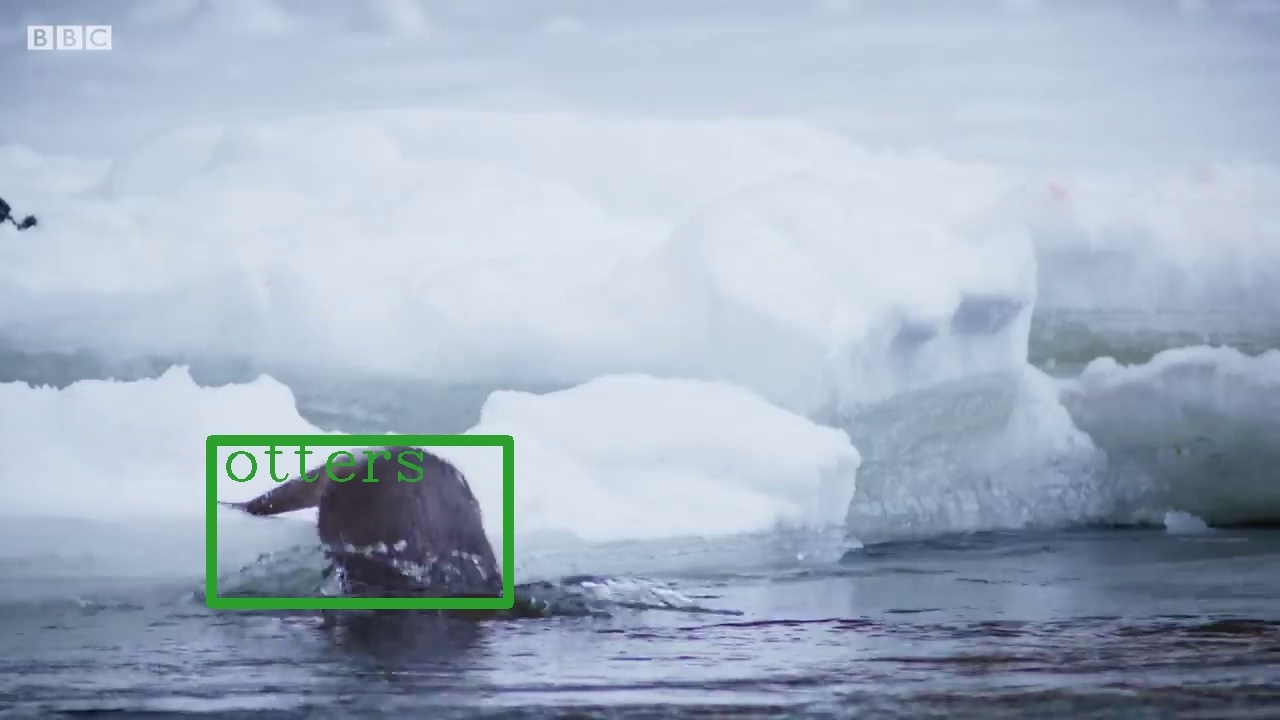} \\

        \includegraphics[width=0.2\linewidth]{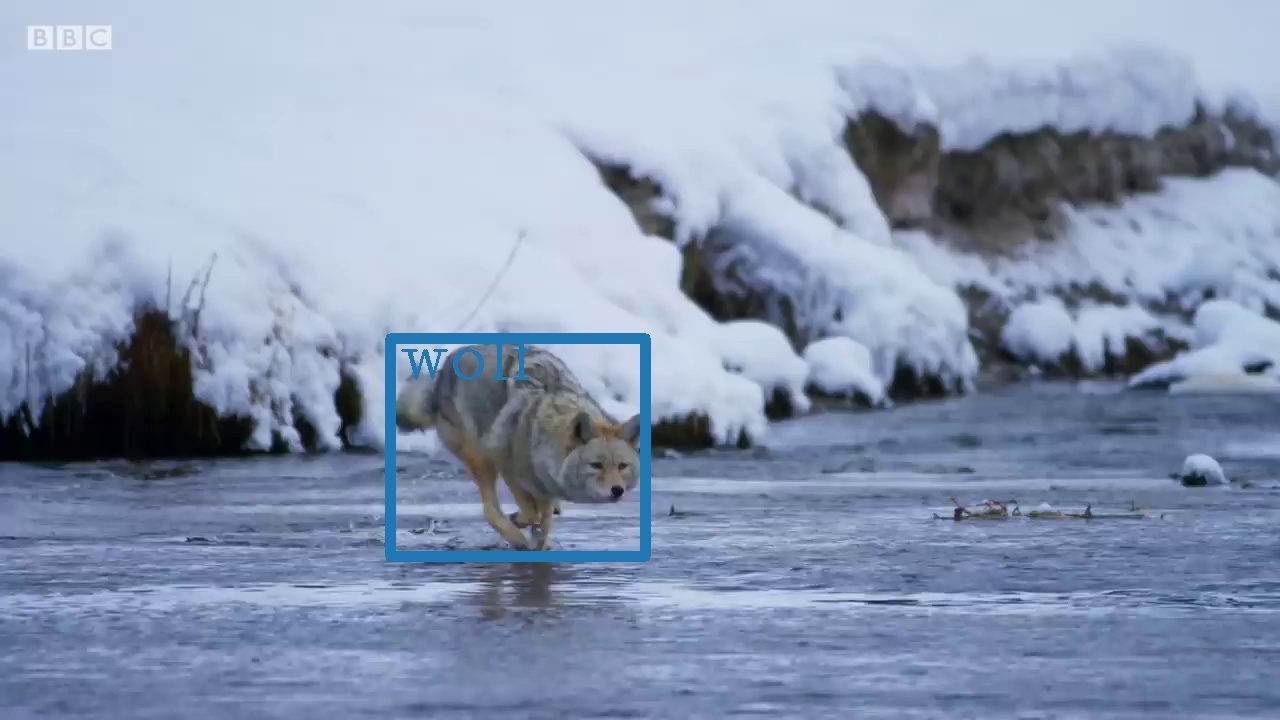} &
        \includegraphics[width=0.2\linewidth]{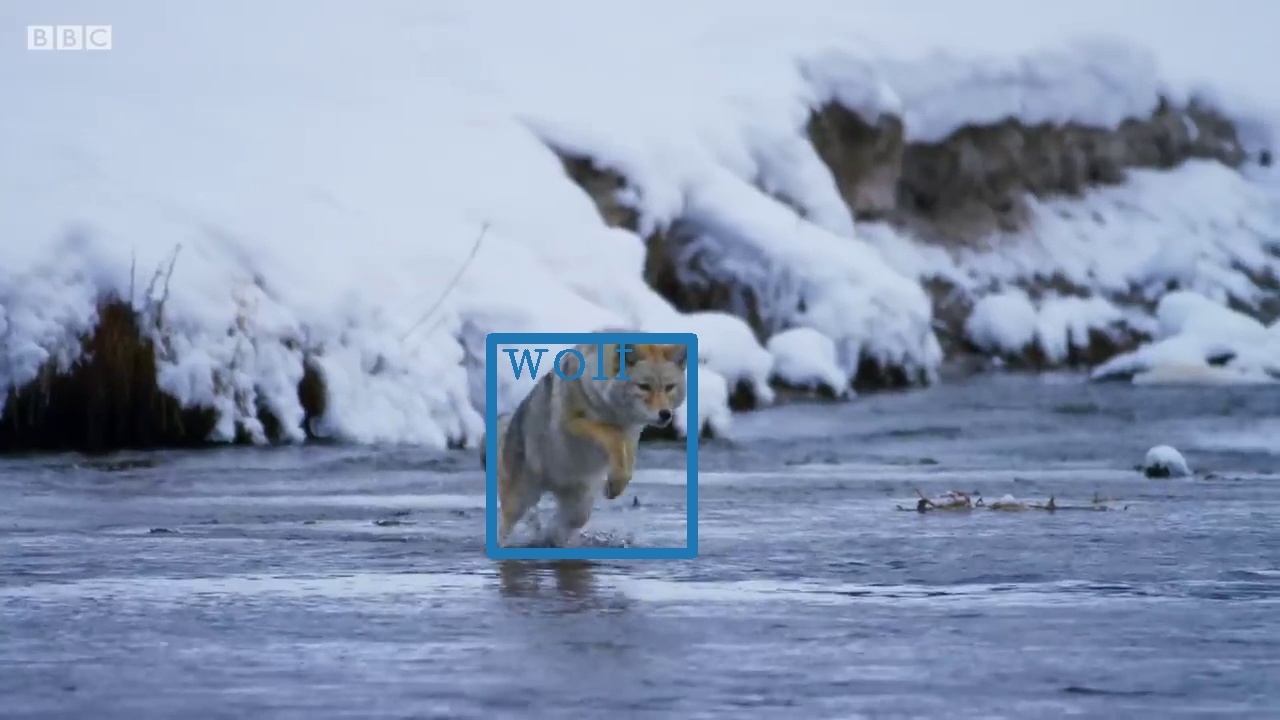} &
        \includegraphics[width=0.2\linewidth]{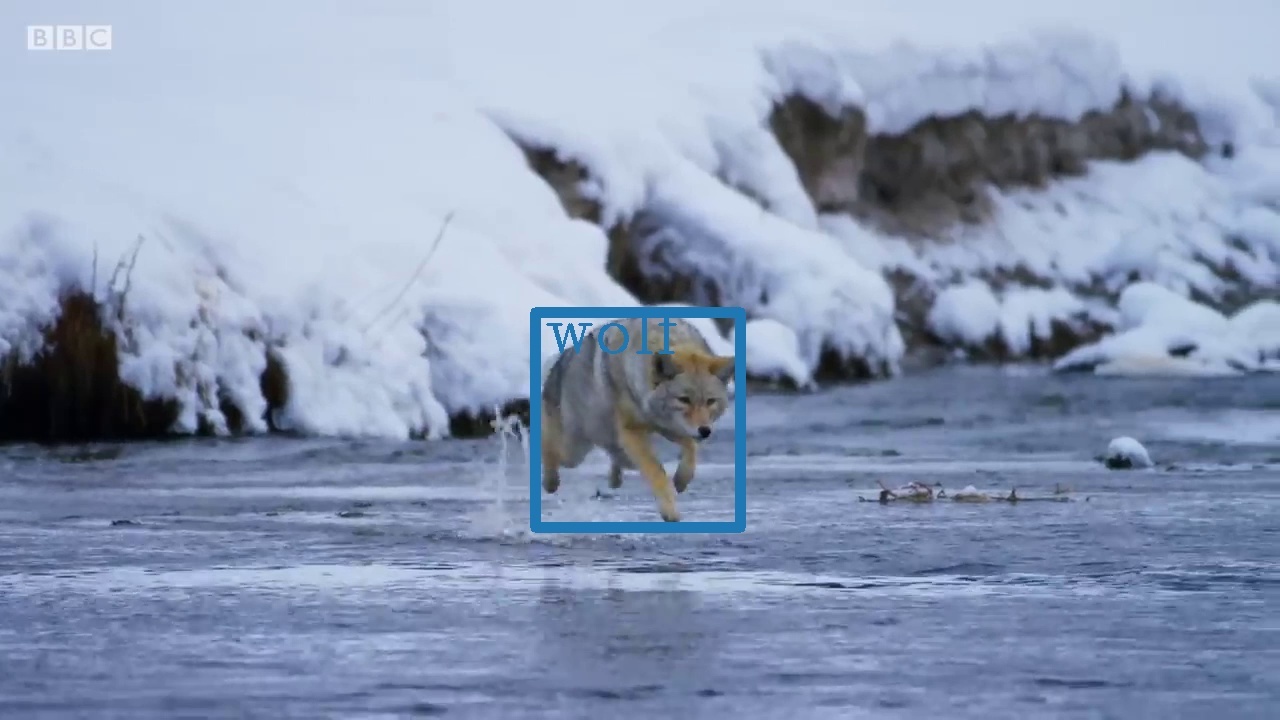} &
        \includegraphics[width=0.2\linewidth]{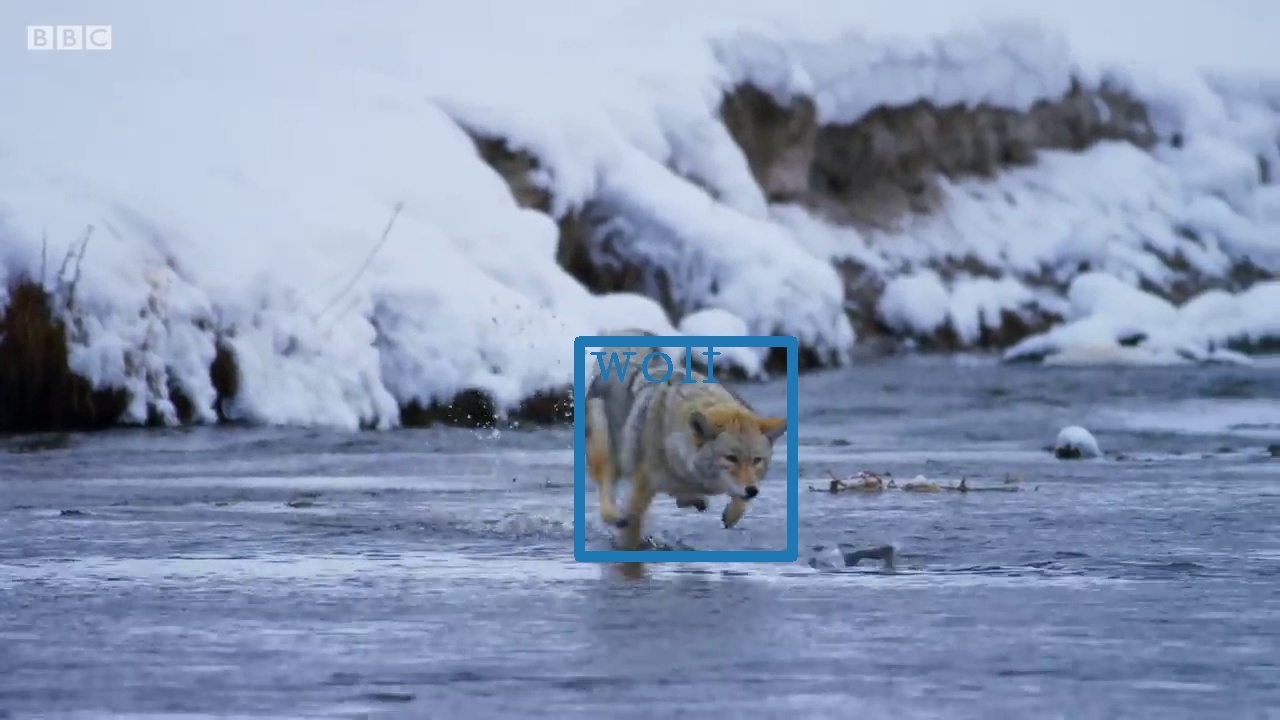} &
        \includegraphics[width=0.2\linewidth]{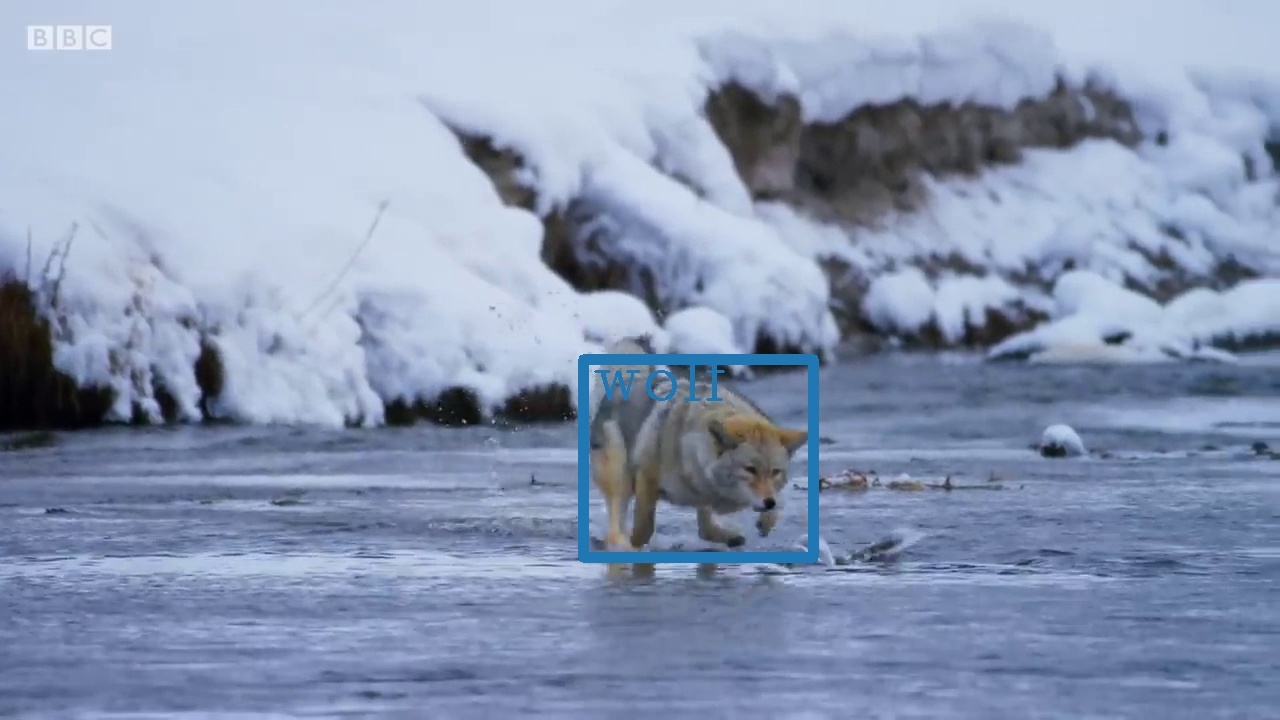} \\
        
        \includegraphics[width=0.2\linewidth]{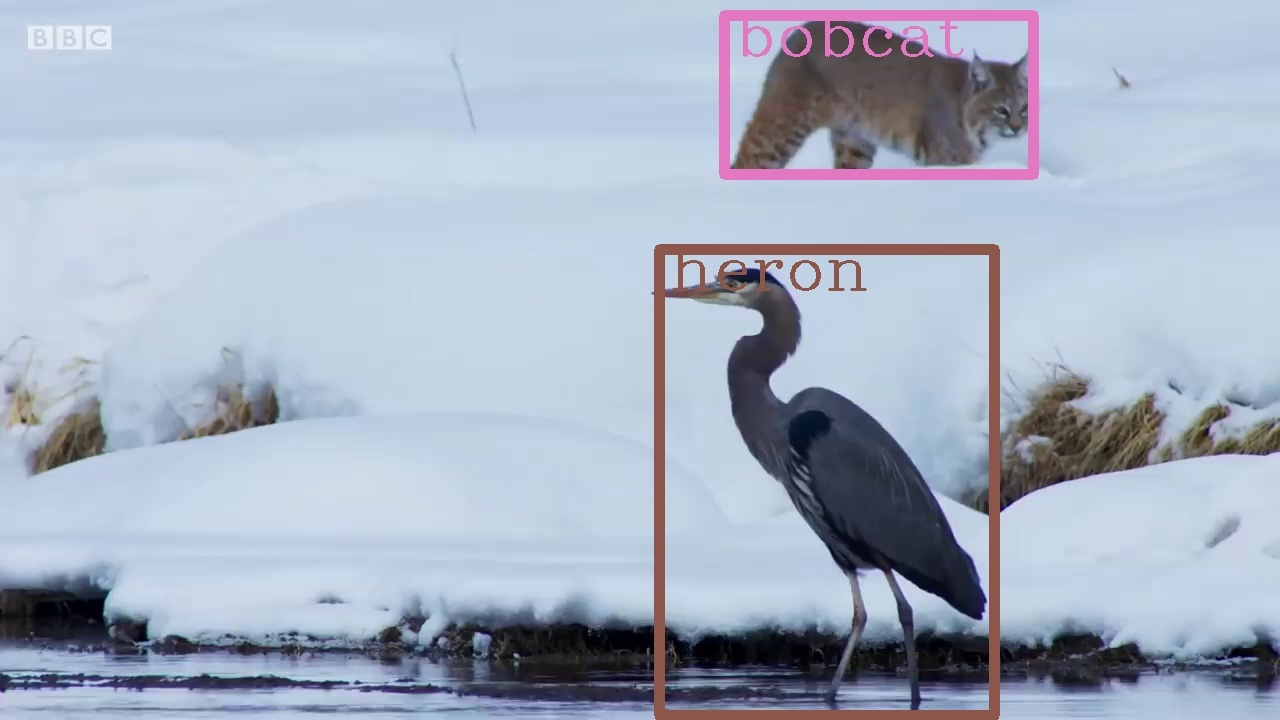} &
        \includegraphics[width=0.2\linewidth]{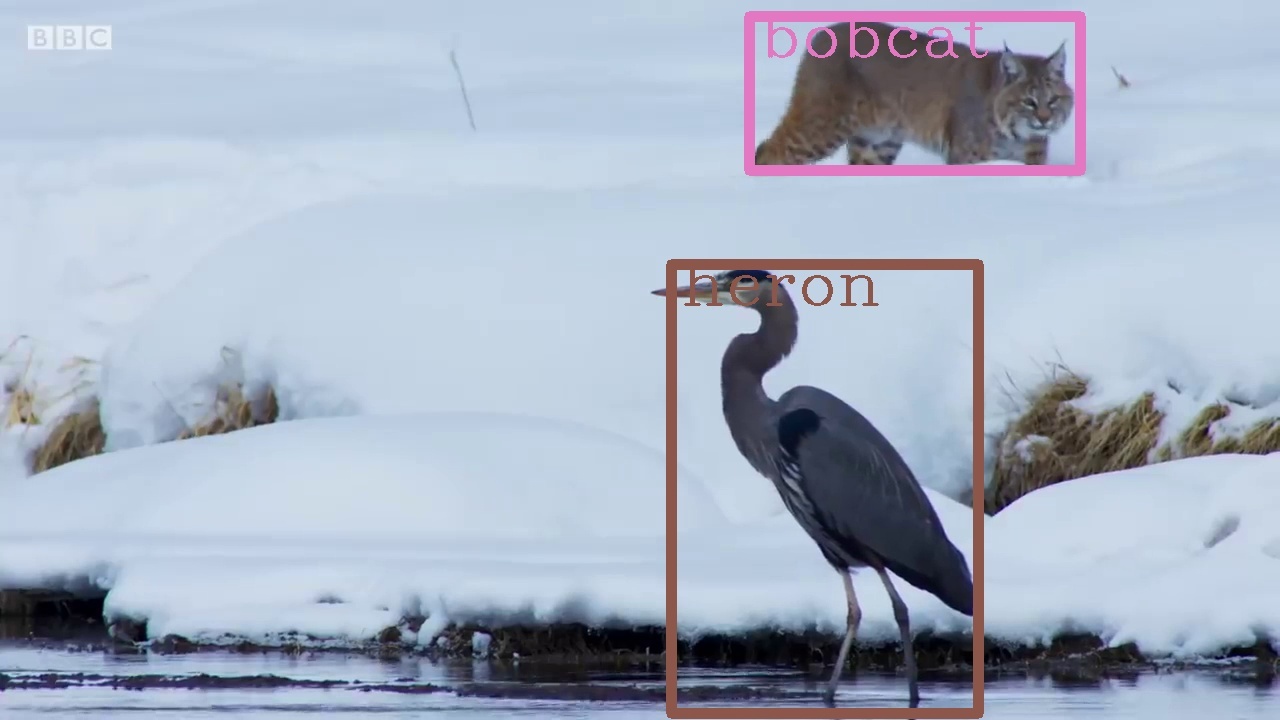} &
        \includegraphics[width=0.2\linewidth]{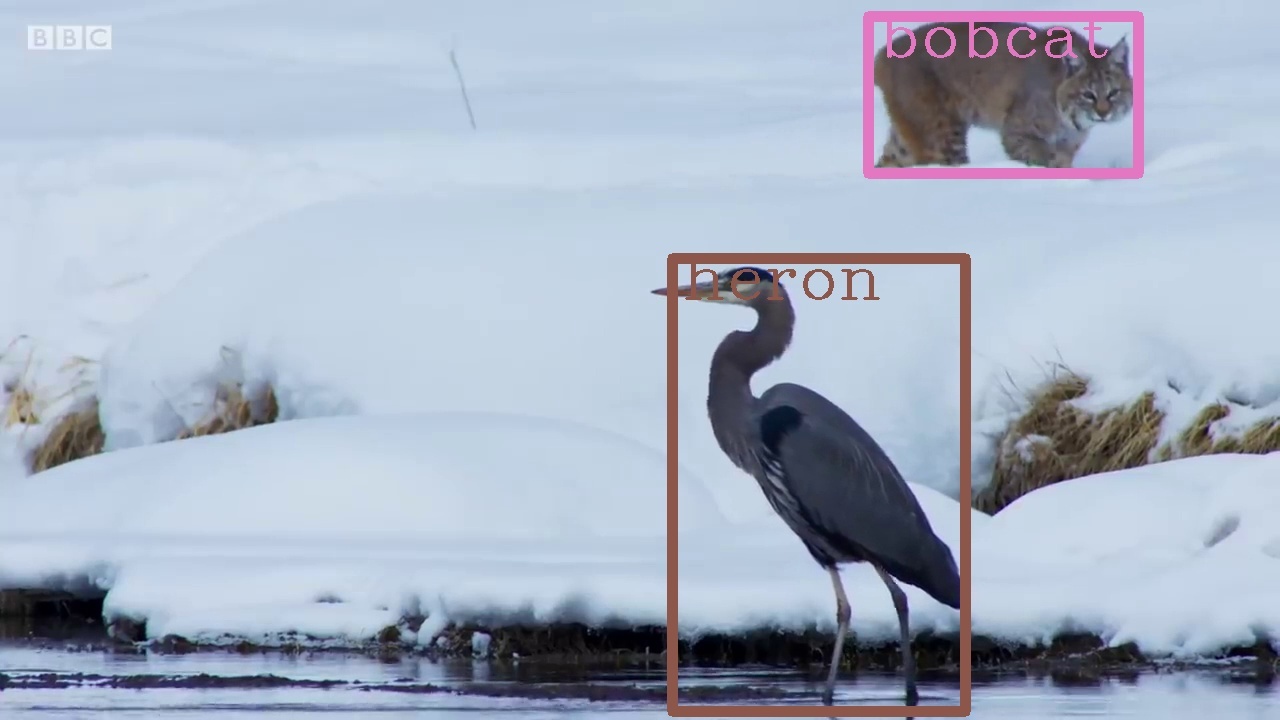} &
        \includegraphics[width=0.2\linewidth]{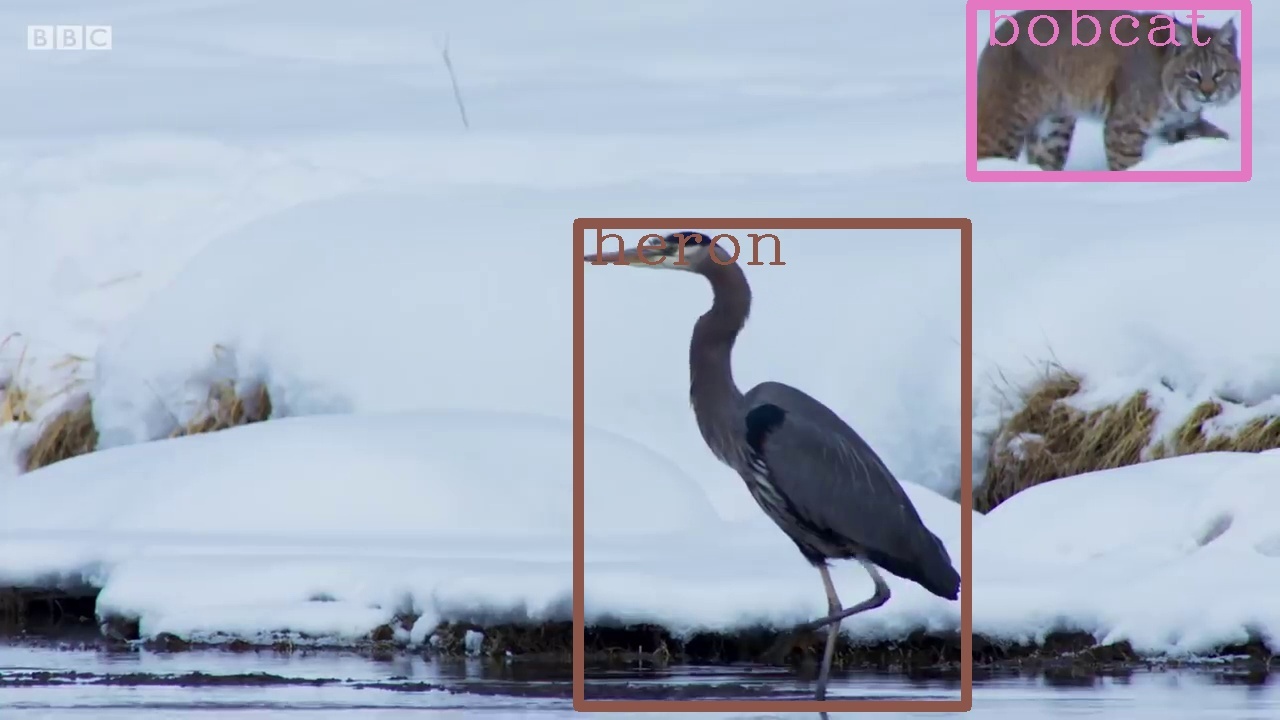} &
        \includegraphics[width=0.2\linewidth]{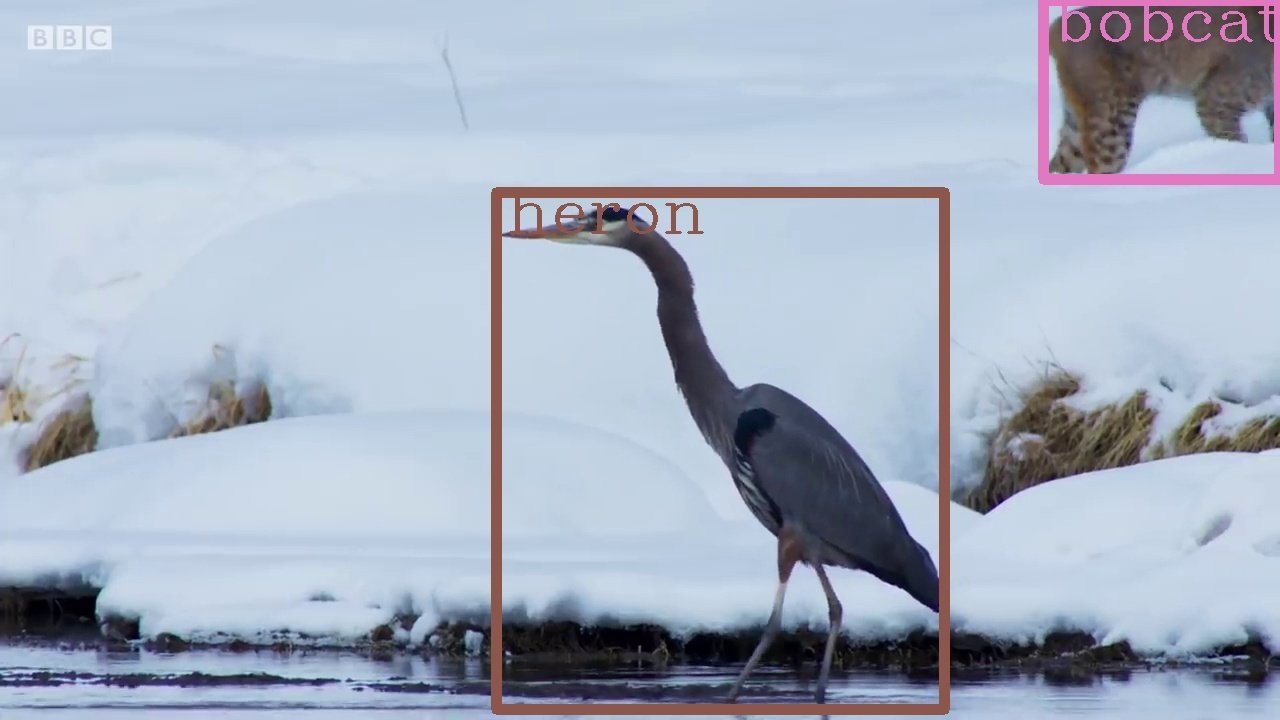} \\
        
        \includegraphics[width=0.2\linewidth]{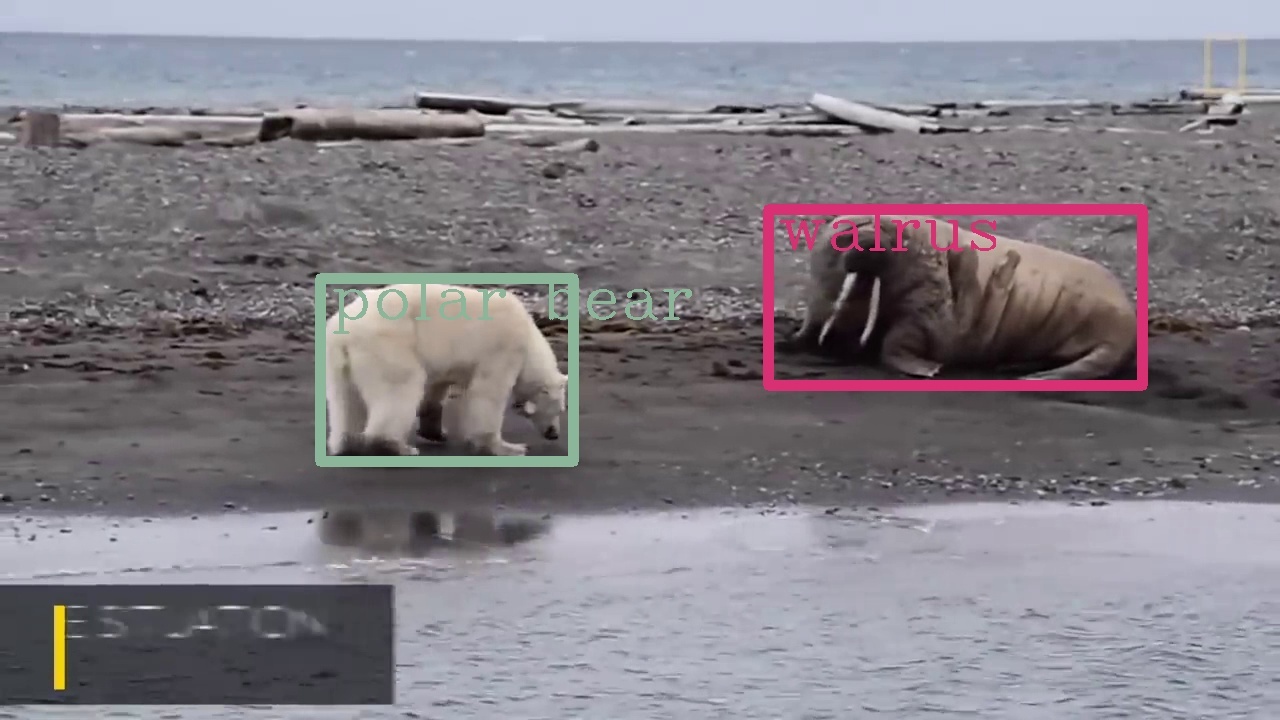} &
        \includegraphics[width=0.2\linewidth]{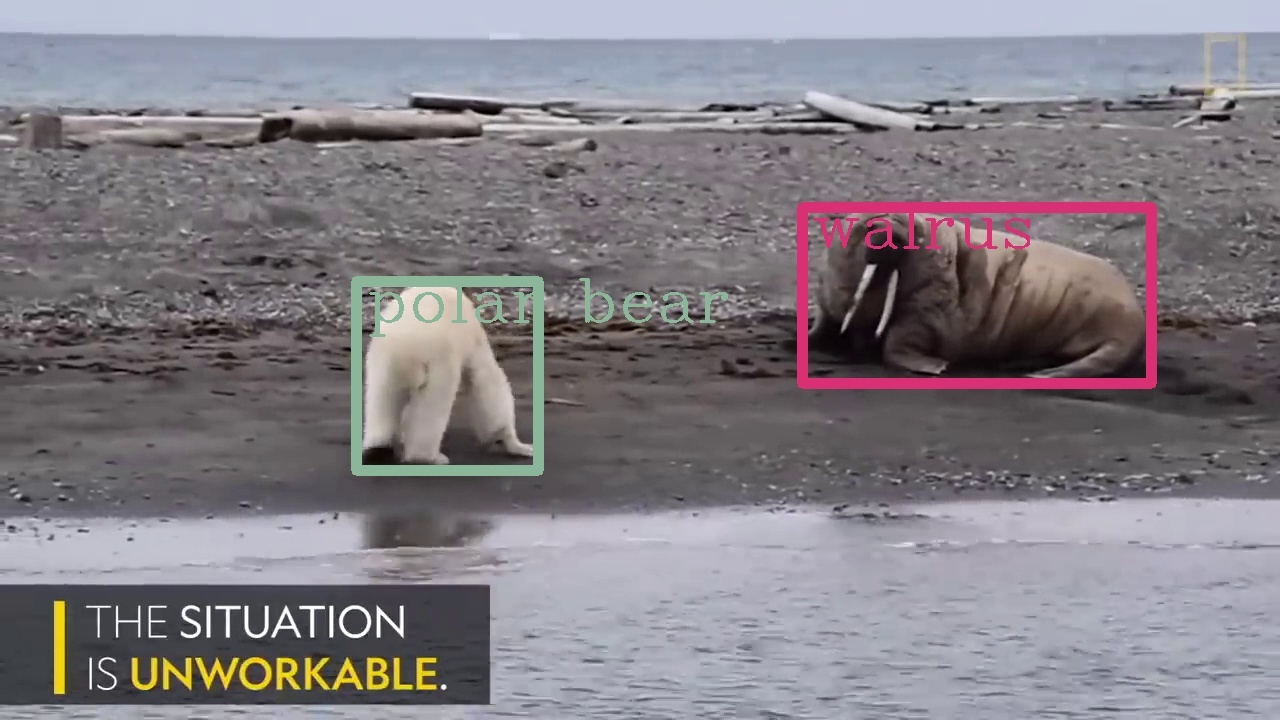} &
        \includegraphics[width=0.2\linewidth]{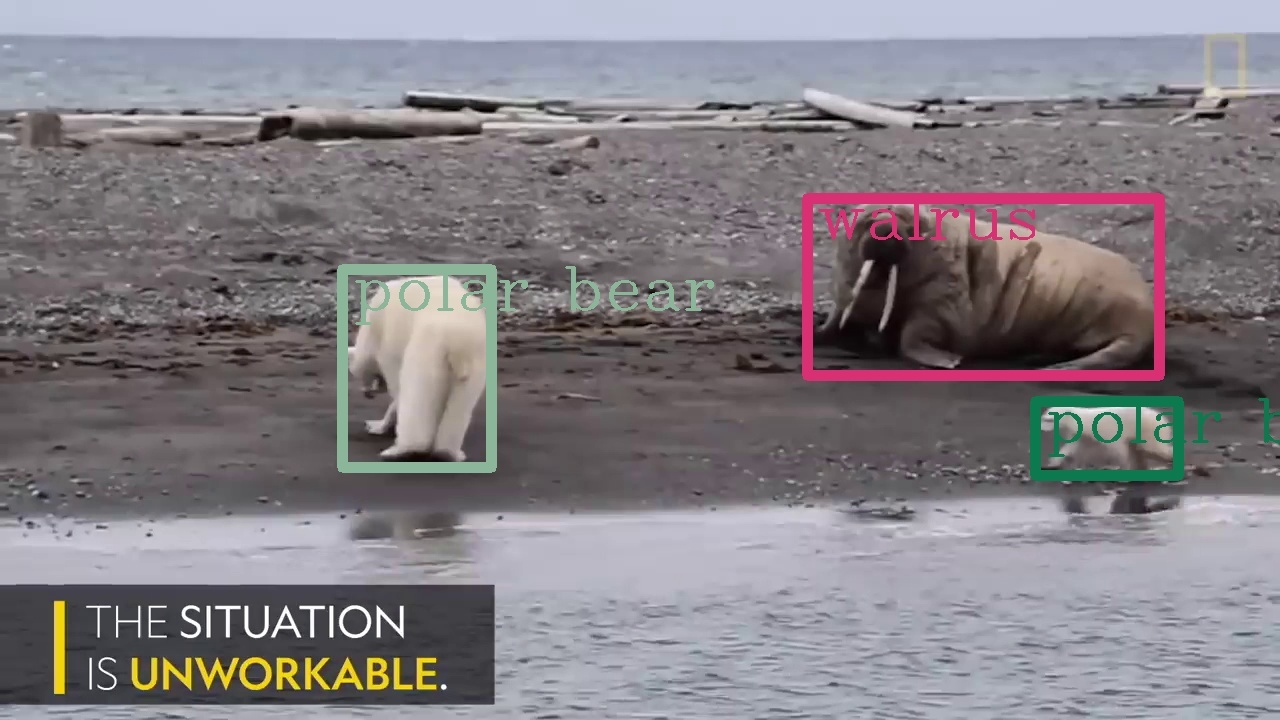} &
        \includegraphics[width=0.2\linewidth]{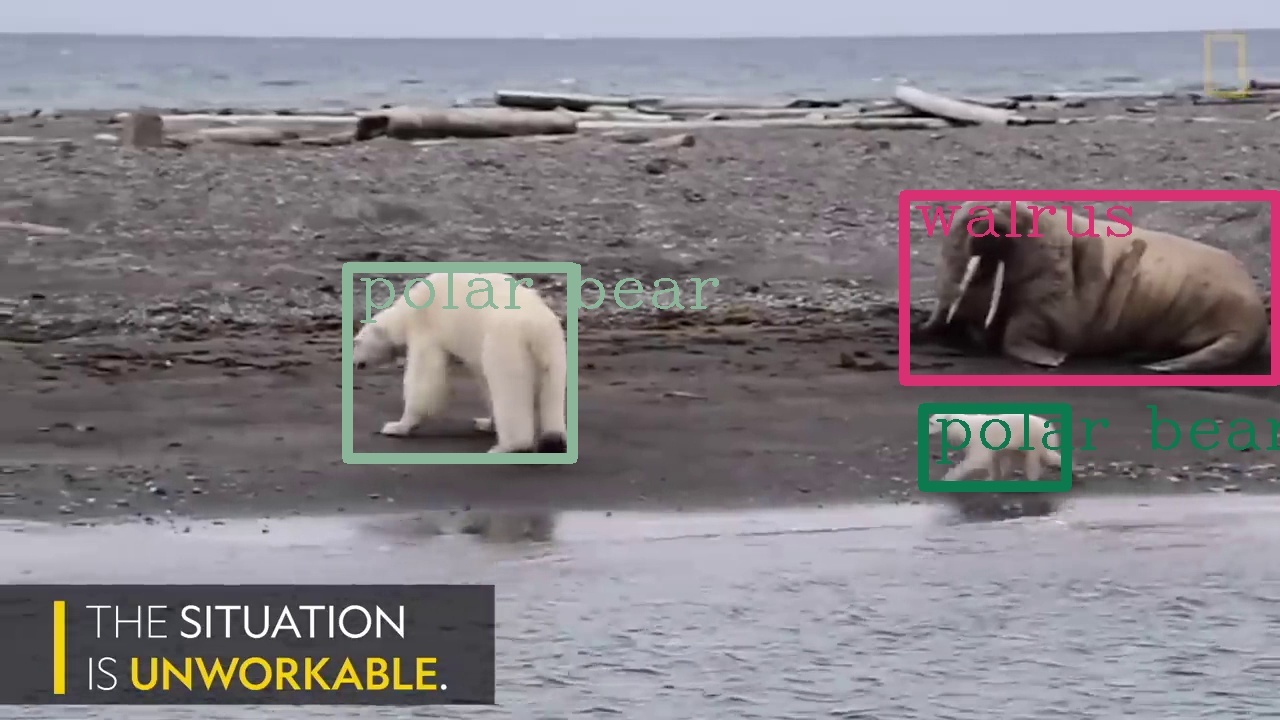} &
        \includegraphics[width=0.2\linewidth]{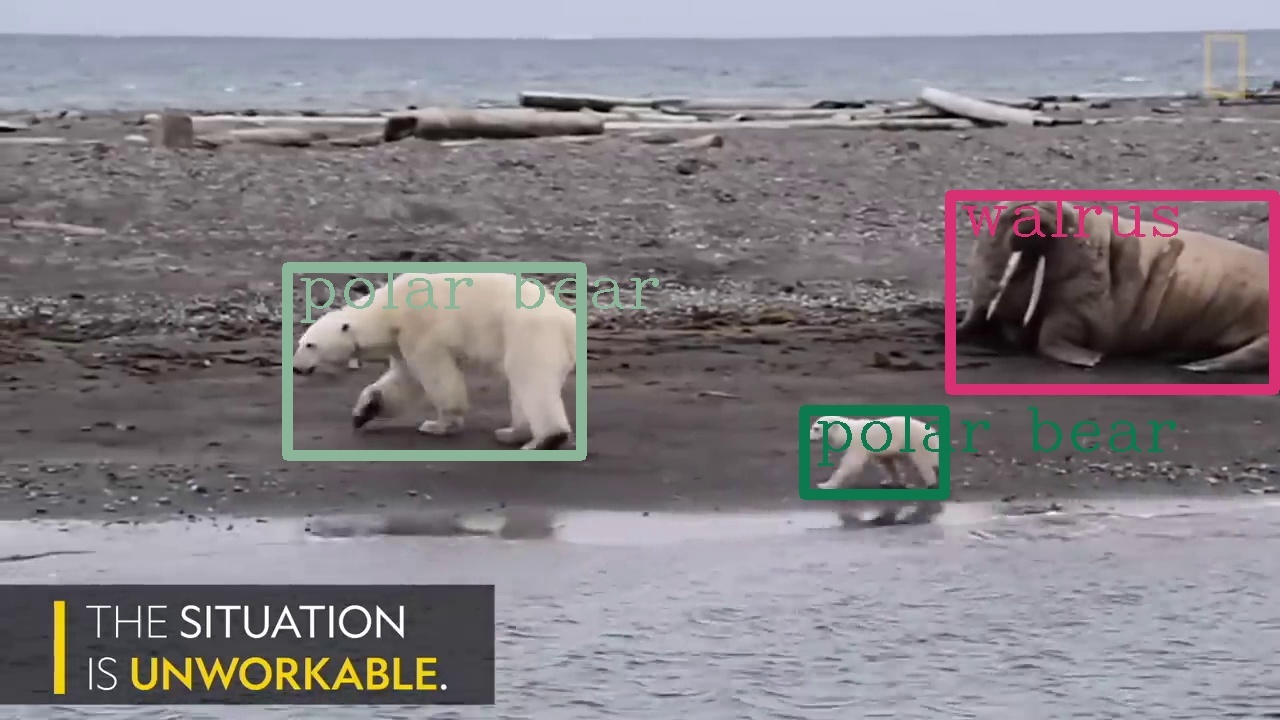} \\
        
        \includegraphics[width=0.2\linewidth]{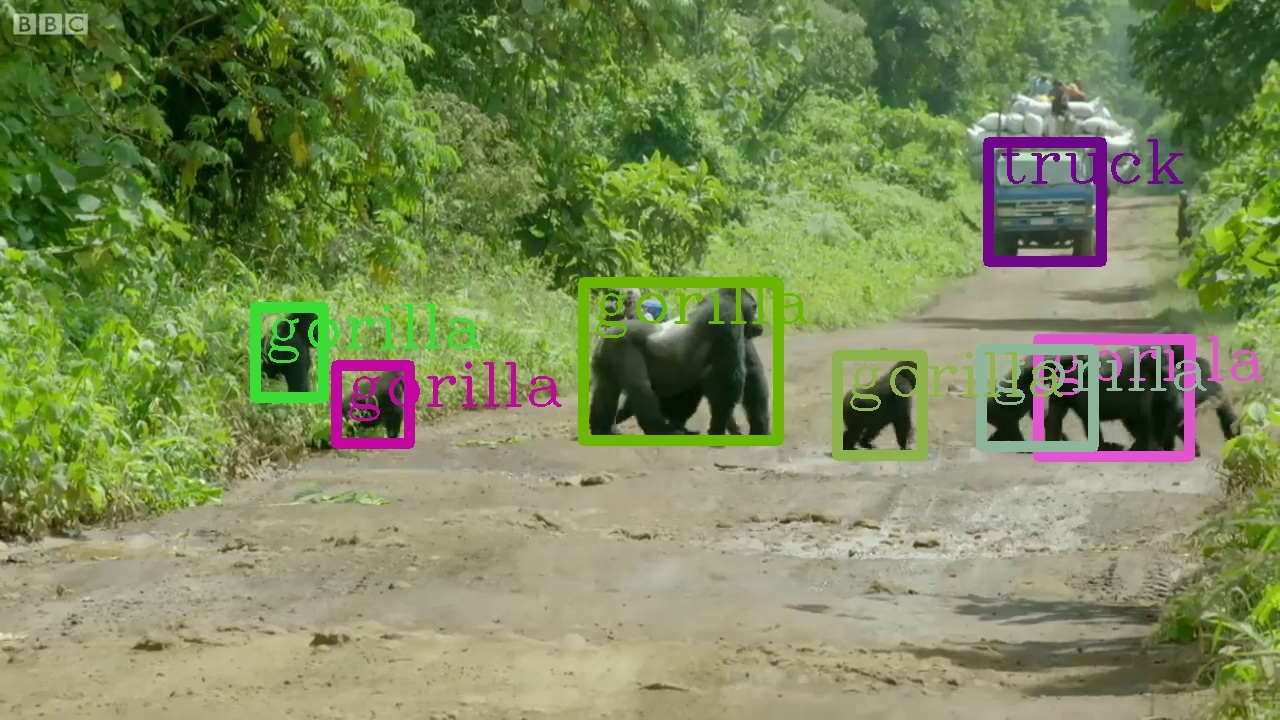} &
        \includegraphics[width=0.2\linewidth]{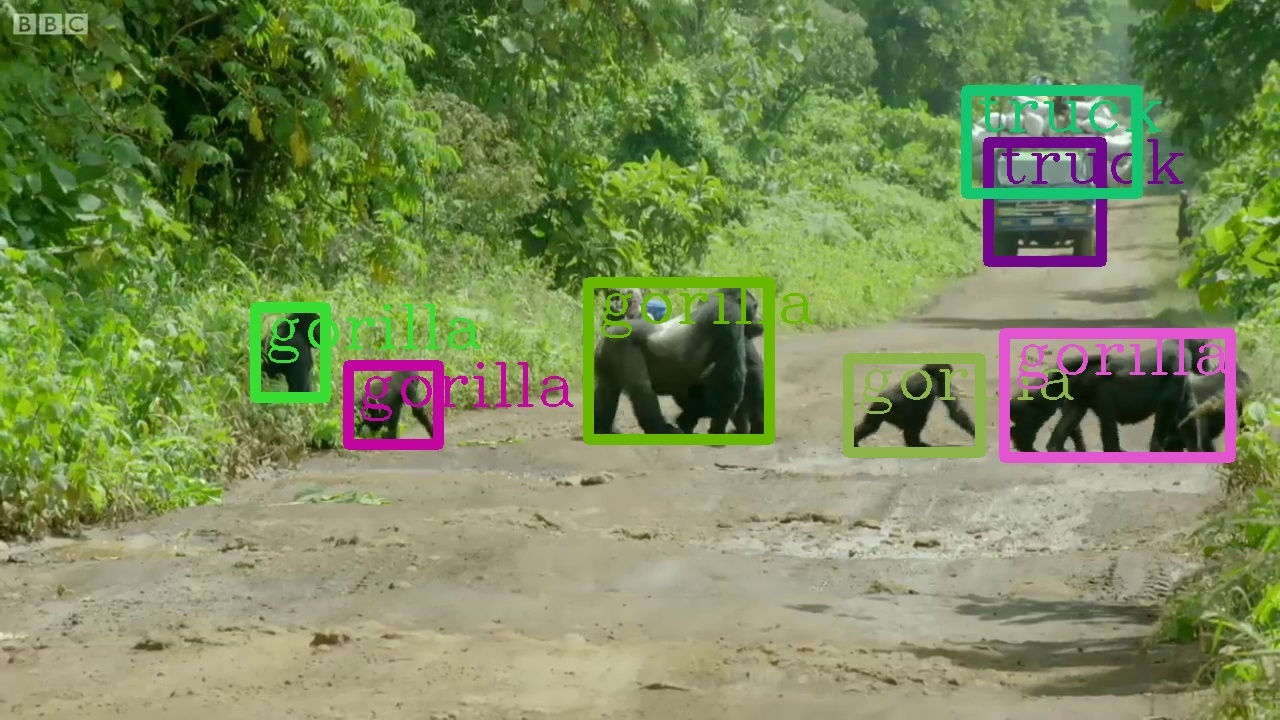} &
        \includegraphics[width=0.2\linewidth]{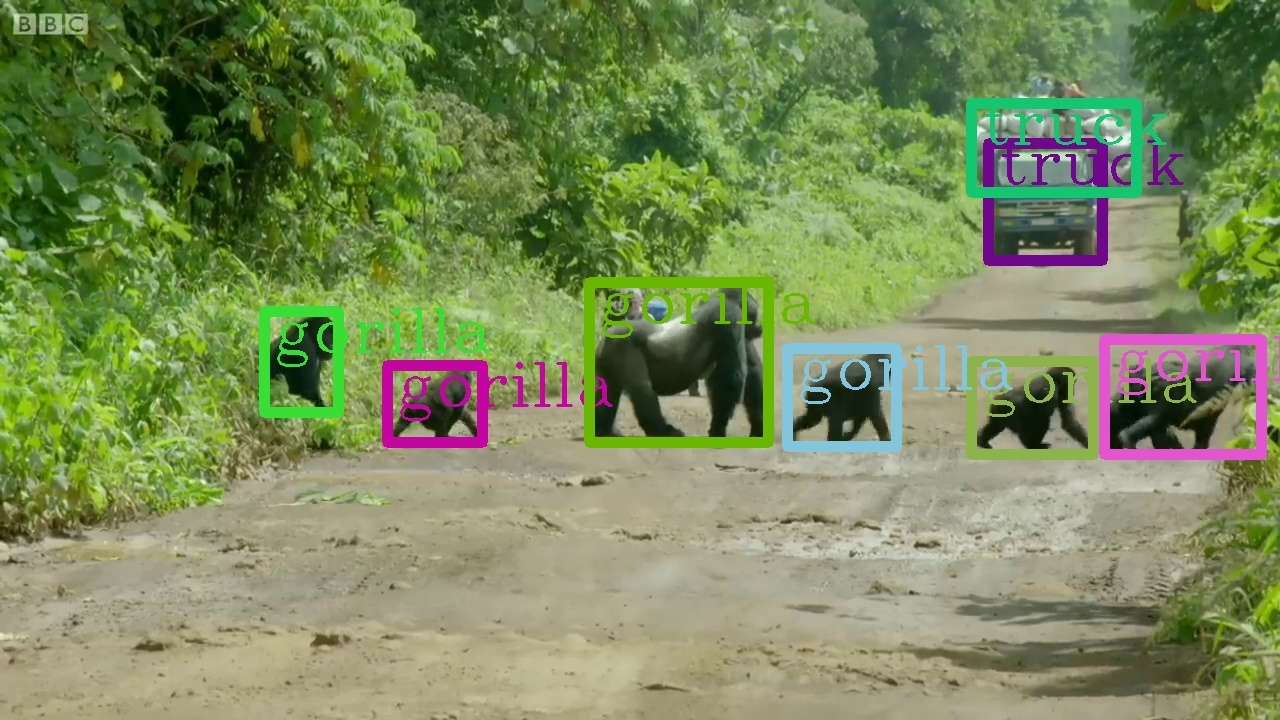} &
        \includegraphics[width=0.2\linewidth]{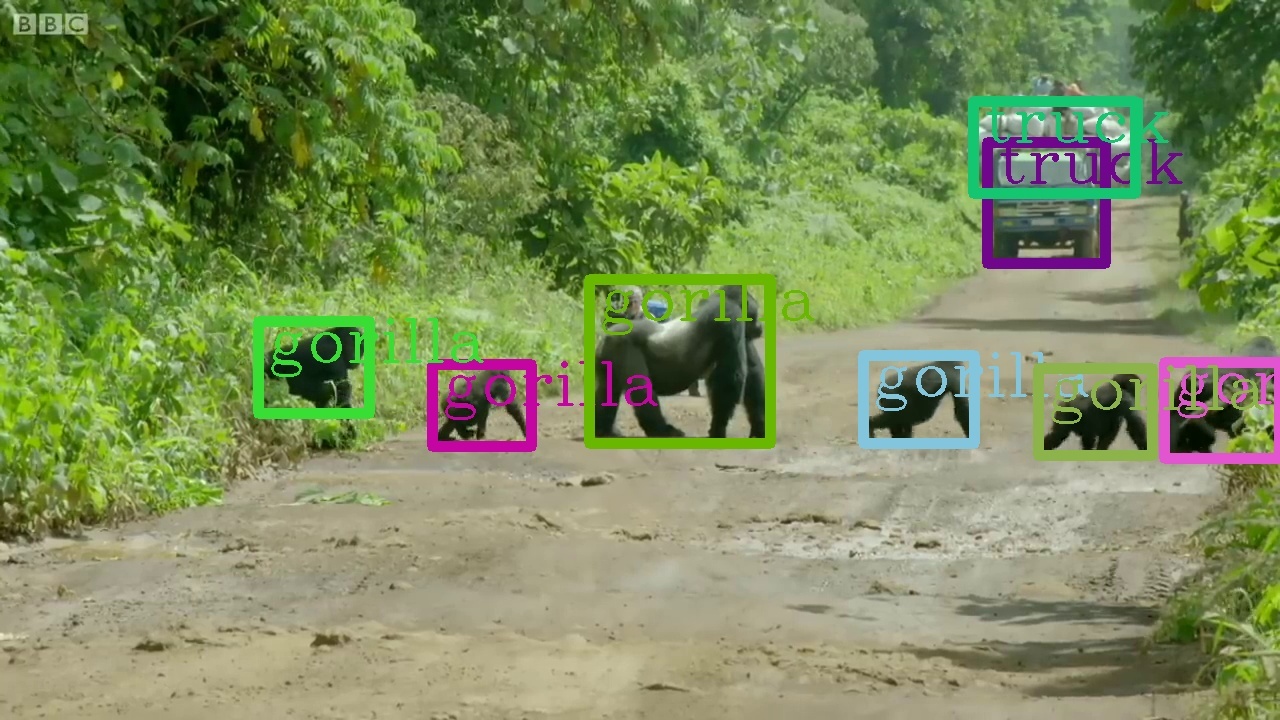} &
        \includegraphics[width=0.2\linewidth]{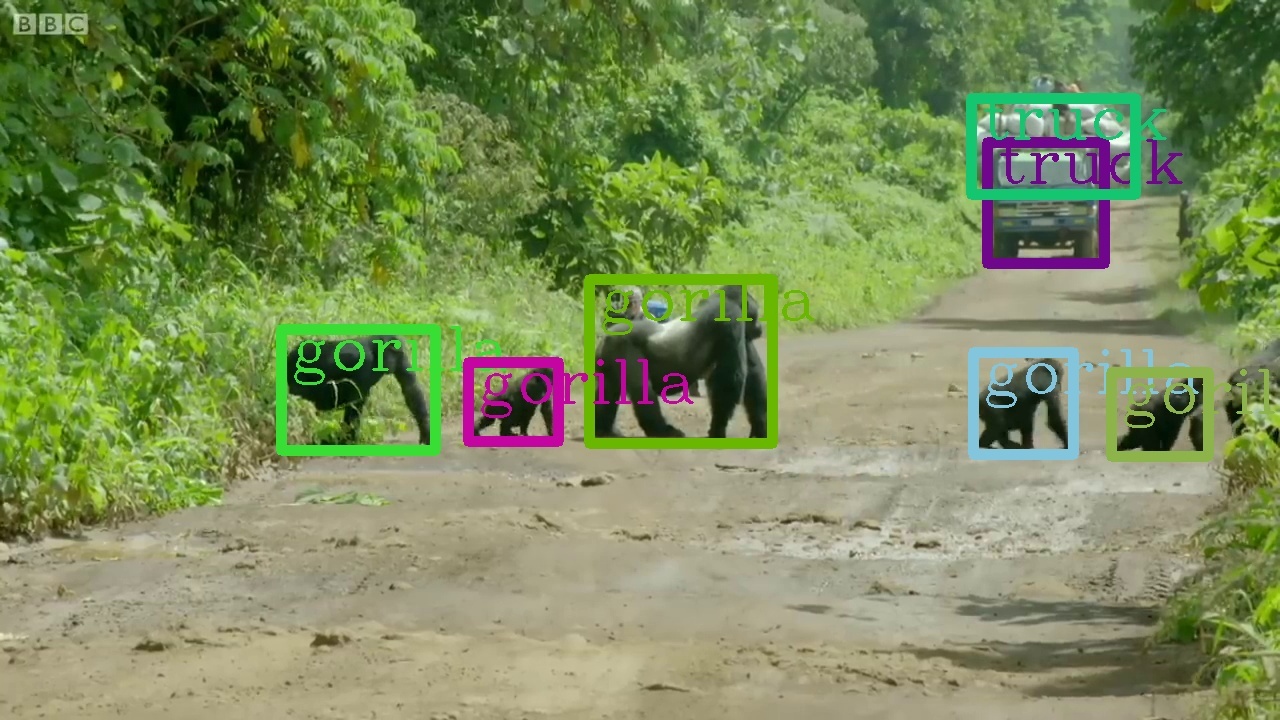} \\
        
        \includegraphics[width=0.2\linewidth]{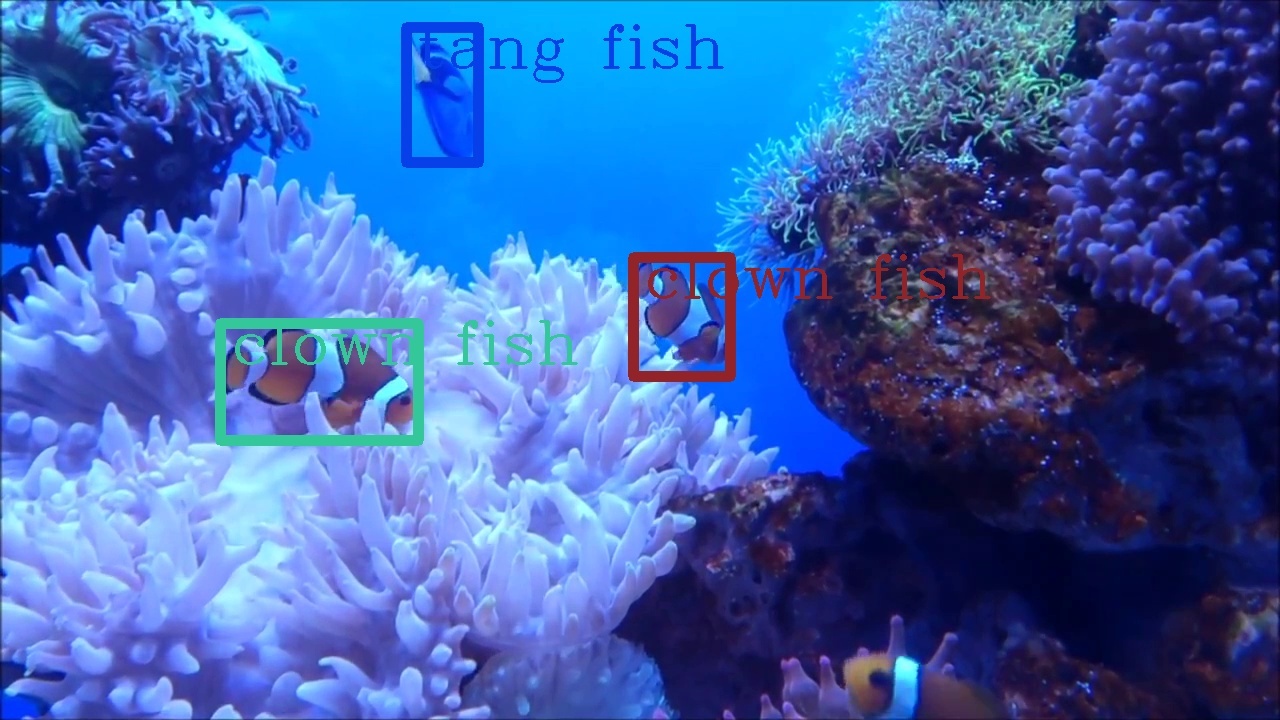} &
        \includegraphics[width=0.2\linewidth]{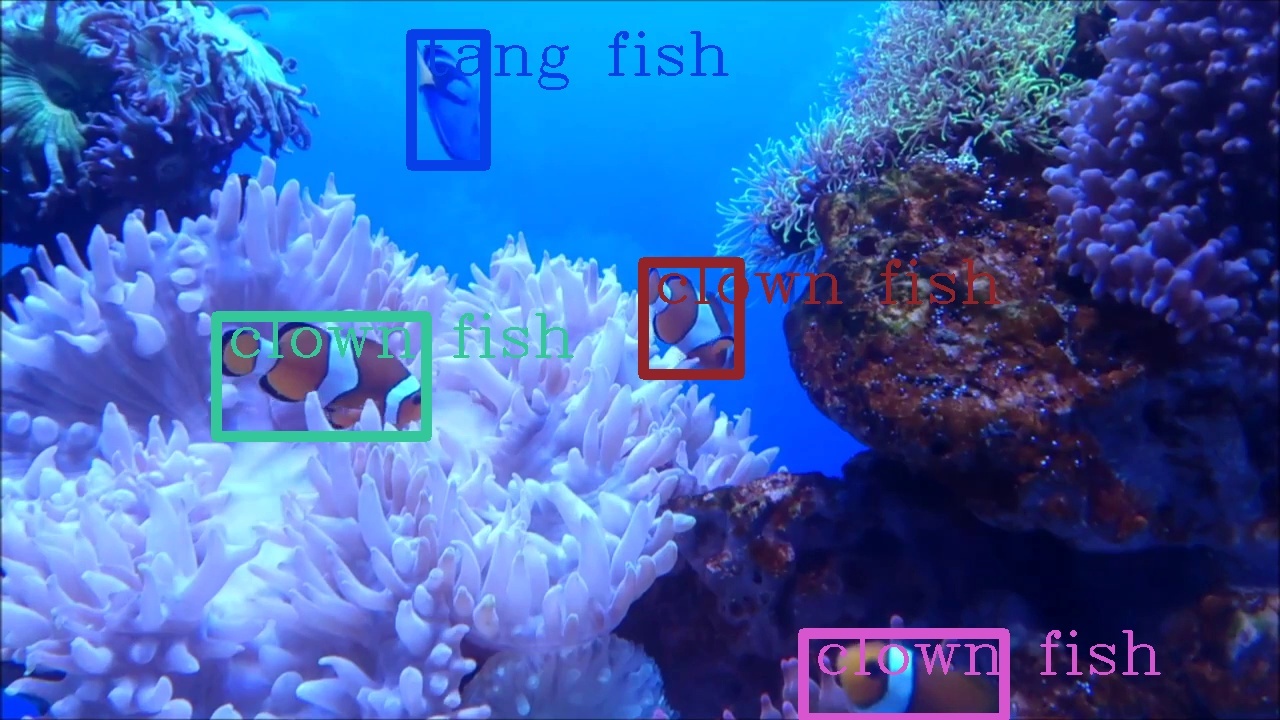} &
        \includegraphics[width=0.2\linewidth]{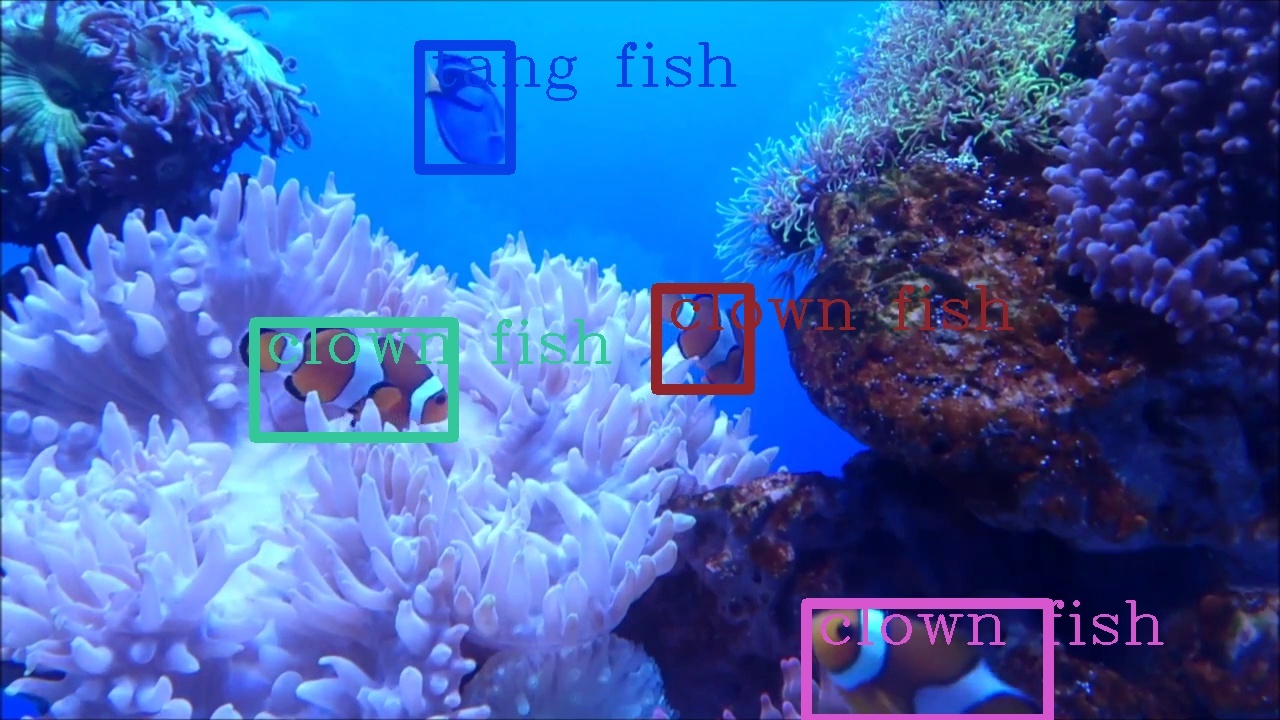} &
        \includegraphics[width=0.2\linewidth]{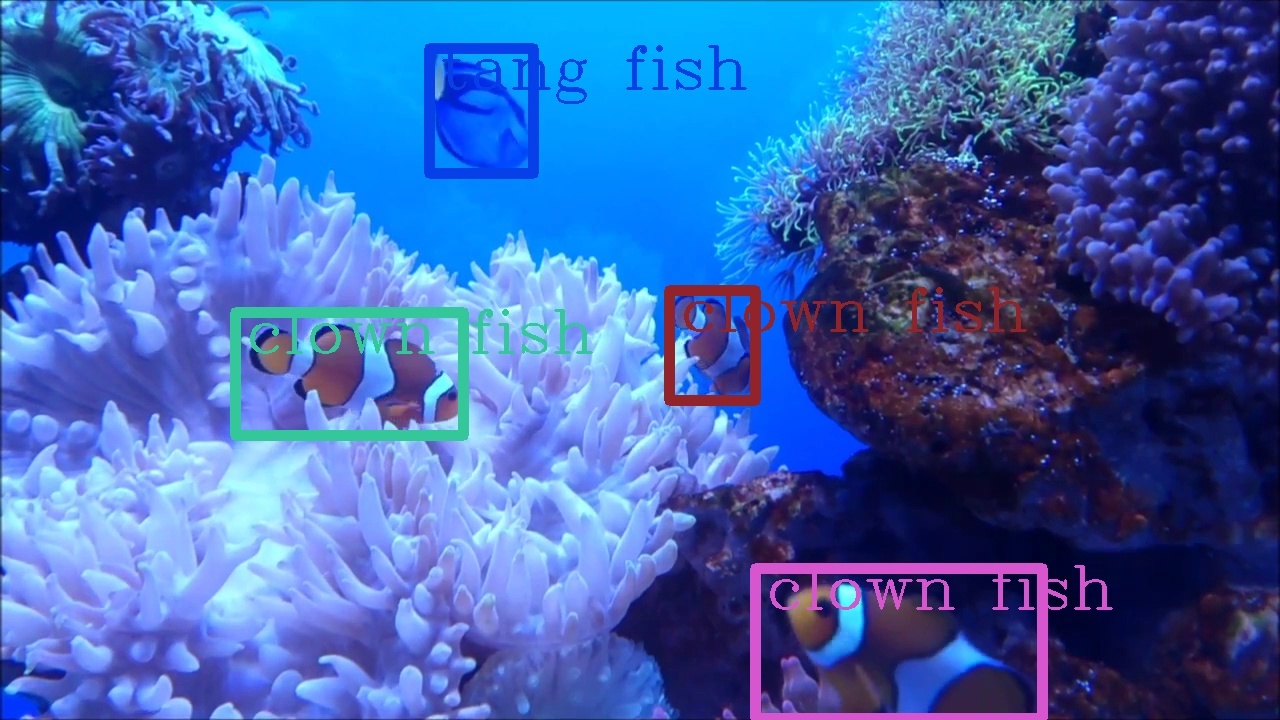} &
        \includegraphics[width=0.2\linewidth]{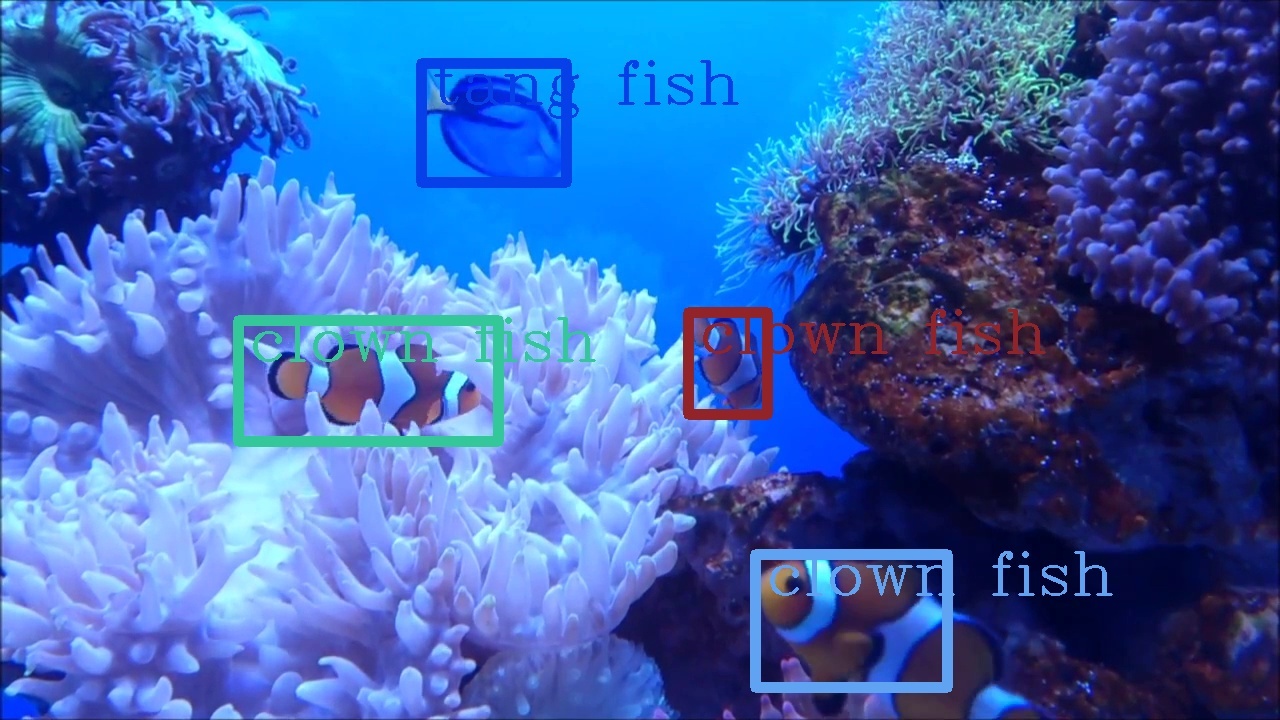} \\
        
        \includegraphics[width=0.2\linewidth]{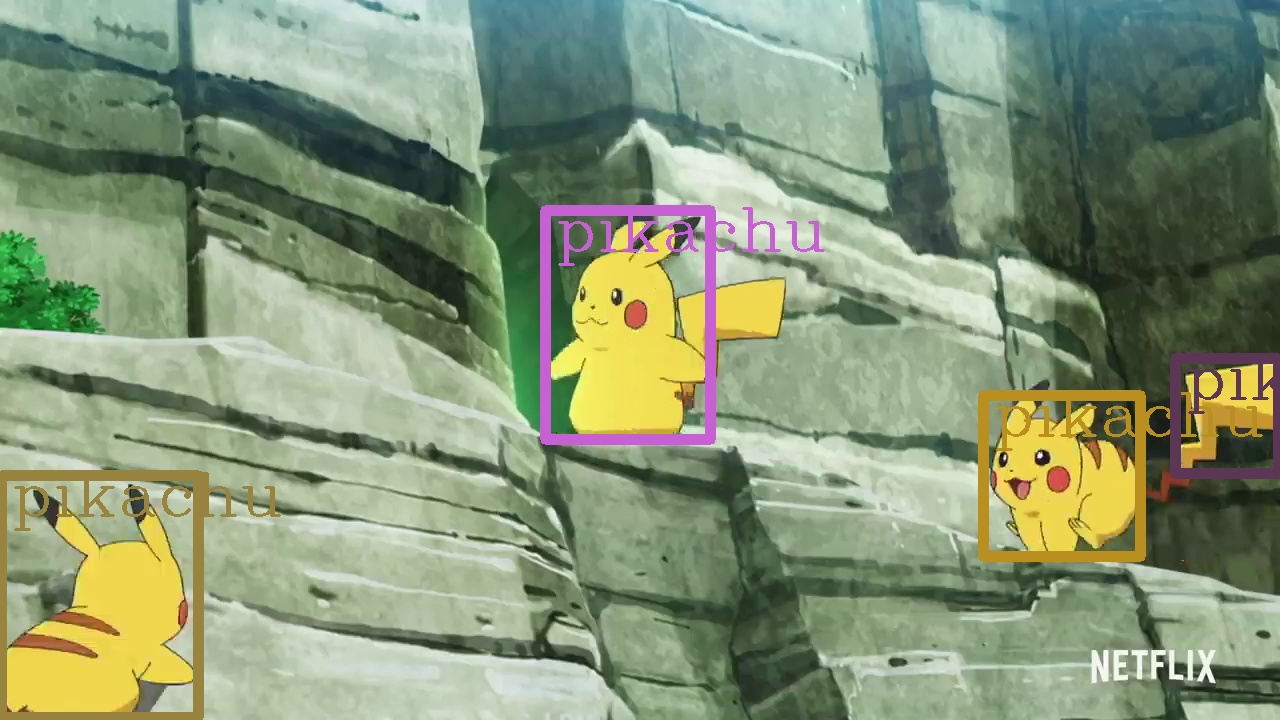} &
        \includegraphics[width=0.2\linewidth]{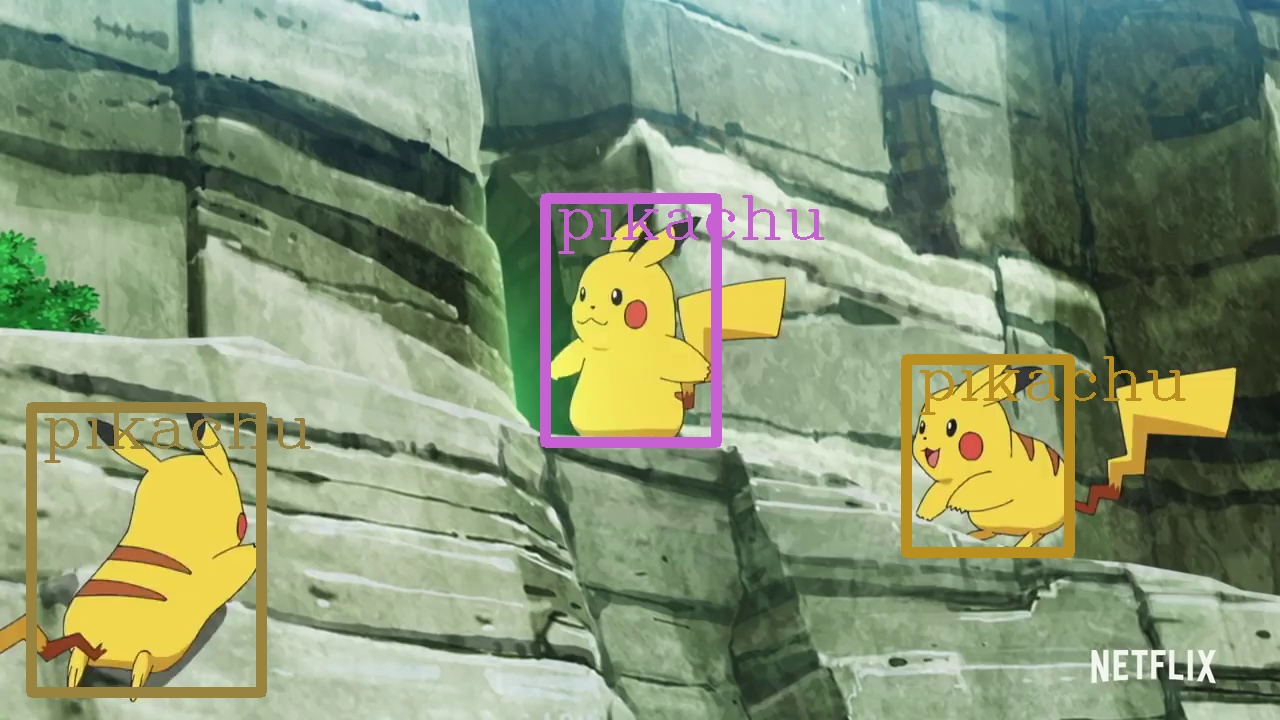} &
        \includegraphics[width=0.2\linewidth]{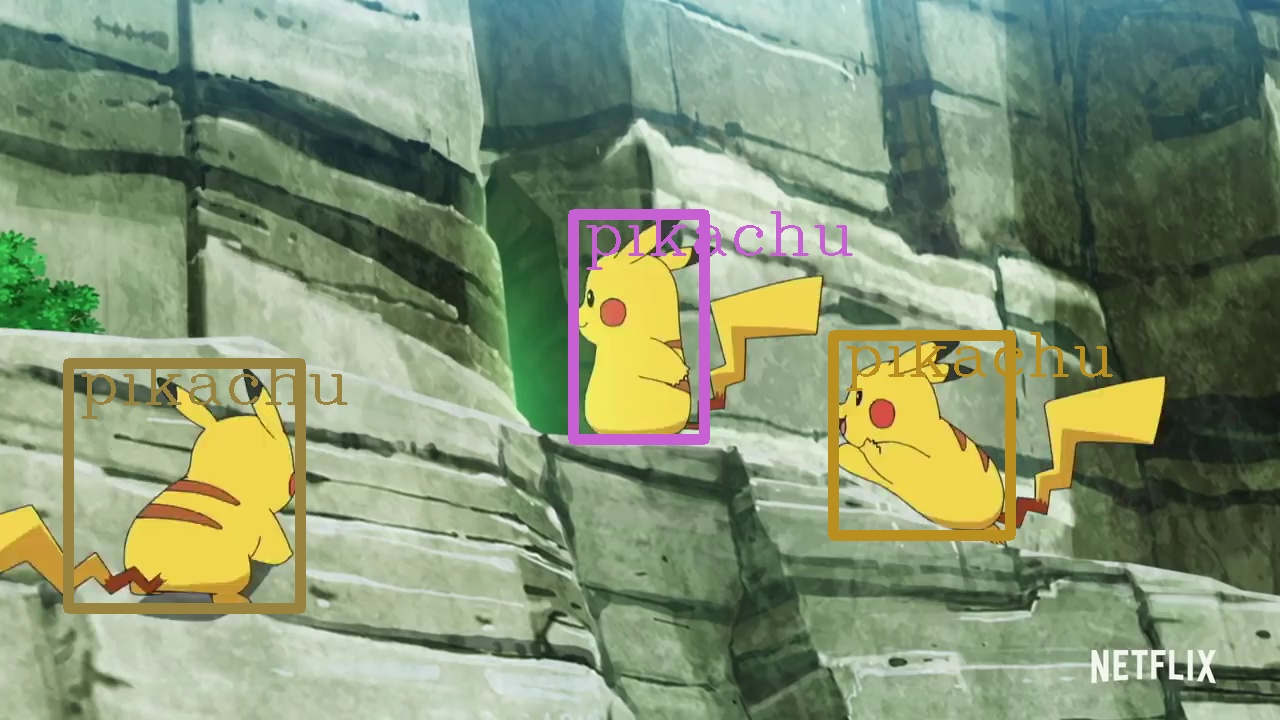} &
        \includegraphics[width=0.2\linewidth]{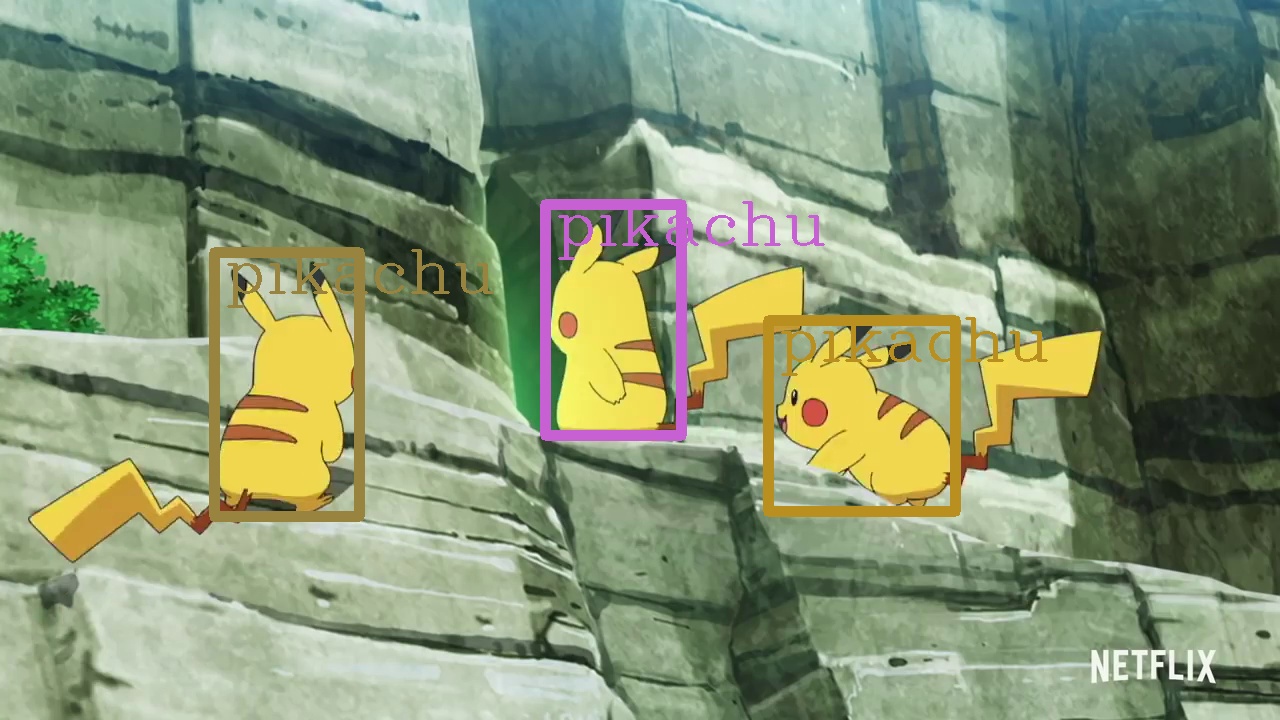} &
        \includegraphics[width=0.2\linewidth]{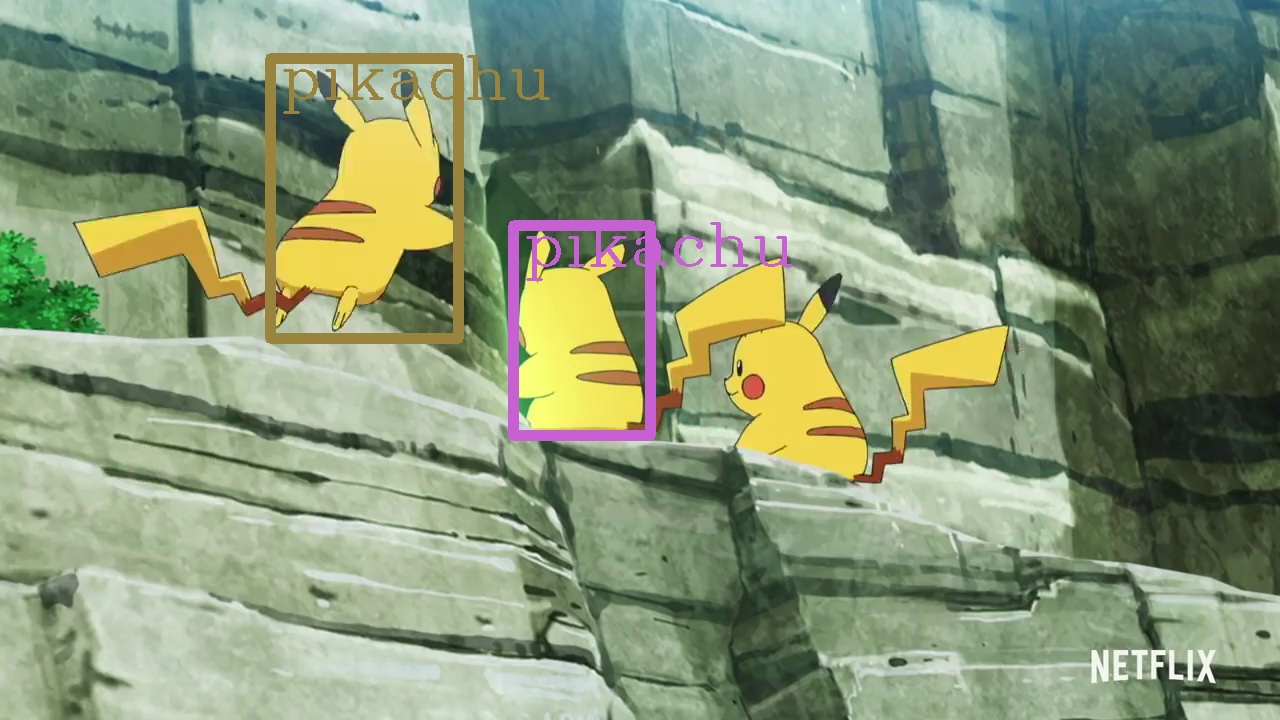} \\

        \includegraphics[width=0.2\linewidth]{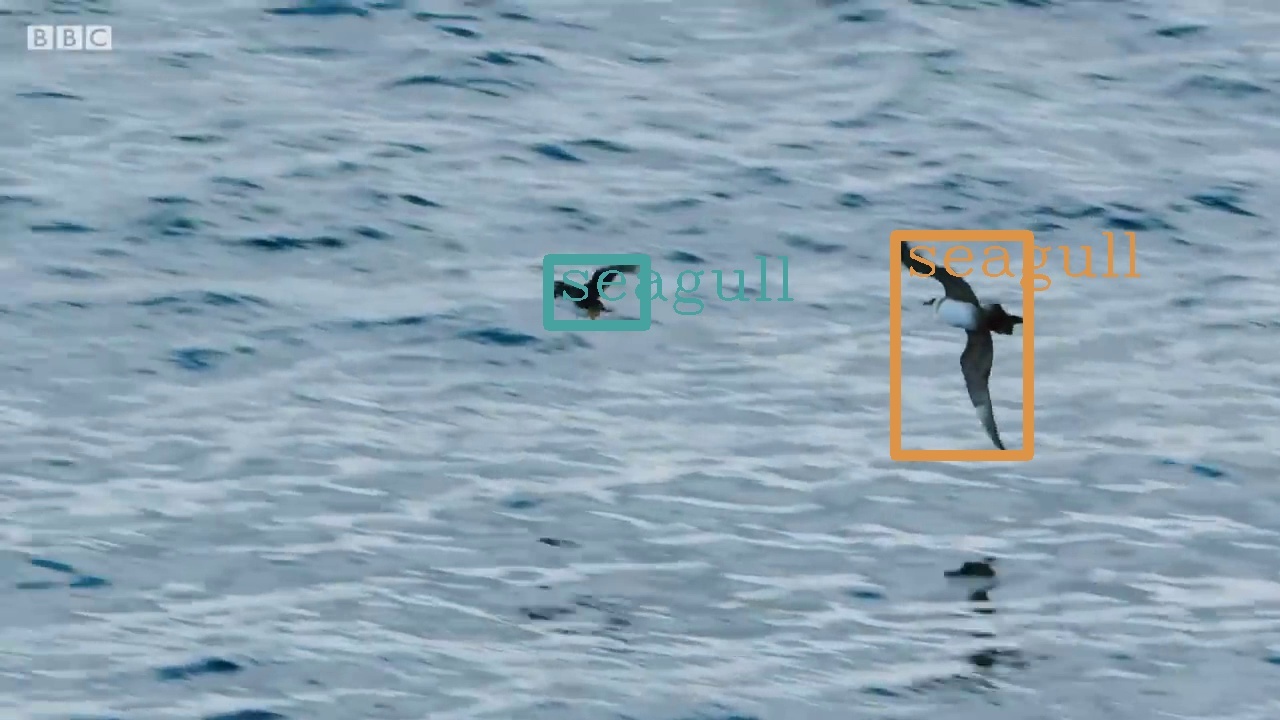} &
        \includegraphics[width=0.2\linewidth]{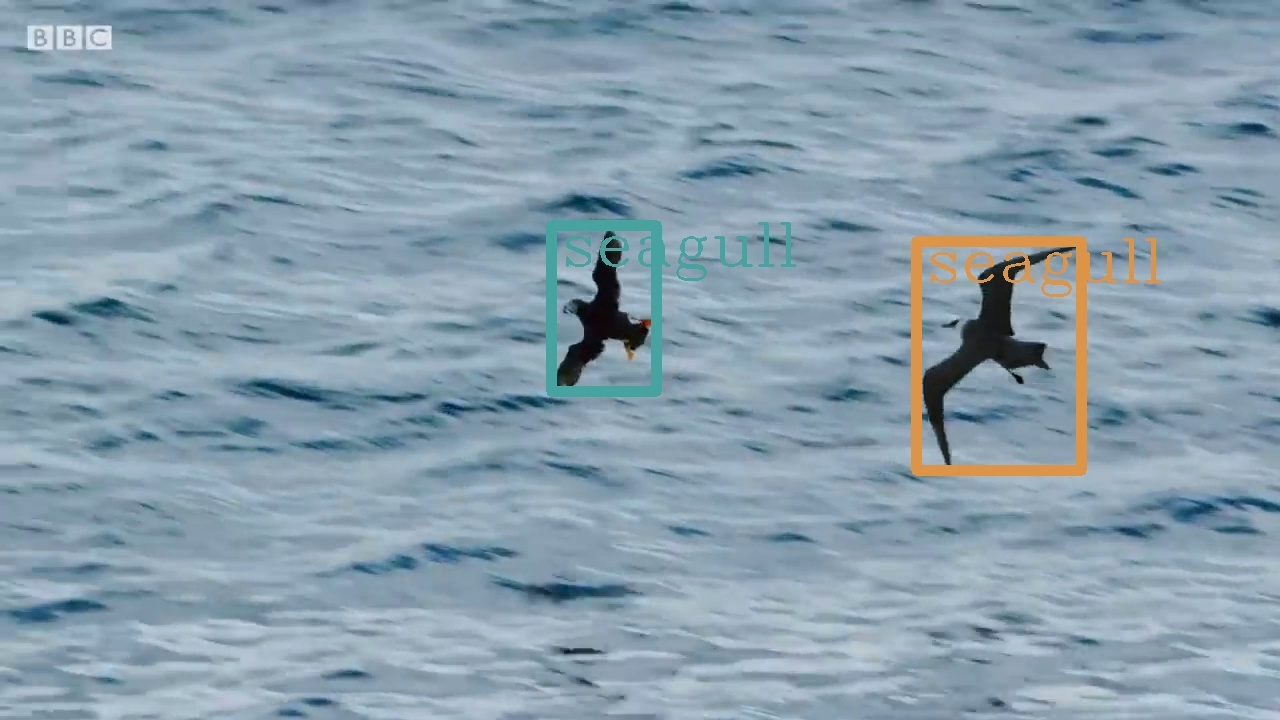} &
        \includegraphics[width=0.2\linewidth]{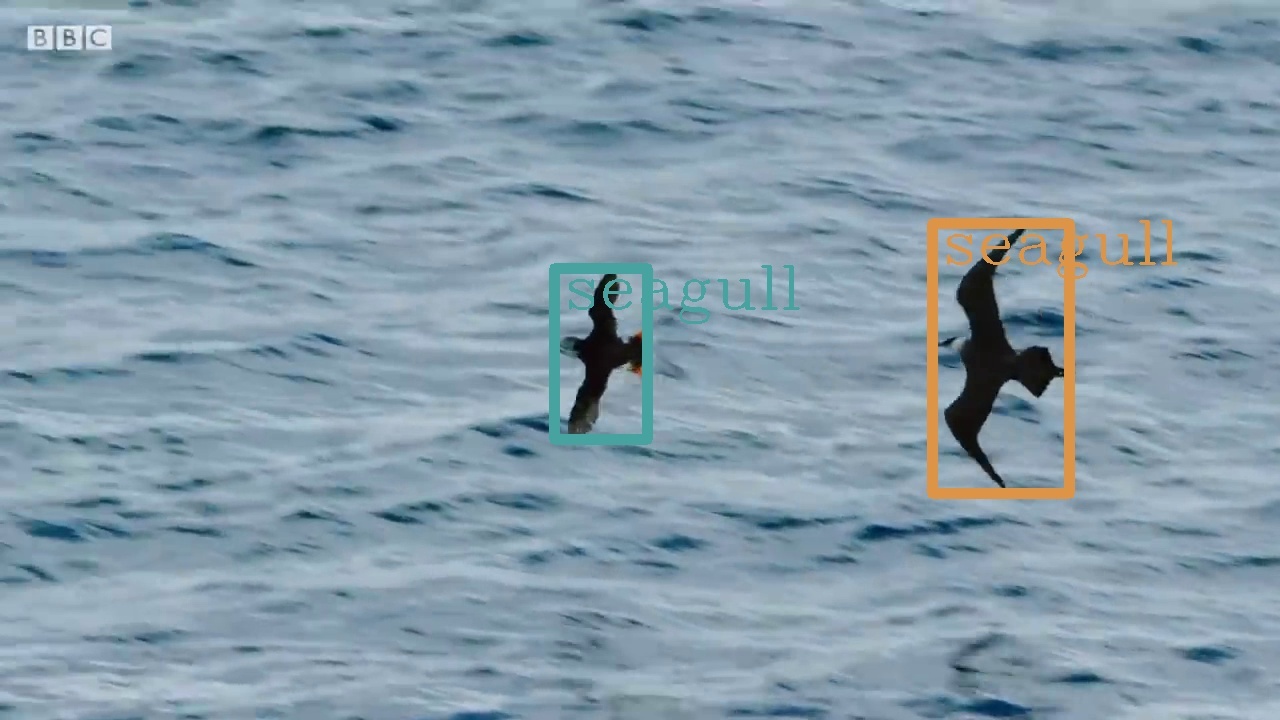} &
        \includegraphics[width=0.2\linewidth]{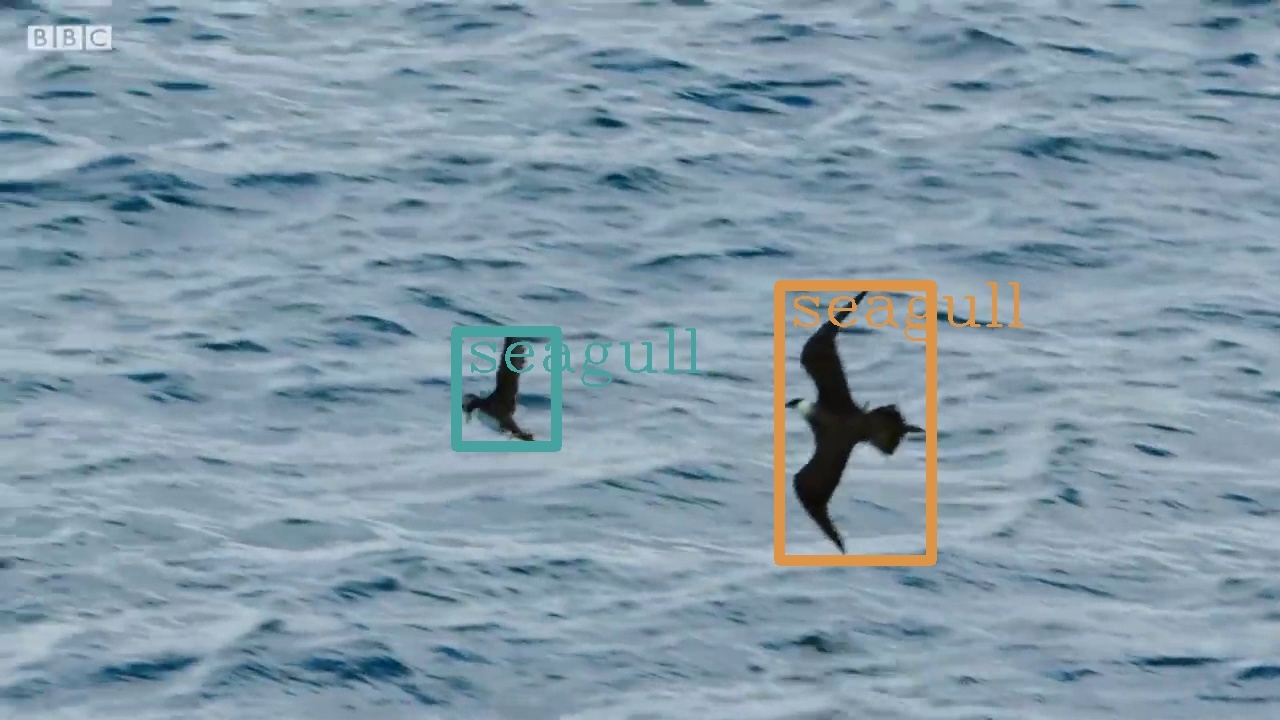} &
        \includegraphics[width=0.2\linewidth]{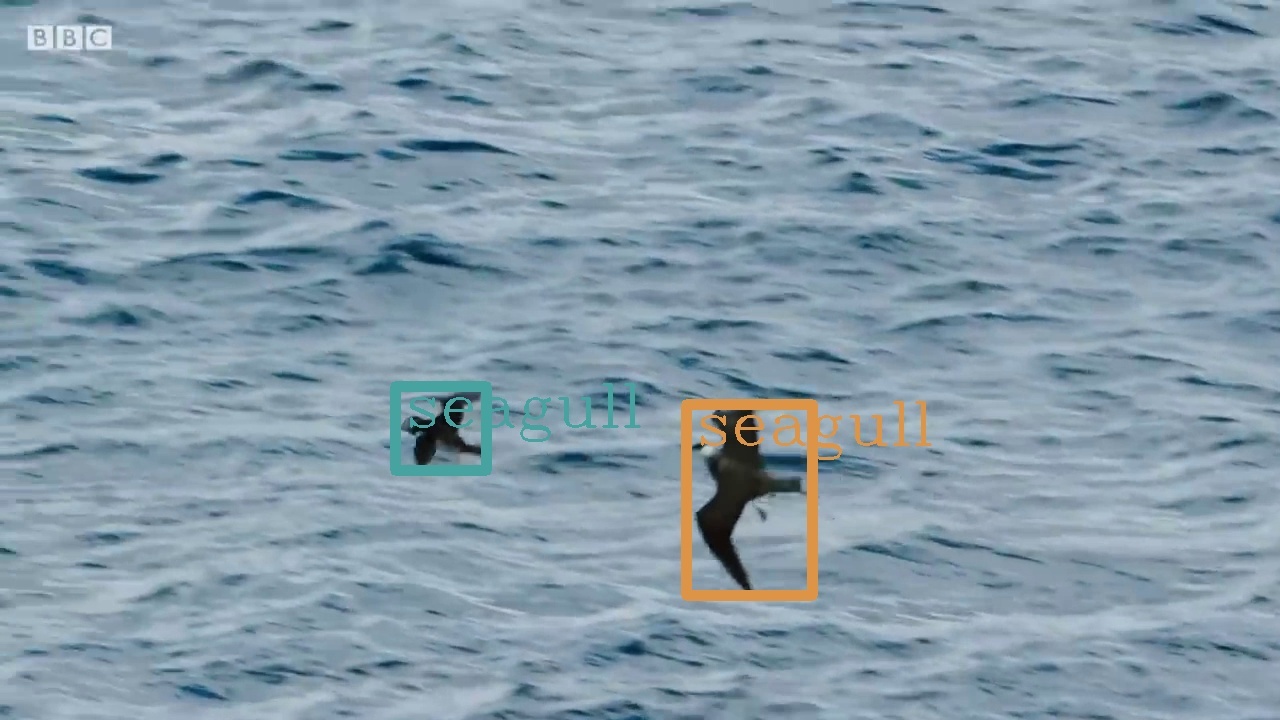} \\
        \bottomrule
    \end{tabular}}
    \caption{\textbf{\modelname qualitative results and failure cases.} We condition our tracker on text prompts unseen during training and successfully track the corresponding objects in the videos. The box color depicts object identity. We choose random internet videos to test our algorithm on diverse real-world scenarios. The bottom row shows the difficulty of fine-grained classification, where our method fails to distinguish the puffin from the sea gull. Best viewed digitally.
    }
    \label{fig:qualitative}
\end{figure*}

\begin{figure*}[t]
    \centering
    \small
    \setlength\tabcolsep{0.5mm}
    \resizebox{1.0\linewidth}{!}{
    \begin{tabular}{cc|cc}
        \toprule
        Generated & Original & Generated & Original \\ \midrule
        \includegraphics[trim={0 3cm 0 6cm},clip,width=0.25\linewidth]{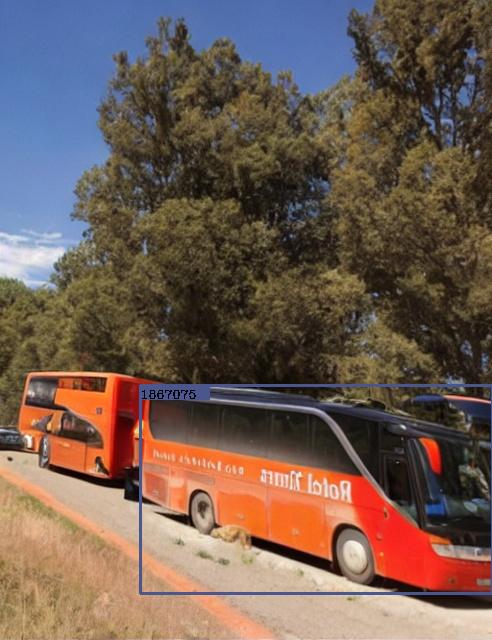} &
        \includegraphics[trim={0 3cm 0 6cm},clip,width=0.25\linewidth]{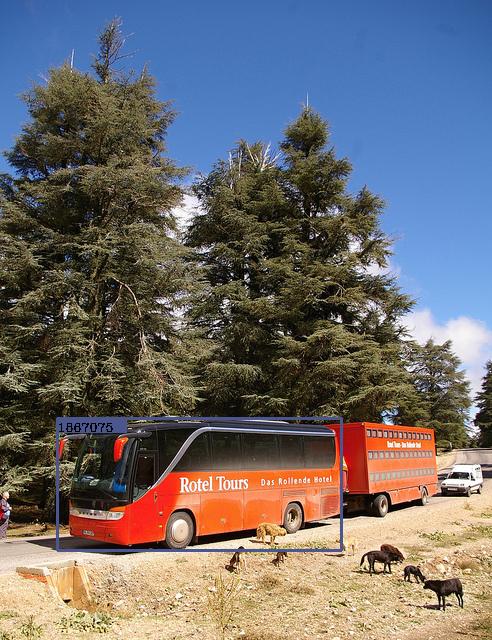} &
        \includegraphics[width=0.25\linewidth]{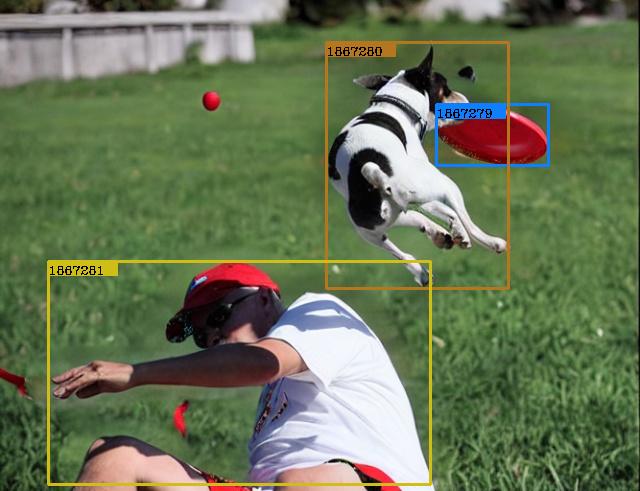} &
        \includegraphics[width=0.25\linewidth]{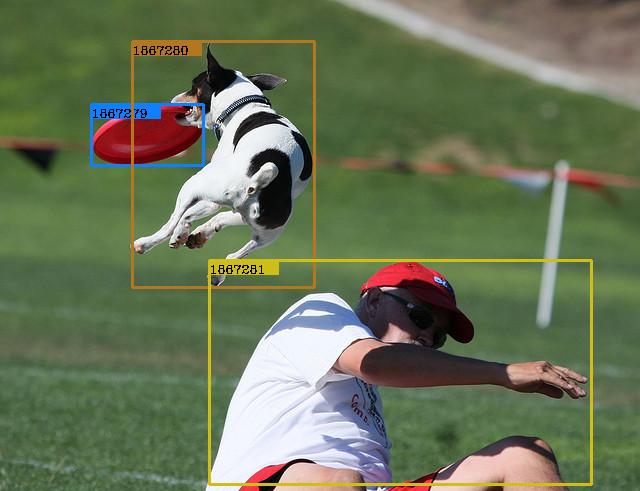} \\

        \includegraphics[trim={1.5cm 0 1.5cm 0},clip,width=0.25\linewidth]{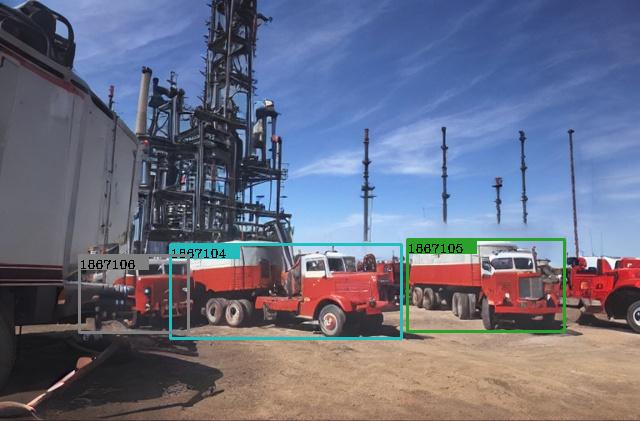} &
        \includegraphics[trim={1.5cm 0 1.5cm 0},clip,width=0.25\linewidth]{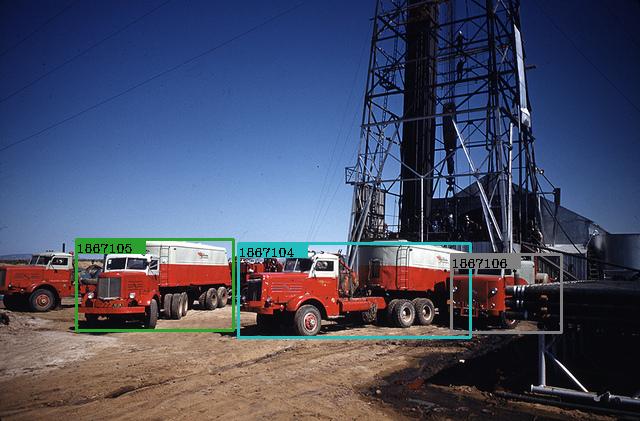} &
        \includegraphics[width=0.25\linewidth]{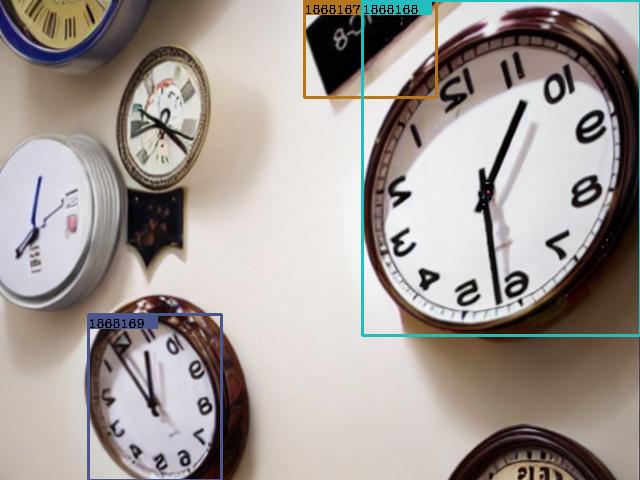} &
        \includegraphics[width=0.25\linewidth]{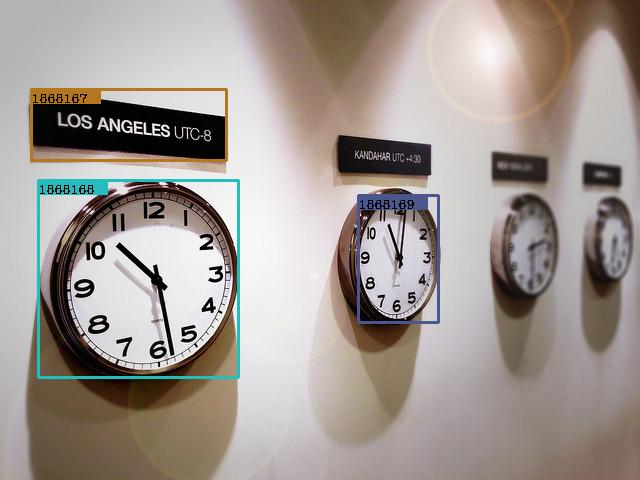} \\

        \includegraphics[width=0.25\linewidth]{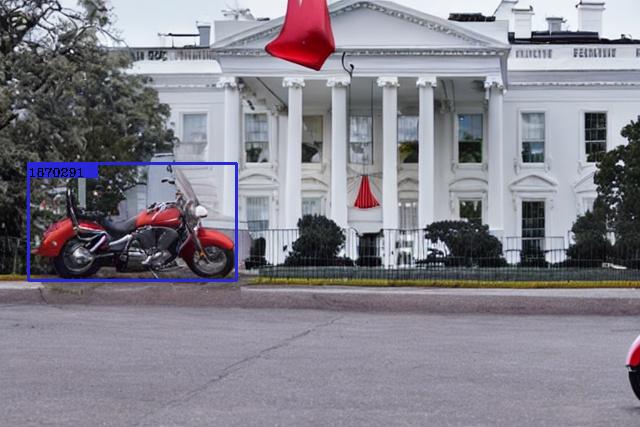} &
        \includegraphics[width=0.25\linewidth]{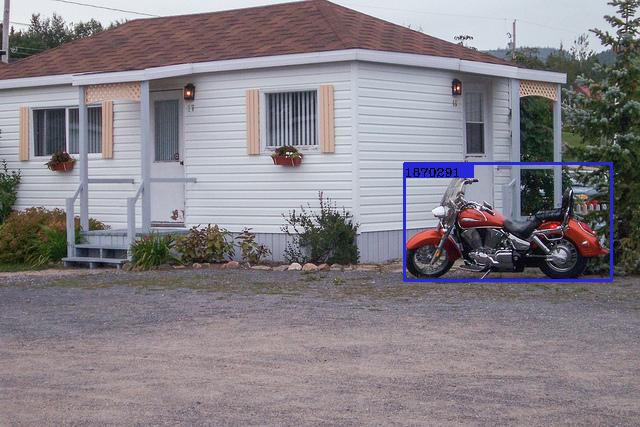} &
        \includegraphics[width=0.25\linewidth]{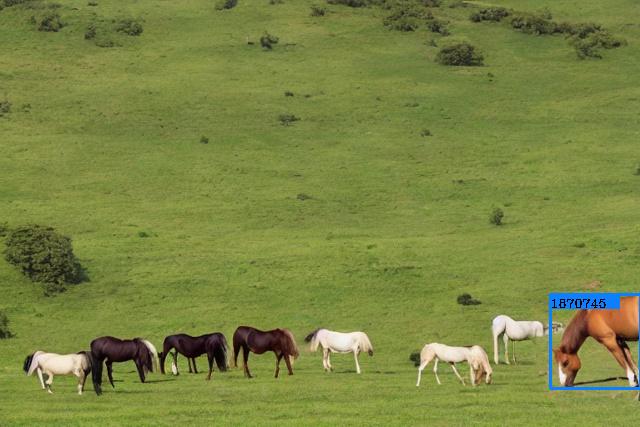} &
        \includegraphics[width=0.25\linewidth]{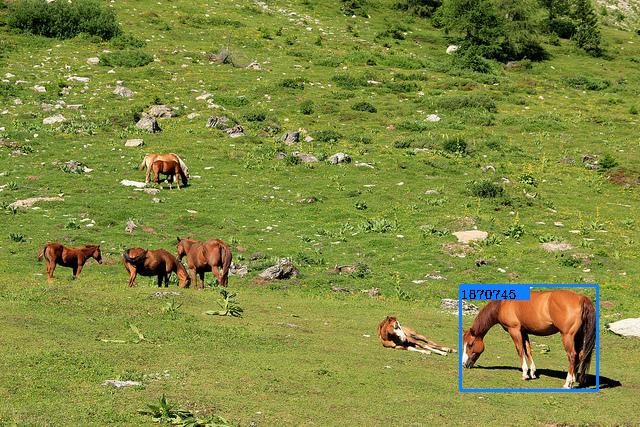} \\

        \includegraphics[width=0.25\linewidth]{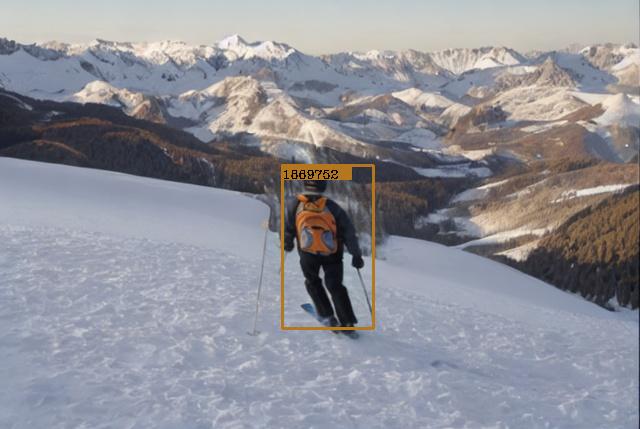} &
        \includegraphics[width=0.25\linewidth]{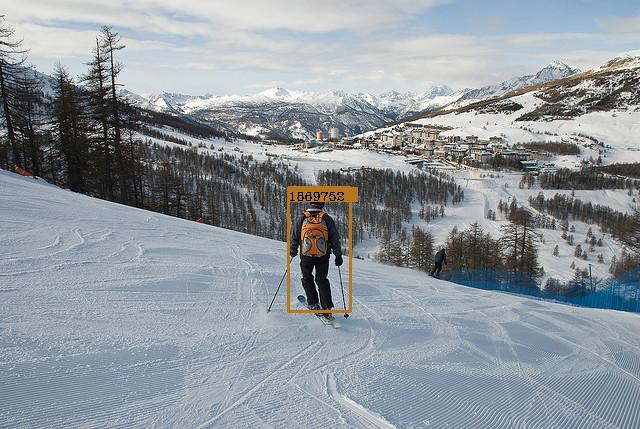} &
        \includegraphics[width=0.25\linewidth]{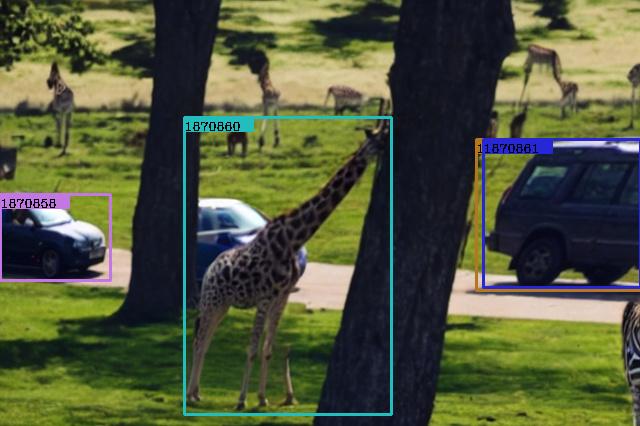} &
        \includegraphics[width=0.25\linewidth]{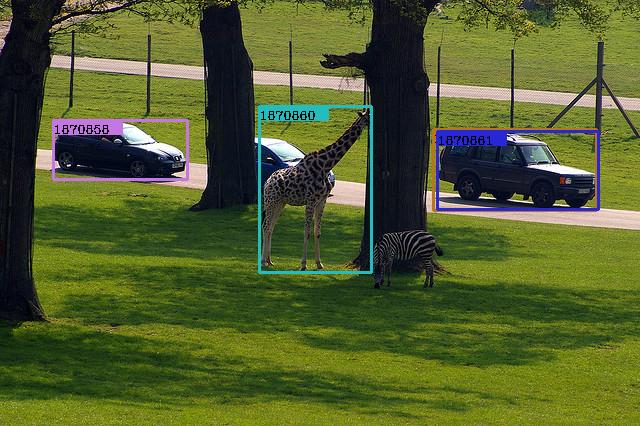} \\

        \includegraphics[width=0.25\linewidth]{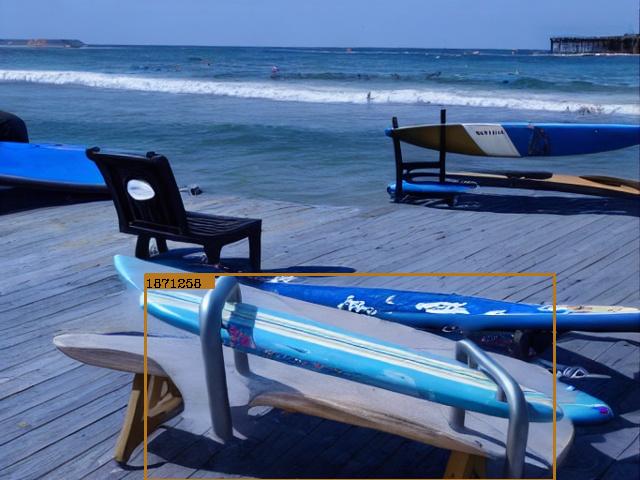} &
        \includegraphics[width=0.25\linewidth]{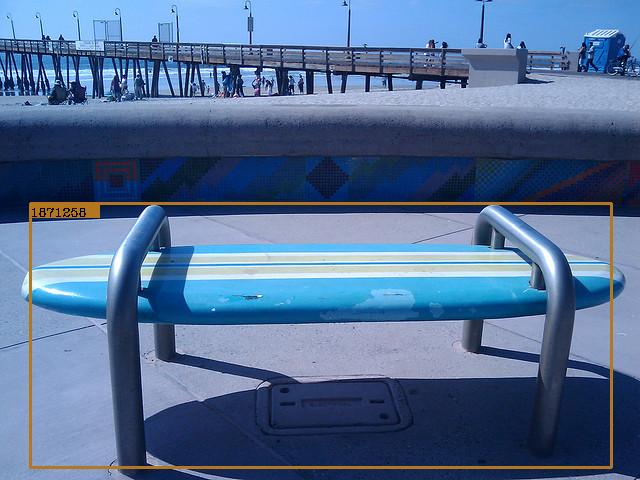} &
        \includegraphics[trim={0.5cm 0 0.5cm 0},clip,width=0.25\linewidth]{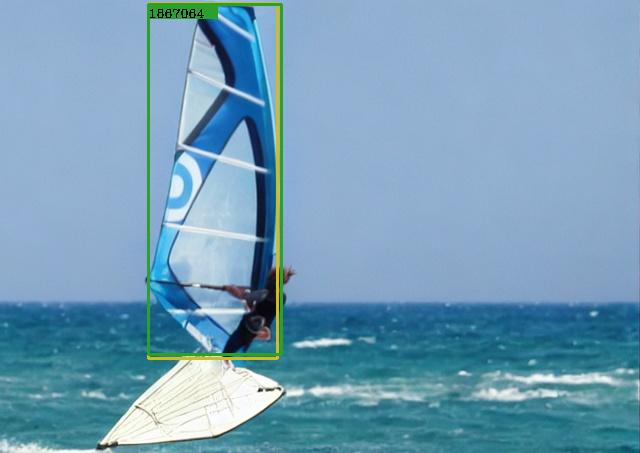} &
        \includegraphics[trim={0.5cm 0 0.5cm 0},clip,width=0.25\linewidth]{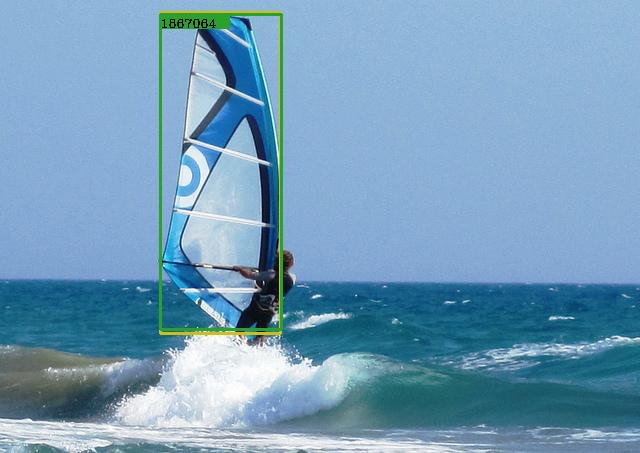} \\
        \bottomrule
    \end{tabular}}
    \caption{\textbf{Data hallucination examples.} We provide examples of our data hallucination strategy including annotations on the LVIS dataset. We plot the generated versions and the original for comparison. The ids on the bounding boxes depict the identity.\vspace{4mm}
    }
    \label{fig:qualitative_diffusion}
\end{figure*}

\subsection{Qualitative results}
\label{sec:qualitative}
 
For the qualitative results in this supplementary material, we set $\gamma = \frac{1}{|\mathcal{C}| + 1}$ where $\mathcal{C}$ is the number of prompts in the video to have a more rigorous detection filtering.

 \parsection{Data hallucination strategy.} We visualize the results of different hyperparameters in our diffusion process in Fig.~\ref{fig:hallucination}. We choose the parameters with the visually most appealing results, $\delta_0 = 0.75$, $K=50$ and $\eta = 0.01$. We observe that choosing a too high $\delta_0$ leads to divergence from the original image content, while too little noise leads to insufficient fidelity. Increasing the number of iterations $K$ does not lead to an obvious improvement in visual quality, so we choose $K=50$ to speed up the image generation process. Finally, having no homogenization,~\ie setting $\eta = 0.0$ leads to noticeable artifacts. On the other hand, a higher $\eta$ of $0.1$ leads to subtle, but significant appearance perturbation of the object, which is also undesirable for preserving its identity.
 
 In addition, we illustrate examples of our final data hallucination strategy in Fig.~\ref{fig:qualitative_diffusion}. We visualize examples from the LVIS dataset, where in each row we plot both annotations and images and show the generated versions and the original images.

 \parsection{Qualitative results and failure cases.} We show qualitative results and failure cases of our method in Fig.~\ref{fig:qualitative}. 
 We observe that our method does well on tracking, and is able to generalize even to very exotic classes, such as pikachu. However, fine-grained classification is still challenging. In particular, in the bottom row of the figure, our method fails to distinguish the sea gull from the puffin, wrongly classifying it as another sea gull. Furthermore, our detection is not perfect, as can be seen by the false negative in the 7th row ($t + 4$). In addition, the 6th row exhibits an ID switch between $t+ 3$ and $t + 4$.

{\small
\bibliographystyle{ieee_fullname}
\bibliography{egbib}
}

\end{document}